\documentclass[sigconf]{acmart}

\usepackage{url}
\usepackage{verbatim}
\usepackage{graphicx}
\usepackage{color}
\usepackage{algorithmic,algorithm}

\AtBeginDocument{%
  }
    
\setcopyright{none}
\renewcommand\footnotetextcopyrightpermission[1]{} 
\copyrightyear{}
\acmYear{}
\acmDOI{}

\acmConference[]{}{}{}
\acmPrice{}
\acmISBN{}

\begin{document}

\title{Quality-diversity in Dissimilarity Spaces}

\author{Steve Huntsman}
\email{steve.huntsman@str.us}
\orcid{0000-0002-9168-2216}


\begin{abstract}
The theory of magnitude provides a mathematical framework for quantifying and maximizing diversity. We apply this framework to formulate quality-diversity algorithms in generic dissimilarity spaces. In particular, we instantiate a very general version of Go-Explore with promising performance for challenging and computationally expensive objectives, such as arise in simulations. Finally, we prove a result on diversity at scale zero that is interesting in its own right and consider its implications for our algorithm.
\end{abstract}



\maketitle

\section{\label{sec:Introduction}Introduction}

The survival of a species under selection pressure is a manifestly challenging optimization problem, jointly solved by evolution many times over despite omnipresent maladaptation \cite{brady2019causes}. This suggests that many objective functions that are hard to optimize (and frequently, also hard to evaluate) may admit a diverse set of inputs that perform well even if they are not local extrema. \emph{Quality-diversity} (QD) algorithms \cite{pugh2016quality,chatzilygeroudis2021quality} such as NSLC \cite{lehman2011evolving}, MAP-Elites \cite{mouret2015illuminating}, and their offshoots, seek to produce such sets of inputs, typically by discretizing the input space into ``cells'' and returning the best input found for each cell. QD differs from multimodal optimization by exploring regions of input space that need not have extrema.
\footnote{
Another practical (if not theoretical) difference between QD and multimodal optimization algorithms is that the former are often explicitly intended to operate in a (possibly latent) low-dimensional behavioral or phenotypical space, but this is mostly irrelevant from the primarily algorithmic point of view we concern ourselves with here. The key algorithmic point is to consider a ``pullback'' dissimilarity: for this, see the archetypal class of objectives discussed in \S \ref{sec:GoExplore}.
}

The link between exploration and diversity in ecosystems is that ``nature abhors a vacuum in the animate world'' \cite{grinnell1924geography}. A similar link informs optimization algorithms \cite{hoffman2022benchmarking,huntsman2022diversity,huntsman2022parallel}. The key construction is a mathematically principled notion of diversity that generalizes information theory by incorporating geometry \cite{leinster2012measuring,leinster2021entropy}. It singles out the Solow-Polasky diversity \cite{solow1994measuring} or \emph{magnitude} of a finite space endowed with a symmetric dissimilarity in relation to the ``correct'' definition \eqref{eq:diversity} of diversity that uniquely satisfies various natural desiderata. The uniqueness of a diversity-maximizing probability distribution has been established \cite{leinster2016maximizing}, though to our knowledge the only applications to optimization are presently \cite{hoffman2022benchmarking,huntsman2022diversity,huntsman2022parallel}.

In this paper, which elaborates on the conference paper \cite{huntsman2023quality}, we apply the notion \eqref{eq:diversity} of diversity to QD algorithms for the first time by producing a formalization of the breakthrough Go-Explore framework \cite{ecoffet2021first} suited for computationally expensive objectives in very general settings. The algorithm requires very little tuning and its only requirements are
\begin{itemize}
	\item a symmetric, nondegenerate dissimilarity (not necessarily satisfying the triangle inequality) that is efficient to evaluate;
	\item a mechanism for globally generating points in the input space (which need not span the entire space, since we can use the output of one run of the algorithm to initialize another);
	\item an efficient mechanism for locally perturbing existing points;
	\item and a mechanism for estimating the objective that permits efficient evaluation: e.g., interpolation using polyharmonic radial basis functions \cite{buhmann2003radial} or a neural network.
\end{itemize}
Other than these, the algorithm's only other inputs are a handful of integer parameters that govern the discretization of the input space and the effort devoted to evaluating the objective and its cheaper estimate. Our examples repeatedly reuse many values for these.

The paper is organized as follows. In \S \ref{sec:WMD}, we introduce the concepts of magnitude and diversity. In \S \ref{sec:GoExplore}, we outline the Go-Explore framework in the context of dissimilarity spaces. In \S \ref{sec:go} we construct a probability distribution for ``going'' that balances exploration and exploitation. We discuss local exploration mechanisms in \S \ref{sec:explore}. In \S \ref{sec:examples} we provide a diverse set of examples. Finally, in \S \ref{sec:zeroScale} we analyze the effects of considering an extremal notion of diversity before concluding in \S \ref{sec:conclusion}. 

\section{\label{sec:WMD}Magnitude and diversity}

For details on the ideas in this section, see \S 6 of \cite{leinster2021entropy}; see also \cite{huntsman2022diversity} for a slightly more elaborate retelling.

The notion of magnitude that we will introduce below has been used by ecologists to \emph{quantify} diversity since the work of Solow and Polasky \cite{solow1994measuring}, but much more recent mathematical developments have clarified the role that magnitude and the underlying concept of weightings play in \emph{maximizing} a more general and axiomatically supported notion of diversity \cite{leinster2016maximizing,leinster2021entropy}. We will describe these concepts in reverse order, moving from diversity to weightings and magnitude in turn before elaborating on them.

A square nonnegative matrix $Z$ is a \emph{similarity matrix} if its diagonal is strictly positive. Now the \emph{diversity of order $q$} for a probability distribution $p$ and similarity matrix $Z$ on the same space is 
\begin{equation}
\label{eq:diversity}
D_q^Z(p) := \exp \left ( \frac{1}{1-q} \log \sum_{j: p_j > 0} p_j (Zp)_j^{q-1} \right )
\end{equation} 
for $1 < q < \infty$, and via limits for $q = 1,\infty$.
\footnote{
The logarithm of \eqref{eq:diversity} is a ``similarity-sensitive'' generalization of the R\'enyi entropy of order $q$. For $Z = I$, the R\'enyi entropy is recovered, with Shannon entropy for $q = 1$.
}
This is the ``correct'' measure of diversity in essentially the same way that Shannon entropy is the ``correct'' measure of information.

If the similarity matrix $Z$ is symmetric, then it turns out that the diversity-maximizing distribution $\arg \max_p D_q^Z(p)$ is actually independent of $q$. There is an algorithm to compute this distribution that we will discuss below, and in practice we can usually perform a nonlinear scaling of $Z$ to ensure the distribution is efficiently computable. We restrict attention to similarity matrices of the form 
\begin{equation}
\label{eq:Z}
Z = \exp[-td]
\end{equation} 
where $(\exp[M])_{jk} := \exp(M_{jk})$,
$t \in (0,\infty)$ is a scale parameter, and $d$ is a square symmetric \emph{dissimilarity matrix}: i.e., its entries are in $[0,\infty]$, with zeros on and only on the diagonal.
\footnote{
Henceforth we assume symmetry and nondegeneracy for $d$ unless stated otherwise. 
We will also write $d$ for a symmetric, nondegenerate \emph{dissimilarity} $d : X^2 \rightarrow [0,\infty]$ with $d(x,x) \equiv 0$ and $x \ne x' \Rightarrow d(x,x') \ne 0$. 
Here $X$ is called a \emph{dissimilarity space}. 
Note that $d$ (as a matrix or function) is \emph{not} assumed to satisfy the triangle inequality.
}

A \emph{weighting} $w$ is a vector satisfying 
\begin{equation}
\label{eq:weighting}
Zw = 1,
\end{equation} 
where the vector of all ones is indicated on the right. A \emph{coweighting} is the transpose of a weighting for $Z^T$. If $Z$ has both a weighting $w$ and a coweighting, then its \emph{magnitude} is $\sum_j w_j$, which also equals the sum of the coweighting components. In particular, if $Z$ is invertible then its magnitude is $\sum_{jk} (Z^{-1})_{jk}$. 
The theory of weightings and magnitude provide a very attractive and general notion of size that 
encodes rich scale-dependent geometrical data \cite{leinster2017magnitude}. 

\begin{example}
\label{ex:3PointSpace}
Take $\{x_j\}_{j=1}^3$ with $d_{jk} := d(x_j,x_k)$ given by $d_{12} = d_{13} = 1 = d_{21} = d_{31}$ and $d_{23} = \delta = d_{32}$ with $\delta<2$. A straightforward calculation of \eqref{eq:weighting} using \eqref{eq:Z} 
is shown in
Figure \ref{fig:3pointSpace20210402} for $\delta = 10^{-3}$. At $t = 10^{-2}$, the ``effective size'' of the nearby points is $\approx 0.25$, and that of the distal point is $\approx 0.5$, so at this scale the ``effective number of points'' is $\approx 1$. At $t = 10$, 
the effective number of points is $\approx 2$. Finally, at $t = 10^4$, 
the effective number of points is $\approx 3$. 

\begin{figure}[h]
  \centering
  \includegraphics[trim = 40mm 105mm 40mm 105mm, clip, width=.75\columnwidth,keepaspectratio]{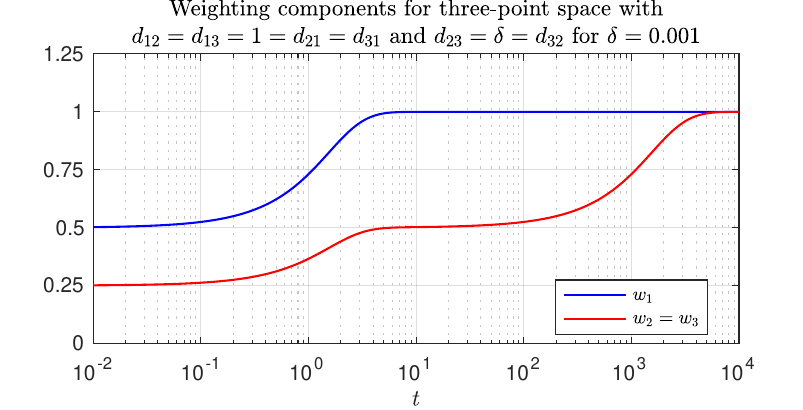}
  \caption{The magnitude $w_1+w_2+w_3$ of an isoceles dissimilarity space is a scale-dependent ``effective number of points.''}
  \label{fig:3pointSpace20210402}
\end{figure}
\end{example}

For a symmetric similarity matrix $Z$, the positive weighting of the submatrix on common row and column indices that has the largest magnitude is unique and proportional to the diversity-saturating distribution for \emph{all} values of the free parameter $q$ in \eqref{eq:diversity}, and this magnitude equals the maximum diversity. Because of the exponential number of subsets involved, in general the diversity-maximizing distribution is $\mathbf{NP}$-hard to compute, though cases of size $\le 25$ are easily handled on a laptop. 

The algorithmic situation improves radically if besides being symmetric, $Z$ is also positive definite
\footnote{
If $d$ is the distance matrix corresponding to (e.g.) a finite subset of Euclidean space, 
$Z = \exp[-td]$ is automatically positive definite: see Theorem 3.6 of \cite{meckes2013positive}.
}  
and admits a positive weighting $w$ (which is unique by positive-definiteness). Then
\begin{equation}
\label{eq:diversityMaximization}
\arg \max_p D_q^Z(p) = \frac{w}{1^T w},
\end{equation}
i.e., this weighting is proportional to the diversity-maximizing distribution, and linear algebra suffices to obtain it efficiently. 
\footnote{
In Euclidean space, the geometrical manifestation of diversity maximization via weightings is an excellent scale-dependent boundary/outlier detection mechanism \cite{willerton2009heuristic,bunch2020practical,huntsman2022diversity}. A technical explanation for this boundary-detecting behavior draws on the potential-theoretical notion of Bessel capacities \cite{meckes2015magnitude}. 
}

To engineer this situation, we take $Z = \exp[t_+d]$, where the \emph{strong cutoff} $t_+$ is the minimal value such that $\exp[-td]$ is positive semidefinite and admits a nonnegative weighting for any $t > t_+$. This is guaranteed to exist and can be computed using the bounds
\begin{equation}
\label{eq:diagonalCutoffBounds}
\frac{\log(n-1)}{\min_j \max_k d_{jk}} \le t_d \le \frac{\log(n-1)}{\min_j \min_{k \ne j} d_{jk}}
\end{equation}
where $t_d$ is the minimal value such that $\exp[-td]$ is diagonally dominant (i.e., $1 > \max_j \sum_{k \ne j} \exp(-t d_{jk})$) and hence also positive definite for any $t > t_d$ \cite{huntsman2022diversity}.

\section{\label{sec:GoExplore}Go-Explore}

The animating principle of Go-Explore is to ``first return, then explore''  \cite{ecoffet2021first}. This is a simple but powerful principle: by returning to previously visited states, Go-Explore avoids two pitfalls common to most sparse reinforcement learning algorithms such as \cite{guo2021geometric}. It does not prematurely avoid promising regions, nor does it avoid underexplored regions that have already been visited.

The basic scheme of Go-Explore is to repeatedly
\begin{itemize}
	\item probabilistically sample and {\bf go} to an elite state;
	\item {\bf explore} starting from the sampled elite;
	\item map resulting states to a cellular discretization of space;
	\item update the elites in populated cells.
\end{itemize}
This scheme has achieved breakthrough performance on outstanding challenge problems in reinforcement learning. Viewing it in the context of QD algorithms, 
we produce in Algorithm \ref{alg:GoExploreDissimilarity} a specific formal instantiation of Go-Explore for computationally expensive objectives on dissimilarity spaces while illustrating ideas and techniques that apply to more general settings and instantiations.

The basic setting of interest to us is a space $X$ endowed with a nondegenerate and symmetric (but not necessarily metric) dissimilarity $d : X^2 \rightarrow [0,\infty]$ and an objective $f : X \rightarrow \mathbb{R}$ for which we want a large number of diverse inputs $x$ (in the sense of \eqref{eq:diversity}) that each produce relatively small values of $f$. We  generally assume that $f$ is expensive to compute, so we want to evaluate it sparingly.

Although we assume the existence of a ``global generator'' that produces points in $X$, it does not need to explore $X$ globally: running our algorithm multiple times in succession (with elites from one run serving as ``landmarks'' in the next) addresses this. Similarly, although we assume the existence of a ``local generator'' in the form of a probability distribution, we do not need to know much about it--only that it be parametrized by a current base point $x \in X$ and some scalar parameter $\theta$, and that we can efficiently sample from it. 

\begin{figure}[h]
  \centering
  \includegraphics[trim = 80mm 110mm 77mm 105mm, clip, width=.3\columnwidth,keepaspectratio]{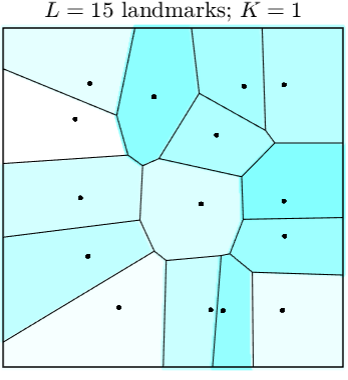}
  \includegraphics[trim = 80mm 110mm 77mm 105mm, clip, width=.3\columnwidth,keepaspectratio]{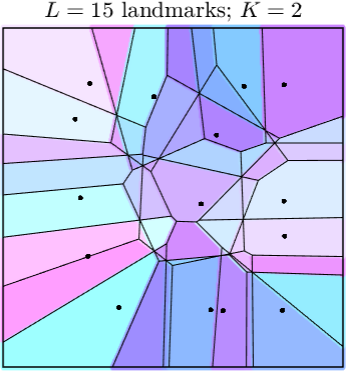}
  \includegraphics[trim = 80mm 110mm 77mm 105mm, clip, width=.3\columnwidth,keepaspectratio]{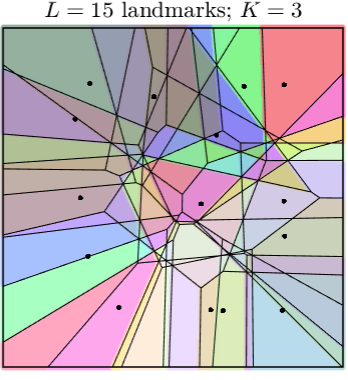}
\caption{(All axes are $[-3,4]^2$.) Algorithm \ref{alg:GoExploreCell} maps states to cells. $L = 15$ landmarks are shown as black dots ($\bullet$): these are iteratively selected from $T = \lceil L \log L \rceil = 41$ states with $G =$ sampling from $\mathcal{U}([-2,3]^2)$ as described in Algorithm \ref{alg:GoExploreLandmarks}. Landmarks are mostly dispersed  near the boundary of the sampling region. (Left; [resp., center; right]) Cells with $K = 1$ (resp., $K = 2; 3$) are all intersections of Voronoi cells of landmarks with $0$ (resp., $1; 2$) landmarks omitted. Many of the $L^K$ notional cells are degenerate.
  }
  \label{fig:cells20220609}
\end{figure}

A global generator suffices to provide a good discretization of a dissimilarity space along the lines of Algorithms \ref{alg:GoExploreLandmarks} and \ref{alg:GoExploreCell}, generalizing the approach of \cite{vassiliades2017using}. Magnitude satisfies an asymptotic (i.e., in the limit $t \uparrow \infty$) inclusion-exclusion formula for compact convex bodies in Euclidean space \cite{gimperlein2021magnitude}, and more generally appears to be approximately submodular.
\footnote{
$F : 2^\Omega \rightarrow \mathbb{R}$ is submodular iff $\forall X\subseteq \Omega$ and $x_1,x_2 \in \Omega \backslash X$ s.t. $x_{1}\neq x_{2}$, $F(X\cup \{x_1\})+F(X\cup \{x_2\})\geq F(X\cup \{x_1,x_2\})+F(X)$. Let $\Omega = \{(1,0),(0,1),(-1,0),(2,0)\}$ endowed with Euclidean distance; let $X = \{(1,0),(0,1)\}$; let $x_1 = (-1,0)$, and $x_2 = (2,0)$. Then (using an obvious notation) $\text{Mag}(X\cup \{x_1\})+\text{Mag}(X\cup \{x_2\})\approx 4.1773$ while $\text{Mag}(X\cup \{x_1,x_2\})+\text{Mag}(X) \approx 4.1815$, so magnitude is not submodular. 
} 
This suggests using the standard greedy approach for approximate submodular maximization \cite{nemhauser1978analysis,krause2014submodular} despite the fact that we do not have any theoretical guarantees. In practice this works well: by greedily maximizing the magnitude of a fixed-size subset of states, Algorithm \ref{alg:GoExploreLandmarks} produces a set of diverse landmarks, as Figure \ref{fig:cells20220609} shows. Ranking the dissimilarities of these landmarks to a query point yields a good locality-sensitive hash \cite{indyk1998approximate,gionis1999similarity,amato2008approximate,chavez2008effective,novak2010locality,tellez2010locality,kang2012robust,silva2014large,leskovec2020mining}.
\footnote{
A dissimilarity on the symmetric group $S_n$ can be used to query: see \S B.1 (supplement).
} See Figure \ref{fig:cells20220609}.


Although we generally assume an ability to produce a computationally cheap surrogate for $f$ and the availability of parallel resources for evaluating $f$ in much the same manner as \cite{gaier2018data,kent2020bop,zhang2022dsa}, most of our approach can be adapted to cases where either of these assumptions do not hold: see \S E of the supplement.

An archetypal class of example objectives is of the form $f = \phi \circ \gamma$, where $\gamma : X \rightarrow Y$ and $\phi : Y \rightarrow \mathbb{R}$ respectively embody a computationally expensive ``genotype to phenotype'' simulation and a fitness function such as $\phi(y) = \min_{\omega \in \Omega} d_Y(y,\omega)$ for some finite $\Omega \subset Y$ (e.g., representing desired outcomes of different tasks) and with $d(x,x') := d_Y(\gamma(x),\gamma(x'))$ a pullback distance.

This degree of generality is useful. In applications $X$ and/or $Y$ might be a pseudomanifold or a discrete structure like a graph
or space of variable-length sequences, so that $d$ is not (effectively anything like) Euclidean. Indeed, a now-classical result (phenomenologically familiar to users of multidimensional scaling) is that finite metrics (to say nothing of more general dissimilarities \emph{per se}) are generally not even embeddable in Euclidean space \cite{bourgain1985lipschitz,linial1995geometry}. 

A class of problems where $Y$ is a graph arises in automatic scenario generation \cite{auguston2005environment,martin2009automatic,nguyen2015innovation,li2020scenario,fontaine2021evaluating} where simulations involve a flow graph or finite automaton of some sort, e.g. testing and controlling machine learning, cyber-physical, and/or robotic systems. 
\footnote{
This is related to fuzzing \cite{bohme2017directed,manes2019art,zeller2019fuzzing,zhu2022fuzzing}, but with an emphasis on phenomenology rather than finding bugs, and a different computational complexity regime.  
}

Another attractive class of potential applications, in which $X$ is a space of variable-length sequences, is the design of diverse proteins and/or chemicals in high-throughput virtual screening \cite{gomez2016design}. In pharmacological settings, a quantitative structure–activity relationship is an archetypal objective \cite{roy2015primer}. For example, consider the problem of protein design in the context of an mRNA vaccine \cite{pardi2018mrna} where diversity could aid in the development of universal vaccines that protect against a diverse set of viral strains \cite{giuliani2006universal,paules2018chasing,koff2021universal,morens2022universal}. Proteins are conveniently represented using nucleic/amino acid sequences, and sequence alignment and related techniques give relevant dissimilarities \cite{rosenberg2009sequence,huntsman2015bruijn}. Using AlphaFold \cite{jumper2021highly,varadi2022alphafold} for protein structures and complementary deep learning techniques such as \cite{mcnutt2021gnina}, it is possible to estimate molecular docking results in about 30 seconds. Meanwhile, more accurate and precise results from absolute binding free energy calculations require about a day with a single GPU \cite{cournia2020rigorous}. 
\footnote{
For chemical drug design, a convenient representation is the SMILES language \cite{weininger1988smiles}, and relevant dissimilarities include \cite{ozturk2016comparative,samanta2020vae}. 
Here, a chemical drug-target interaction objective can be estimated using deep learning approaches such as \cite{gao2018interpretable,shin2019self,karimi2019deepaffinity}. 
}
Our approach appears to offer favorable use of parallel resources in realistic applications of this sort \cite{jayachandran2006parallelized}.

In still other applications, $Y$ might be a latent space of the sort produced by an autoencoder or generative adversarial network \cite{gaier2020discovering,fontaine2021illuminating}. Such applications are not as directly targeted by our instantiated algorithm since in practice such spaces are treatable as locally Euclidean (though only as pseudomanifolds \emph{versus} manifolds \emph{per se} due to singularities) and more specialized instantiations might be suitable. By the same token, we basically ignore the ``behavioral'' focus of many QD algorithms: in our intended applications, the local (and frequently global) generators already operate on reasonably low-dimensional feature (or latent) spaces, or the dissimilarity is pulled back from such a space. That said, engineering landmarks rather than using Algorithm \ref{alg:GoExploreLandmarks} can produce cells that follow a grid pattern and/or enable internal behavioral representations.

\begin{algorithm}
  \caption{\textsc{GoExploreDissimilarity}$(f,d,L,T,K,G,M,g,\mu)$}
  \label{alg:GoExploreDissimilarity}
\begin{algorithmic}[1]
  \REQUIRE Objective $f : X \rightarrow \mathbb{R}$, dissimilarity $d : X^2 \rightarrow [0,\infty]$, number of landmarks $L$, number of initial states $T \ge L$, rank cutoff $K \ll L$, global generator $G : \mathbb{N} \rightarrow X$, evaluation budget $M$, local generator probability distributions $g(x'|x,\theta)$, and maximum ``per-expedition'' exploration effort $\mu$ 
  \STATE Generate landmarks $\subset$ initial states $X'$ using $G$ \hfill \emph{// Algorithm \ref{alg:GoExploreLandmarks}}
  \STATE Evaluate $f$ on $X'$
  \STATE Evaluate $\sigma^{(K)}$ on $X'$ and initialize history $h$ \hfill \emph{// Algorithm \ref{alg:GoExploreCell}}
  \WHILE{$|h|<M$}
   \STATE $E \leftarrow \bigcup_\tau \{ \arg \min_{x \in h: \sigma^{(K)}(x) = \tau} f(x) \}$ \hfill \emph{// Elites}
   \STATE Compute weighting $w$ at scale $t_+$ on $E$
   \STATE Form $p \propto \exp([\log w]_\land - [f|_E]_\land)$ \hfill \emph{// \S \ref{sec:go}} 
   \STATE Compute best feasible lower bound $b \le \mathbb{E}_p(C_{\lceil |E|/2 \rceil})$ \hfill \emph{// \S C}
    \FOR{$b$ steps}
      \STATE Sample $x \sim p$ 
      \STATE Compute exploration effort $\mu_*\in [\mu]$ \hfill \emph{// \S \ref{sec:explorationEffort}}
      \STATE $U \leftarrow \text{set of the} \min \{|h|,\lceil \mu/2 \rceil \}$ states in $h$ nearest to $x$
      \STATE $V \leftarrow \{y \in h: \sigma^{(K)}(y) = \sigma^{(K)}(x)\}$
      \STATE Form local estimate $\hat f$ of $f$ using data on $U \cup V$ \hfill \emph{// \S \ref{sec:objectiveEstimate}}
      \STATE $\theta \leftarrow \max_{y \in E} d(x,y)$ \hfill \emph{// \S \ref{sec:bandwidth}}
      \STATE Sample $(x'_j)_{j=1}^{2\mu} \sim g^{\times 2\mu}(\cdot | x,\theta)$
      \WHILE{$|\{x'_j : \sigma^{(K)}(x'_j) = \sigma^{(K)}(x) \} | < 2\mu/4$}
        \STATE $\theta \leftarrow \theta/2$; sample $(x'_j)_{j=1}^{2\mu} \sim g^{\times 2\mu}(\cdot | x,\theta)$  
      \ENDWHILE
      \STATE Get weighting $w'$ at scale $t'_+$ on $\Lambda := \{x'_j\}_{j=1}^{2\mu} \cup U \cup V$
      \STATE Normalize $\hat f|_\Lambda$ and $-w'$ to zero mean and unit variance
      \STATE Compute Pareto domination of $(\hat f|_\Lambda,-w')$ \hfill \emph{// \S \ref{sec:pareto}}
      \STATE $X' \leftarrow \min \{\mu_*,M-|h| \}$ least dominated pairs 
    \ENDFOR
    \STATE Evaluate $f$ and $\sigma^{(K)}$ on $X'$; update $h$
  \ENDWHILE
  \ENSURE $h$ (and $E$ of globally diverse/locally optimal elites as above)
\end{algorithmic}
\end{algorithm}


\begin{algorithm}
  \caption{\textsc{GenerateDiverseLandmarks}$(d,L,T,G)$}
  \label{alg:GoExploreLandmarks}
\begin{algorithmic}[1]
  \REQUIRE Dissimilarity $d : X^2 \rightarrow [0,\infty]$, number of landmarks $L$, number of state generations $T \ge L$, global state generator $G : \mathbb{N} \rightarrow X$ that is one-to-one on $[T]$
  \FOR{$i$ from 1 to $L$}
    \STATE $x_i \leftarrow G(i)$ \hfill \emph{// State}
    \STATE $\mathcal{I}(i) \leftarrow i$ \hfill \emph{// Landmark index}
  \ENDFOR
  \STATE $D \leftarrow d|_{\{x_{\mathcal{I}(i)} : i \in [L] \}}$ \hfill \emph{// Dissimilarity matrix}
  \STATE $t \leftarrow t_+(D)$ \hfill \emph{// Strong cutoff}
  \STATE $w \leftarrow \exp[-t D] \backslash 1$ \hfill \emph{// Weighting}
  \STATE $M \leftarrow 1^T w$ \hfill \emph{// Magnitude}
  \FOR{$i$ from $L+1$ to $T$}
    \STATE $i' \leftarrow \arg \min w$ \hfill \emph{// State with least weighting component}
    \STATE $x_i\leftarrow G(i)$ \hfill \emph{// Candidate state}
    \STATE $\mathcal{I}' \leftarrow \mathcal{I}$ \hfill \emph{// Candidate landmark indices}
    \STATE $\mathcal{I}'(i') \leftarrow i$ \hfill \emph{// Try current index for landmark}
    \STATE $D' \leftarrow d|_{\{x_{\mathcal{I}'(i)} : i \in [L] \}}$ \hfill \emph{// Candidate dissimilarity matrix} 
    \STATE $w' \leftarrow \exp[-t D'] \backslash 1$ \hfill \emph{// Candidate weighting}
    \STATE $M' \leftarrow 1^T w'$ \hfill \emph{// Candidate magnitude}
    \IF{$M' > M$}
      \STATE $\mathcal{I} \leftarrow \mathcal{I}'$, $M \leftarrow M'$ \hfill \emph{// Accept candidate} 
    \ENDIF
  \ENDFOR
  \ENSURE Initial states $\{x_i\}_{i = 1}^{T}$, $L$ landmark indices in $\mathcal{I} \subseteq [T]$
\end{algorithmic}
\end{algorithm}

    %
    %
    
\begin{figure}[h]
  \centering
  \includegraphics[trim = 40mm 75mm 40mm 75mm, clip, width=.5\columnwidth,keepaspectratio]{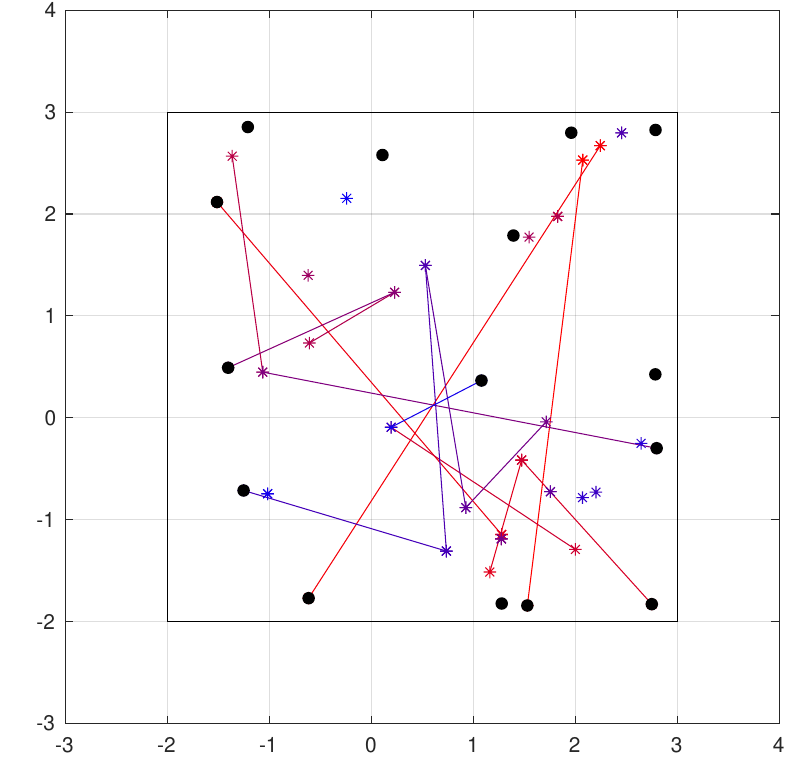}
\caption{Algorithm \ref{alg:GoExploreLandmarks} generates $L = 15$ landmarks shown as black dots ($\bullet$) among $T = \lceil L \log L \rceil = 41$ states with $G =$ sampling from $\mathcal{U}([-2,3]^2)$. Non-landmark states are shown as asterisks ($*$) colored from {\color{red}red} to {\color{blue}blue} in order of generation. Replacements of prior landmarks are indicated with lines in the same color scheme. At the end, landmarks are dispersed.
  }
  \label{fig:landmarks20220609}
\end{figure}


\begin{algorithm}
  \caption{$\textsc{StateCell}(d,\{x_i\}_{i = 1}^{T}, \mathcal{I},K,x)$}
  \label{alg:GoExploreCell}
\begin{algorithmic}[1]
  \REQUIRE Dissimilarity $d : X^2 \rightarrow [0,\infty]$, states $\{x_i\}_{i = 1}^{T}$, landmark indices $\mathcal{I}$ with $L := |\mathcal{I}| \le T$, rank cutoff $K \le L$, state $x \in X$ 
  \STATE Sort to get $\sigma$ such that $d(x,x_{\mathcal{I}(\sigma(1))}) \le \dots \le d(x,x_{\mathcal{I}(\sigma(L))})$ 
  \STATE $\sigma^{(K)}(x) \leftarrow (\sigma(1), \dots, \sigma(K))$ \hfill \emph{// $K$ closest landmarks (sorted)}
  \ENSURE Cell identifier $\sigma^{(K)}(x)$
\end{algorithmic}
\end{algorithm}

\section{\label{sec:go}Going}

A distribution over cells drives the ``go'' process. We balance exploration of the space $X$ and exploitation of the objective $f$, respectively via a diversity-saturating probability distribution \eqref{eq:diversityMaximization} and a distribution proportional to $\exp(-\beta f)$. Rather than taking $\beta$ as a variable regularizer, we use it to remove a degree of freedom. Define
\begin{equation}
\label{eq:quantileNormalization}
[\phi]_\land := \frac{\phi-\text{median }\phi}{\text{max }\phi-\text{median }\phi}
\end{equation}
and the ``go distribution'' 
\begin{equation}
\label{eq:goDistribution}
p \propto \exp([\log w]_\land - [f|_E]_\land),
\end{equation} 
where the weighting $w$ is computed for the set $E$ of elites at scale $t_+$.
\footnote{
The weighting $w$ at scale $t_+$ (and \eqref{eq:diversityMaximization}) has one entry equal to zero by construction. Therefore the corresponding entry of $\log w$ is $-\infty$ and we cannot get finite minima or moments, which motivates the normalization \eqref{eq:quantileNormalization}.
}
The distribution $p$ encourages visits to cells whose elites contribute to diversity and/or have the lowest objective values. As elites improve, $p$ changes over time, reinforcing the ``go'' process.

We want to sample from $p$ to go to enough elites for exploration, but not so often as to be impractical. Meanwhile, to mitigate bias, we sample over the course of discrete epochs. As \S C of the supplement details and Algorithm \ref{alg:GoExploreDissimilarity} requires, we efficiently compute a good lower bound on the expected time for the event $C_m$ of visiting $m$ of $n$ elites via IID draws from the distribution $p \equiv (p_1,\dots,p_n)$. This lower bound is the number of iterations in the inner loop of Algorithm \ref{alg:GoExploreDissimilarity}.

\section{\label{sec:explore}Exploring}

In general, the entire exploration mechanism in Algorithm \ref{alg:GoExploreDissimilarity} should be tailored to the problem under consideration. Nevertheless, for the sake of striking a balance between specificity and generality we outline elements of an exploration mechanism that works ``out of the box'' on a variety of problems, as demonstrated in \S \ref{sec:examples}.

\subsection{\label{sec:explorationEffort}Exploration Effort}

If the last two visits to a cell have (resp., have not) improved its elite, we spend more (resp., less) effort exploring on the next visit. Let $f_-$ denote the elite objective from the penultimate visit, and $f_0$ the elite objective from the last visit. Let $\mu_-$ and $\mu_0$ be the corresponding efforts (i.e., number of exploration steps). If we assume $f$ is (empirically) normalized to take values in the unit interval, then $-1 \le -(f_0 - f_-) \le 1$. At the lower end, $f_0 - f_- = 1$, which is the worst; at the upper end, $f_0 - f_- = -1$, which is the best. We therefore assign exploration effort 
\begin{equation}
\label{eq:effort}
\mu_+ = \lceil \min \{ \max \{1, 2^{-(f_0 - f_-)} \mu_0 \}, \mu \} \rceil,
\end{equation} 
where $\mu \in \mathbb{N}$ is the maximum exploration effort per ``expedition.'' 

If and when $\mu_+$ is unity for long enough across cells, it makes sense to halt Algorithm \ref{alg:GoExploreDissimilarity} before reaching $M$ function evaluations: however, we do not presently do this.

\subsection{\label{sec:objectiveEstimate}Estimating the Objective Function}

It is generally quite useful to leverage an estimate or \emph{surrogate} $\hat f$ to optimize an objective $f$ whose evaluation is computationally expensive. It is possible to estimate objectives in very general contexts: e.g., for $X$ a (di)graph endowed with a symmetric dissimilarity $d$, via graph learning and/or signal processing techniques \cite{xia2021graph}.

In the more pedestrian and common settings $X \in \{\mathbb{R}^n, \mathbb{Z}^n, \mathbb{F}_2^n\}$, a straightforward approach to estimation is furnished by radial basis function interpolation \cite{buhmann2003radial} using linear (or more generally, polyharmonic) basis functions that require no parameters. Our code is built with this particular approach in mind.

\subsection{\label{sec:bandwidth}Bandwidth for Random Candidates}

The scalar \emph{bandwidth} parameter $\theta$ in the ``local generator'' $g(x'|x,\theta)$ corresponds to, e.g., the standard deviation in a spherical Gaussian for $X \in \{\mathbb{R}^N, \mathbb{Z}^N\}$, 
\footnote{
Perhaps surprisingly, sampling from the discrete Gaussian on $X = \mathbb{Z}^N$ is very hard
\cite{aggarwal2015solving}. Even producing a discrete Gaussian with specified moments is highly nontrivial
\cite{agostini2019discrete}. For these reasons we prefer the simple expedient of rounding (away from zero) samples from a continuous Gaussian.
}
or the parameter in a Bernoulli trial. The idea is that $\theta$ governs the locality properties of $g$, with $\theta$ large (resp., small) approximating a uniform (resp., point) distribution on $X$. Note that if we have \emph{any} mechanism for locally perturbing $x$, then repeating this $\theta$ times allows us to introduce a notion of bandwidth.

Now we want to be able to set $\theta$ so that we actually do explore, but only locally (for otherwise we might as well just evaluate $f$ on points produced by $G$), i.e., we want to explore within and possibly just next to a cell. There is a simple way to do this provided that everything except evaluating $f$ is fast and efficient (which we assume throughout). The idea is to sample from $g(\cdot | x,\theta)$ with a default (large) value for $\theta$ to generate  a large number of ``probes'' widely distributed in $X$. We then decrease $\theta$ and sample again until a significant plurality of probes are in the same cell as the elite $x$.

Initially, we set $\theta = \max_{y \in E} d(x,y)$. Although this makes sense for $g$ a Gaussian and still generally works for bit flips (albeit at the minor cost of probing until $\theta < 1$), some care should be exercised to ensure the definition of $g$ and the initialization of $\theta$ are both compatible with the problem instance at hand.

\subsubsection{\label{sec:generatorAdaptation}Generator Adaptation}

An alternative approach is to estimate the optimal parameter (here not necessarily scalar) of a local generator $g$. It should often be possible to borrow from CMA-ES \cite{hansen2016cma} by taking the best performing probes and using them to update the covariance of (e.g.) a Gaussian. We forego this here in the interest of generality, though it is likely to be very effective \cite{fontaine2020covariance}.

\subsection{\label{sec:pareto}Pareto Dominance}

We assume that we can evaluate $f$ in parallel and produce an estimate/surrogate $\hat f$ that is inexpensive to evaluate. 
It is advantageous to evaluate $f$ at points where we expect it to be more optimal based on $\hat f$, as well as at points that increase potential diversity (i.e., a weighting component) relative to prior evaluation points.

Given competing objectives $\alpha_i$ on a set $\Lambda$, we say the \emph{Pareto domination} of $\lambda \in \Lambda$ is $\max_{\lambda' \in \Lambda} \min_i [\alpha_i(\lambda) - \alpha_i(\lambda')]$. In our algorithm, we take $\Lambda$ to be the union of the set of probes with a set $U$ of nearby states in the evaluation history and the set $V$ of states in the evaluation history that belong to the same cell as the current elite $x$. We take $\alpha_1 = [\hat f]_\|$ and $\alpha_2 = [-w']_\|$, where $w'$ is the weighting on $\Lambda$ at scale $t_+$ and $[\cdot]_\|$ indicates a normalization to zero mean and unit variance. The points that are least Pareto dominated form a reasonable trade space between exploration and evaluation.

\section{\label{sec:examples}Examples}

\subsection{\label{sec:real}$X = \mathbb{R}^N$}

\begin{figure}[h]
  \centering
  \includegraphics[trim = 35mm 70mm 35mm 75mm, clip, width=.49\columnwidth,keepaspectratio]{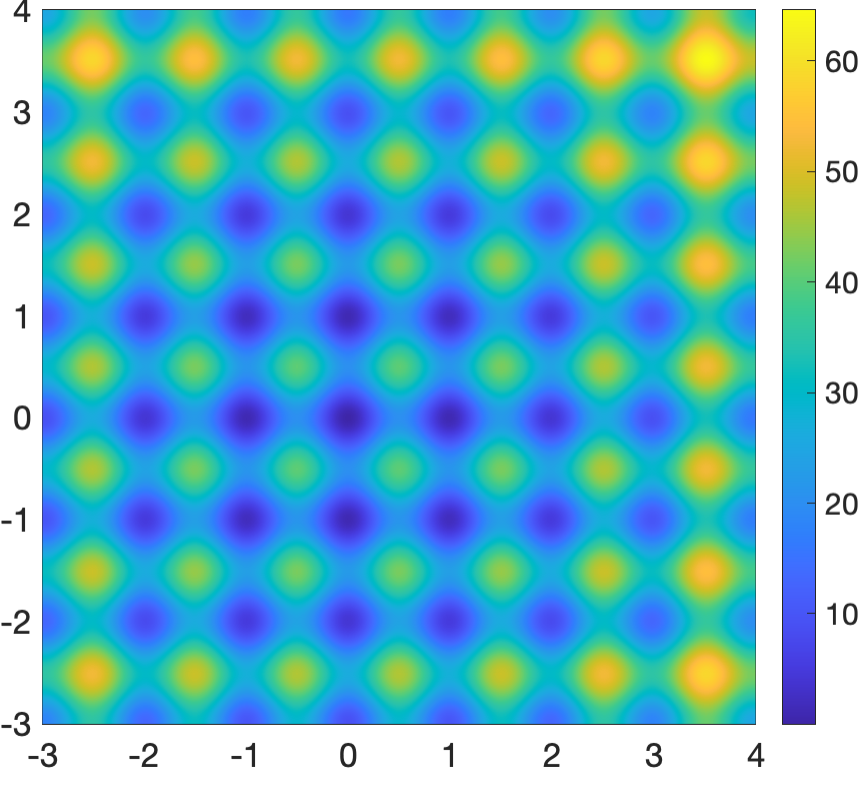}
  \includegraphics[trim = 35mm 70mm 35mm 75mm, clip, width=.49\columnwidth,keepaspectratio]{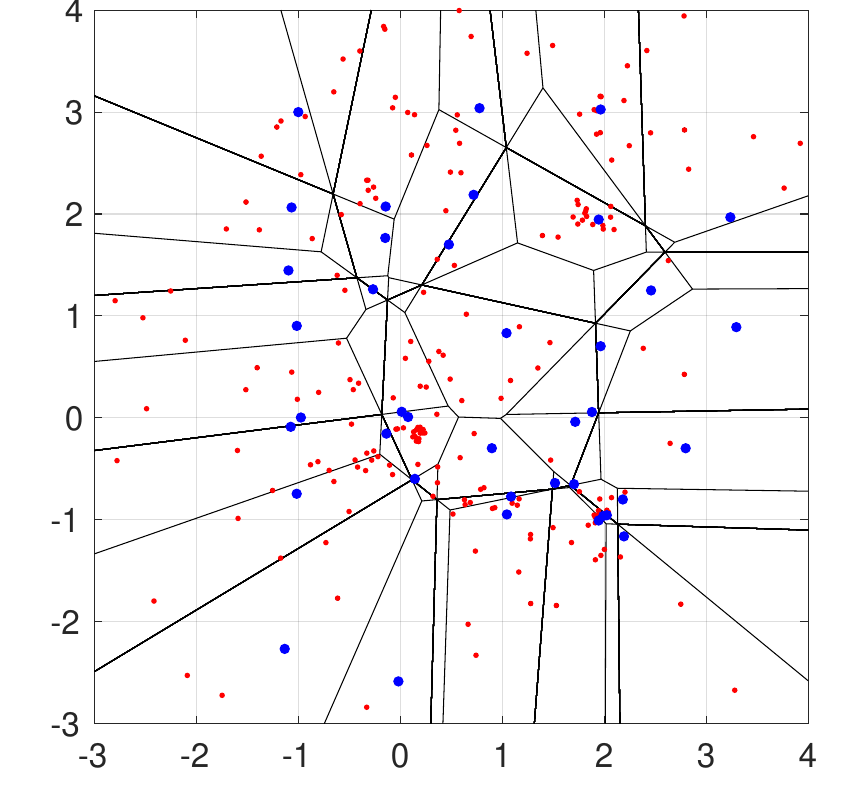}
\caption{(Left) Plot of \eqref{eq:Rastrigin} on $[-3,4]^2$. (Right) Output of Algorithm \ref{alg:GoExploreDissimilarity} for $M = 300$ evaluations of \eqref{eq:Rastrigin} for $N = 2$ with $L = 15$, $T = \lceil L \log L \rceil = 41$, and $K = 2$. To break symmetry, $G =$ sampling from $\mathcal{U}([-2,3]^N)$. Landmarks/cells are as in Figure \ref{fig:cells20220609}. $g(\cdot |x,\theta) = $ sampling from $\mathcal{N}(x,\theta^2 I)$; maximum per-expedition exploration effort $\mu = 128$. Evaluated points are shown as {\color{red}small red dots ($\cdot$)}; elites as {\color{blue}larger blue dots ($\bullet$)}.
  }
  \label{fig:Rastrigin20221212}
\end{figure}

\begin{figure}[h]
  \centering
  \includegraphics[trim = 35mm 70mm 35mm 75mm, clip, width=.49\columnwidth,keepaspectratio]{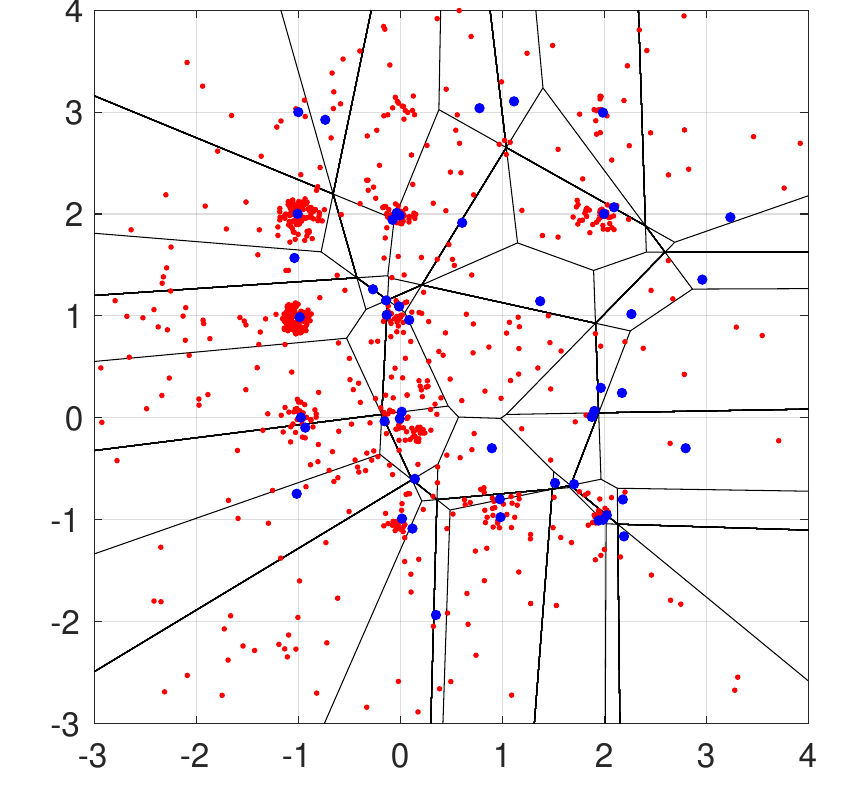}
  \includegraphics[trim = 35mm 70mm 35mm 75mm, clip, width=.49\columnwidth,keepaspectratio]{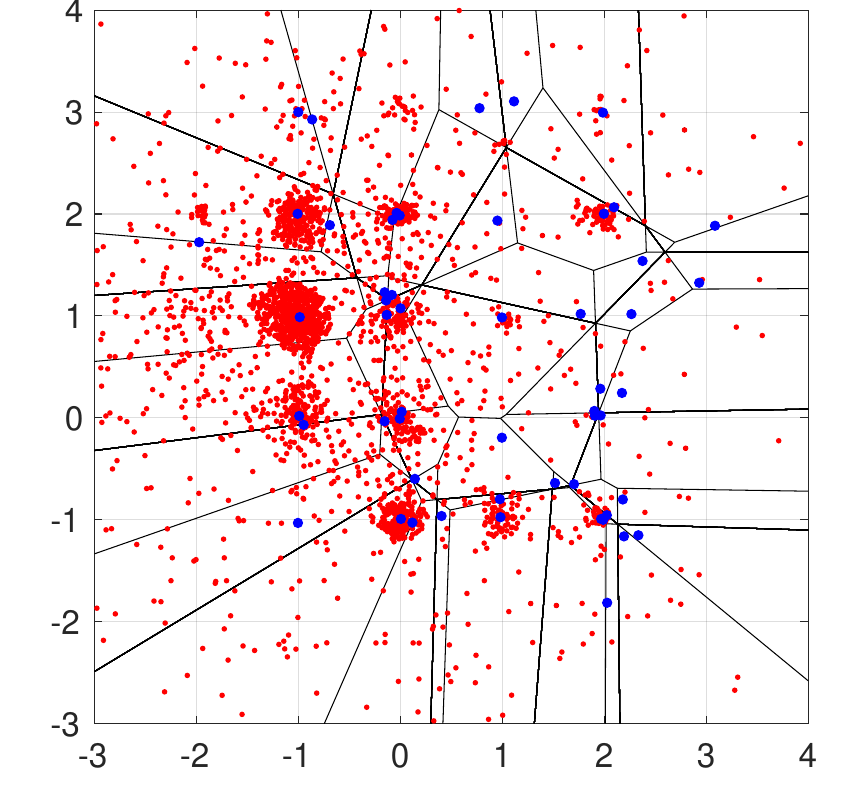}
\caption{(Left) As in the right panel of Figure \ref{fig:Rastrigin20221212}, but for $M = 1000$. The elite near $(1,1)$ moved near $(1,0)$. (Right) $M = 3000$.
  }
  \label{fig:Rastrigin_3000_20221212}
\end{figure}

We consider the Rastrigin function (see left panel of Figure \ref{fig:Rastrigin20221212})
\begin{equation}
\label{eq:Rastrigin}
f(x) = A \cdot N+\sum _{j=1}^N \left ( x_j^2-A\cos(2\pi x_j) \right )
\end{equation}
on $\mathbb{R}^N$ with the usual choice $A = 10$. This has a single global minimum at the origin and local minima on $\mathbb{Z}^N$. It (or a variant thereof) is commonly used as a test objective for QD \cite{cazenille2019comparing,cully2021multi} as well as global optimization problems. Figure \ref{fig:Rastrigin20221212} (right) and Figure \ref{fig:Rastrigin_3000_20221212} show the output of Algorithm \ref{alg:GoExploreDissimilarity} for varying evaluation budgets.

\subsubsection{\label{sec:benchmarkingReal}Benchmarking}

Because the cells produced by Algorithm \ref{alg:GoExploreLandmarks} are irregular, it is difficult to compare our approach to most QD algorithms, especially using a QD score \cite{pugh2016quality}, which is the special case $WQD(1,f)$ of
\begin{equation}
\label{eq:WQD}
WQD(w,\hat f) := |\text{supp } w| \cdot \frac{ w^T \hat f }{w^T 1},
\end{equation}
where $w$ is typically a solution of \eqref{eq:weighting}.
\footnote{
The usual QD score $WQD(1,f)$ typically approximates \eqref{eq:WQD} well in practice.
}
There are two possible workarounds: substitute regular cells, or consider a more ``lightweight'' version of the same approach. While the former option is superficially attractive in that we can compare to a notional reference algorithm, in practice there are only reference implementations of otherwise partially specified algorithms. Aligning a reference algorithm with Algorithm \ref{alg:GoExploreLandmarks} would force us to consider low-dimensional problems that are comparatively disadvantageous as discussed shortly below. Since we already know Go-Explore is a high-performing class of algorithms, the latter option therefore is more informative (and though it does not preclude the former, we restrict consideration to it here). 

\begin{figure}[h]
  \centering
  \includegraphics[trim = 64mm 103mm 65mm 102mm, clip, width=.49\columnwidth,keepaspectratio]{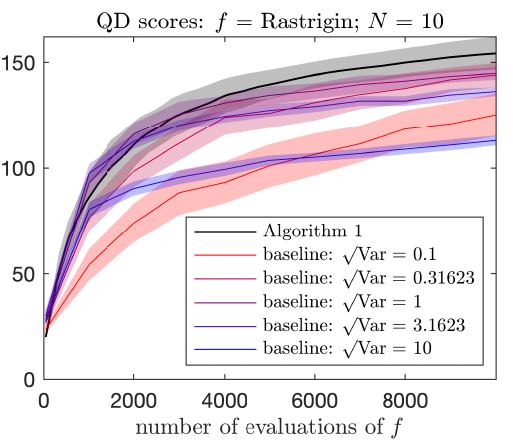}
  \includegraphics[trim = 64mm 103mm 65mm 102mm, clip, width=.49\columnwidth,keepaspectratio]{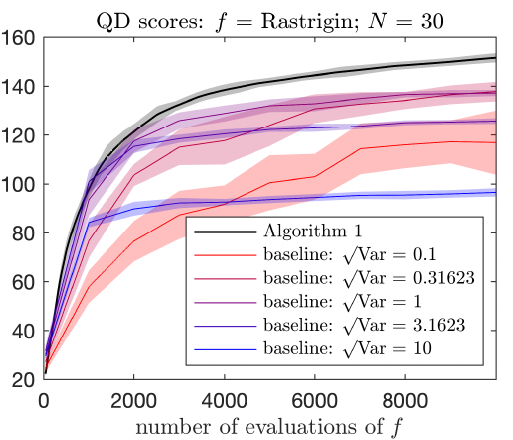}
\caption{QD scores for Algorithm \ref{alg:GoExploreDissimilarity} and a ``baseline'' Go-Explore variant for varying bandwidths applied to the objective \eqref{eq:Rastrigin} (normalized to be in $[0,1]$) in dimensions (left) $N = 10$ and (right) $N = 30$. Results are averaged with standard deviations indicated. (We elected not to normalize by the nominal number of possible cells.) 
}
  \label{fig:RastriginScore_midDim}
\end{figure}

Specifically, we consider a ``baseline'' version of Go-Explore with several specific and simple choices as described in \S F of the supplement: note in particular that the chosen covariances for $g$ therein are fairly well matched to the geometry of the problem.
\footnote{
The bandwidth for which the resulting Gaussian most nearly approximates the objective at the origin (up to an affine transformation) is approximately 0.2.
}
The results of this and Algorithm \ref{alg:GoExploreDissimilarity} with the same settings as in \S \ref{sec:real} are shown in Figure \ref{fig:RastriginScore_midDim}. There is a significant advantage for Algorithm \ref{alg:GoExploreDissimilarity} in dimensions $N \in \{10,30\}$. For dimensions 3 and 100, the advantage lessens (not shown). In the former case, this is due to the low dimensionality that outweighs any marginal gains due to more precise exploitation of local minima. In the latter case, the problem becomes sufficiently difficult and the evaluation budget low enough that the advantages of Algorithm \ref{alg:GoExploreDissimilarity} cannot yet manifest.

\subsection{\label{sec:integer}$X = \mathbb{Z}^N$}

Only minor changes are required to demonstrate Algorithm \ref{alg:GoExploreDissimilarity} on a nontrivial problem on a discrete lattice. Here we take a scale parameter $\lambda \in \mathbb{Z}_+$ and consider \eqref{eq:Rastrigin} on $\mathbb{Z}^2$ under the substitution $x \leftarrow x/\lambda$, while also scaling the domain in the same way. This has minima at $(\lambda \mathbb{Z})^2$, with the global minimum at the origin as before. By using the same pseudorandom number generator seed (which we normally do anyway for reproducibility), we can see a very similar variant of \S \ref{sec:real} emerge: see \S G of the supplement.

\subsection{\label{sec:binary}$X = \mathbb{F}_2^N$}

For binary problems, we embed $\mathbb{F}_2$ in $\mathbb{Z}$ via $0 \mapsto 0$ and $1 \mapsto 1$. 

\begin{figure}[h]
  \centering
  \includegraphics[trim = 35mm 80mm 35mm 80mm, clip, width=.75\columnwidth,keepaspectratio]{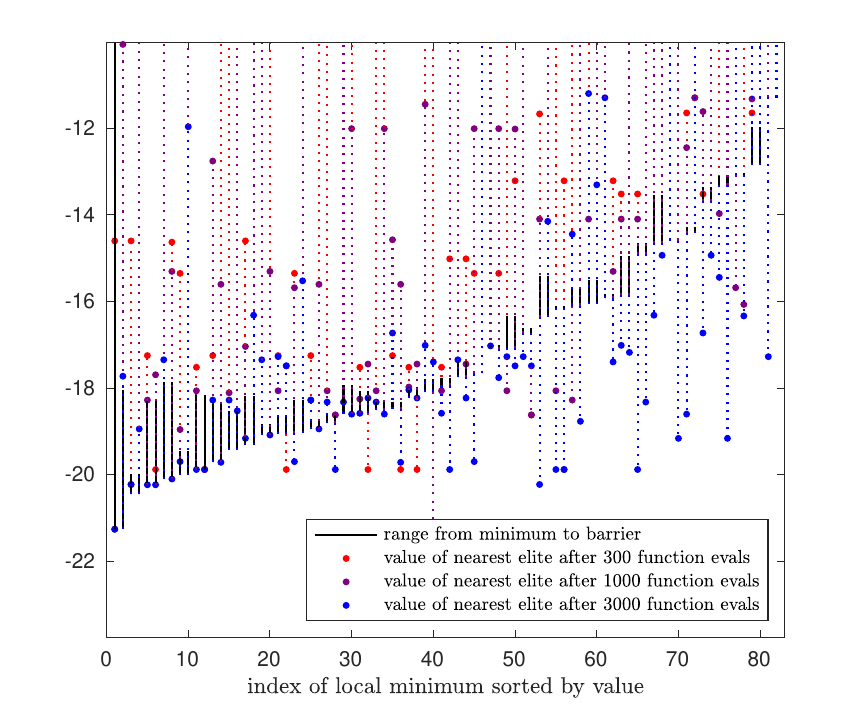}
\caption{Performance of Algorithm \ref{alg:GoExploreDissimilarity} on \eqref{eq:SK} with $N = 20$, $d(s,s') := d_H^{1/2}(b,b')$ for $d_H =$ Hamming distance and $b := (s+1)/2$, $L = 10$, $T = \lceil L \log L \rceil = 24$, $K = 2$, $G =$ sampling from $\mathcal{U}(\mathbb{F}_2^N)$, $g$ a Bernoulli bit flipper, and $\mu = 128$. There are $2^N \approx 10^6$ possible bit/spin configurations, but after just 3000 evaluations of \eqref{eq:SK}, many of the lowest minima have apparently had their barriers traversed \emph{en route}. After 300, 1000, and 3000 evaluations, there are respectively 57, 75, and 85 elites (of $L^K = 100$ nominally possible): meanwhile, there are 82 local minima. Many minima have nearest elites with lower values because (e.g.) the minima are in unexplored cells or because a cell contains more than one minimum. 
}
  \label{fig:barriersSK20221212}
\end{figure}

The Sherrington-Kirkpatrick (SK) spin glass objective is \cite{bolthausen2007spin,panchenko2012sherrington}
\begin{equation}
\label{eq:SK}
\textstyle f(s) = \frac{1}{\sqrt{N}} \sum_{jk} J_{jk} s_j s_k,
\end{equation}
where $s \in \{\pm 1\}^N$ is a spin configuration and $J$ is a symmetric $N \times N$ matrix with IID $\mathcal{N}(0,1)$ entries. It is a classic result that optimizing \eqref{eq:SK} is $\mathbf{NP}$-hard, although there is an $O(N^2)$ algorithm that approximates the optimum with high probability \cite{montanari2021optimization}. 
\footnote{
NB. For $N$ large, the software described in \cite{perera2020chook} can be used to generate hard optimization problems similar to \eqref{eq:SK} with known minima.
}
The underlying phenomenology is that (SK and more general) spin glasses have barrier trees 
\footnote{
The barrier tree of a function is a tree whose leaves and internal vertices repsectively correspond to local minima and minimal saddles connecting minima \cite{becker1997topology}. Barrier trees of spin glasses can be efficiently computed using the \texttt{barriers} program described in \cite{flamm2002barrier} and available at \url{https://www.tbi.univie.ac.at/RNA/Barriers/}.
}
that exhibit fractal and hierarchical characteristics \cite{fontanari2002fractal,zhou2009energy}. These same adjectives describe the landscape of \eqref{eq:SK}, explaining why a near-optimum can be rapidly identified (i.e., via a coarse approximation of the landscape) yet an exact optimum still requires exponential time to identify in general. Figure \ref{fig:barriersSK20221212} shows the performance of Algorithm \ref{alg:GoExploreDissimilarity} on an instance of \eqref{eq:SK} with $N = 20$.

\subsection{\label{sec:maze}$X = $ Nondecreasing Bijections on $[0,1]$}

As a penultimate example, we consider a maze-like problem in which we start with a suitable function $f_0 : [0,1]^2 \rightarrow \mathbb{R}$ and subsequently consider the line integral objective
\begin{equation}
\label{eq:lineIntegral}
f(\gamma) := \int_\gamma f_0
\end{equation}
for $\gamma$ a nondecreasing path from $(0,0)$ to $(1,1)$. This problem has several interesting and attractive features:
\begin{itemize}
\item it is infinite-dimensional, and any useful discretization is either very high-dimensional or such that the notion of dimension itself is inapplicable to the space $X$ of paths;
\item it is maze-like, with multiple local minima, opportunities for ``deception,'' straightforward visualization, etc.;
\item we can use an estimate of $f_0$ to estimate $f$ in turn;
\item the global optimum can be efficiently approximated by computing the shortest path through a weighted DAG (or by dynamic programming \emph{per se}).
\end{itemize}

\begin{figure}[h]
  \centering
  \includegraphics[trim = 65mm 105mm 65mm 100mm, clip, width=.49\columnwidth,keepaspectratio]{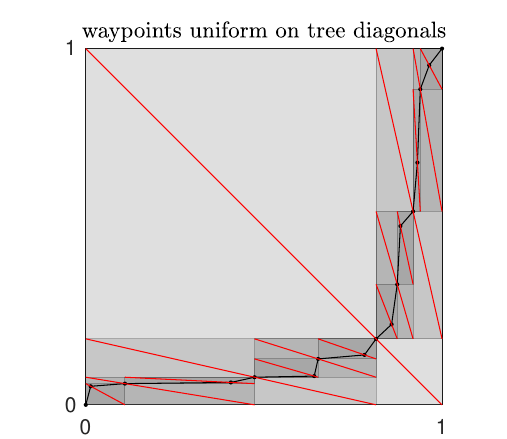}
  \includegraphics[trim = 65mm 105mm 65mm 100mm, clip, width=.49\columnwidth,keepaspectratio]{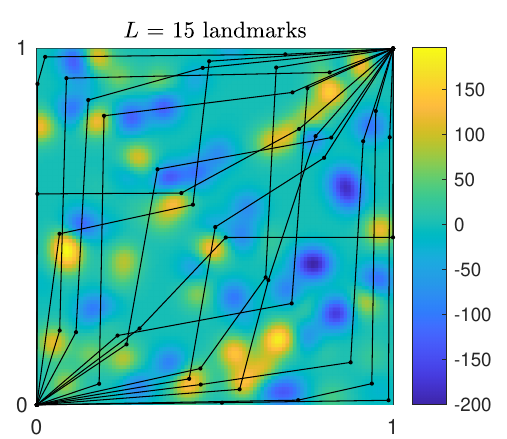}
\caption{(Left) A path connecting a binary tree of waypoints that are uniformly sampled on diagonals. (Right) Landmarks superimposed on a plot of $f_0$.
  }
  \label{fig:landmarksL2}
\end{figure}

A suitable global generator $G$ requires a bit of engineering: uniformly random lattice paths in $(\lambda \mathbb{Z})^2$ from the origin to $(1,1)$ are easy to construct, but are exponentially concentrated along the diagonal joining the two endpoints, which obstructs exploration. Instead, we take the approach indicated in the left panel of Figure \ref{fig:landmarksL2}, uniformly sampling a point on the diagonal between $(0,1)$ and $(1,0)$, then recursively uniformly sampling points on diagonals of rectangles induced by parent points in a binary tree structure. The right panel of Figure \ref{fig:landmarksL2} shows $L = 15$ landmarks obtained this way with a tree depth of 2, $T = \lceil L \log L \rceil = 41$, and the usual $L^2$ distance.

The design of a suitable local generator $g$ is also not entirely trivial: it is necessary to preserve the nondecreasing property of paths. For our instantiation, if a path has $n$ waypoints (including the endpoints), then we perform the following procedure $n' := \lceil (n-2) \theta \rceil$ times: i) we sample from a PDF on the set $\{insert, move, delete\}$ that depends on $n'$; ii) according to the sample, we either insert a new waypoint in a rectangle with adjacent waypoints as corners that is subsequently scaled by $\theta$; delete a waypoint if tenable; or move a waypoint, again in a rectangle with corners determined by adjacent waypoints and subsequently scaled by $\theta$. 

Finally, we take $f_0$ as a weighted sum of $100$ Gaussians with uniformly random centers and covariance $10^{-3} I$; the weights are $\sim \mathcal{U}([-1,1])$. We take $K = 2$, $M = 10000$, and $\mu = 128$. In line with the spirit of our developments, we also estimate \eqref{eq:lineIntegral} by performing a linear RBF interpolation of $f_0$ using waypoints in paths near the current one. However, we also evaluate $f_0$ on $\approx 100$ points along paths to produce a reasonably accurate value for the objective \eqref{eq:lineIntegral}.

\begin{figure}[h]
  \centering
  \includegraphics[trim = 65mm 105mm 65mm 100mm, clip, width=.49\columnwidth,keepaspectratio]{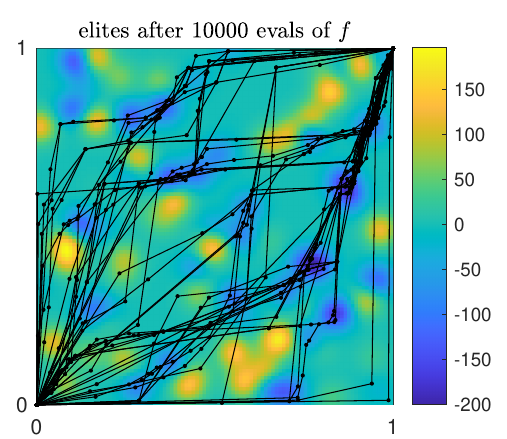}
  \includegraphics[trim = 65mm 105mm 65mm 100mm, clip, width=.49\columnwidth,keepaspectratio]{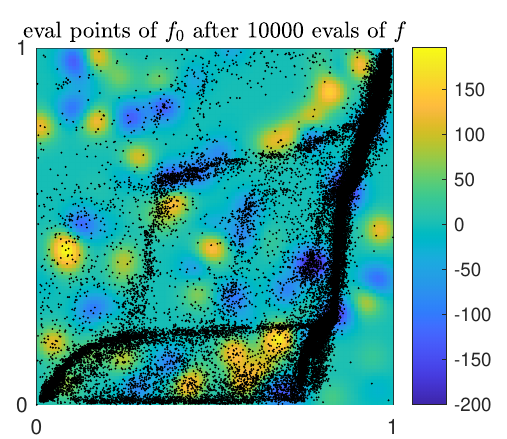}
\caption{(Left) Elites. (Right) Waypoints on evaluated paths. 
  }
  \label{fig:eliteL2}
\end{figure}

\begin{figure}[h]
  \centering
  \includegraphics[trim = 65mm 105mm 65mm 100mm, clip, width=.49\columnwidth,keepaspectratio]{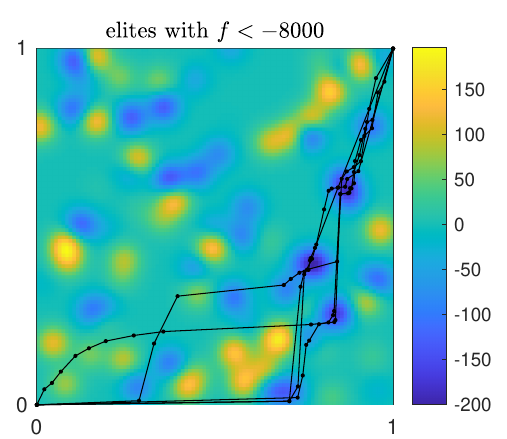}
  \includegraphics[trim = 65mm 105mm 65mm 100mm, clip, width=.49\columnwidth,keepaspectratio]{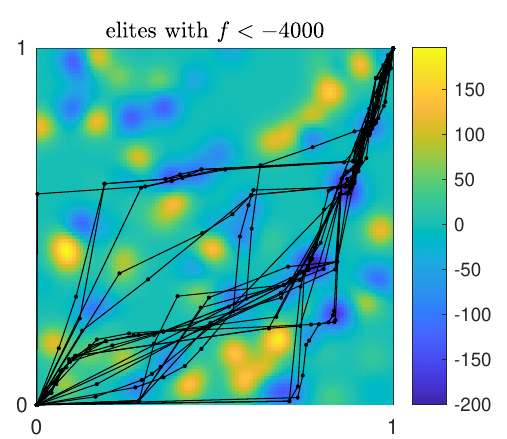}
\caption{As in the right panel of Figure \ref{fig:eliteL2}, but for elites with $f < -8000$ (resp., $-4000$) on the left (resp., right).
  }
  \label{fig:eliteLevel24L2}
\end{figure}

\begin{figure}[h]
  \centering
  \includegraphics[trim = 65mm 105mm 65mm 100mm, clip, width=.49\columnwidth,keepaspectratio]{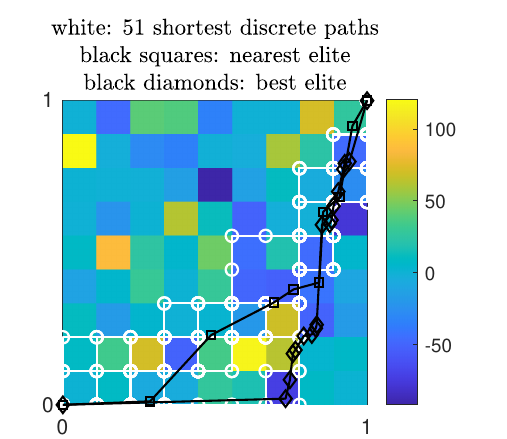}
  \includegraphics[trim = 65mm 105mm 65mm 100mm, clip, width=.49\columnwidth,keepaspectratio]{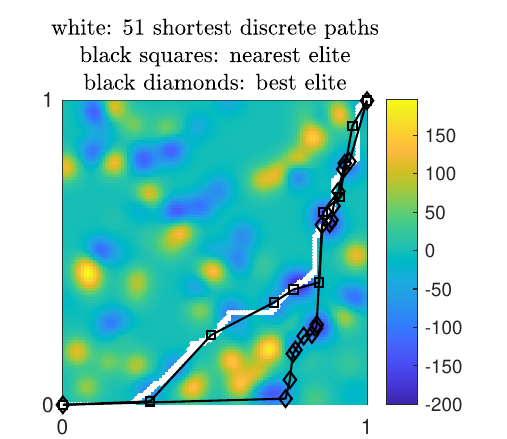}
\caption{(Left) The 46 (= number of elites) shortest paths (white) through a discrete approximation with $\lambda^{-1} = 10$ subdivisions per dimension and the elites that are closest to the shortest path (black squares) and that are best performing (black diamonds). (Right) As on the left, but with $\lambda^{-1} = 100$.
  }
  \label{fig:discreteL2}
\end{figure}

The resulting elites are shown in the left panel of Figure \ref{fig:eliteL2}; the right panel shows all of the waypoints considered along the way. Higher-performing elites are shown in Figure \ref{fig:eliteLevel24L2}. We can see that elites hierarchically coalesce and diverge. Figure \ref{fig:discreteL2} shows the 46 (= number of elites) shortest paths through weighted DAGs with edges from $\lambda \cdot (j,k)$ to $\lambda \cdot (j+1,k)$ and $\lambda \cdot (j,k+1)$ for $\lambda \in \{1/10,1/100\}$ and weights given by $\lambda \cdot f_0(\textnormal{edge midpoint})$; also shown is the closest elite to the shortest path and the elite with the best value of $f$. Note that although there is an elite that approximates the shortest paths well, none of these paths are close to the high-performing ``upper branch'' paths shown in Figure \ref{fig:eliteLevel24L2}. 

%

\subsection{\label{sec:CFG}$X = S^{N-1}$}

As a final example to further demonstrate the versatility of our approach, we construct and fuzz (i.e., we aim to comprehensively evaluate behavior on the basis of randomly generated inputs to) \cite{bohme2017directed,manes2019art,zeller2019fuzzing,zhu2022fuzzing} 100 toy programs that move a unit vector on a high-dimensional sphere $S^{N-1} \subset \mathbb{R}^N$ according to the following procedure: 

First, in each case we generate a program ``skeleton'' using 300 productions from the probabilistic context free grammar \cite{visnevski2007syntactic}
\begin{equation}
\label{eq:skeleton}
\texttt{S} \rightarrow \texttt{S; S} \ | \ \texttt{if b; S; fi} \ | \ \texttt{while b; S; end}
\end{equation}
where $\texttt{S}$ is shorthand for a line separator: the production probabilities are respectively $0.6$, $0.1$, and $0.3$. The tokens $\texttt{S}$ and $\texttt{b}$ respectively represent statements/subroutines and Boolean predicates. 

Second, we form the resulting \emph{control flow graph} (CFG) \cite{cooper2011engineering} by associating vertices with lines in the skeleton and edges according to Table \ref{tab:CFG}.
\begin{table}
  \caption{CFG edge: $[\cdot] := $ line number of matching token.}
  \label{tab:CFG}
  \begin{tabular}{ccl}
    \toprule
    source at line $j$& target($\top$)&target($\bot$)\\
    \midrule
    \texttt{if b} & $j+1$& [\texttt{fi}]+1\\
    \texttt{while b} & $j+1$& [\texttt{end}]+1\\
    \texttt{end} & [\texttt{while}]& -\\
    \texttt{fi} or \texttt{S} & $j+1$& -\\
  \bottomrule
\end{tabular}
\end{table}
Next, we fully instantiate a program by i) replacing a token \texttt{S} on line $j$ of the program by the assignment $x \leftarrow S_j x$, where $S_j$ is a orthogonal matrix of dimension $N = 10$ sampled uniformly at random \cite{stewart1980efficient}; 
and ii) replacing a token \texttt{b} on line $k$ of the program by the predicate $b_k^T x > 0$, where $b_k \sim \mathcal{U}(S^{N-1})$.

Third, we literally draw the CFG using MATLAB's \texttt{layered} option and define an objective function to be the depth--literally, the least vertical coordinate in the drawing--reached when traversing $2 \max d_{CFG}(\texttt{START},\cdot)$ edges in the CFG by dynamically executing the program, where here the program entry and asymmetric digraph distance on the CFG are indicated. The other inputs for the algorithm are mostly familiar from other examples above: $d =$ distance on $S^{N-1}$ (we actually get slightly better results with the ambient Euclidean distance [not shown]); $L = 15$; $T = \lceil L \log L \rceil = 41$; $K = 2$; $G =$ sampling from $\mathcal{U}(S^{N-1})$; $g(\cdot | x,\theta)$ the unit-length normalization of sampling from $\mathcal{N}(x,\theta^2 I)$; $\mu = 128$; and $M = 1000$. Again, we do this for 100 different programs. 

\begin{figure}[h]
  \centering
  \includegraphics[trim = 45mm 95mm 45mm 90mm, clip, width=.49\columnwidth,keepaspectratio]{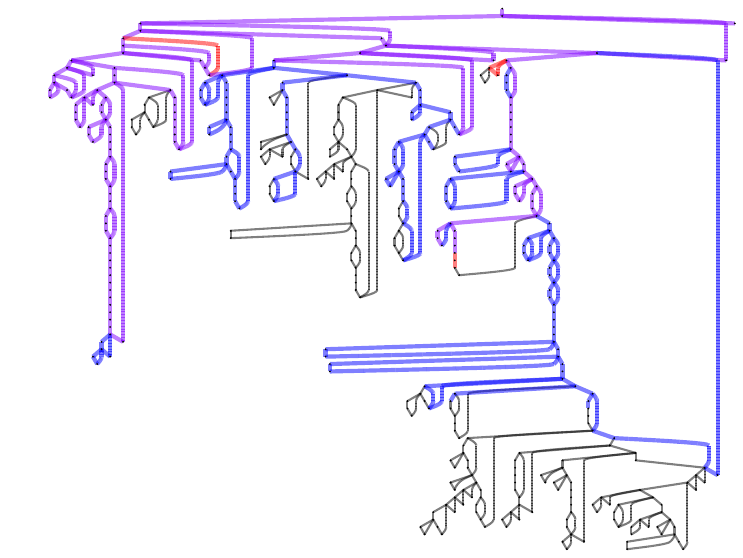}
  \includegraphics[trim = 45mm 95mm 45mm 90mm, clip, width=.49\columnwidth,keepaspectratio]{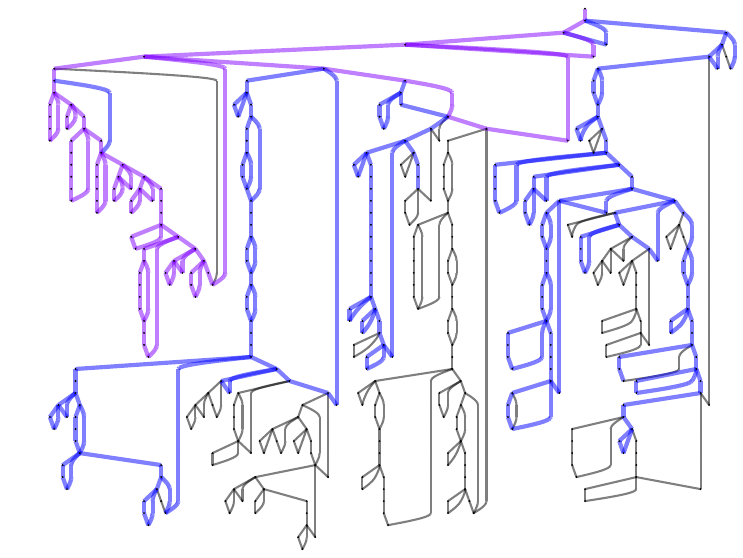}
\caption{(Left) Coverage for the first of 100 CFGs by Algorithm \ref{alg:GoExploreDissimilarity} as described in the text, \emph{versus} by repeatedly executing the global generator $G$ for $M = 1000$ times. {\color{blue}Edges exercised only by Algorithm \ref{alg:GoExploreDissimilarity} are blue}; {\color{red}edges exercised only by $G$ are red}; and {\color{violet}edges exercised by both are purple}.
(Right) As in the left panel, but for the last of 100 CFGs.
  }
  \label{fig:CFG_coverage}
\end{figure}

\begin{figure}[htbp]
  \centering
  \includegraphics[trim = 50mm 100mm 55mm 95mm, clip, width=.49\columnwidth,keepaspectratio]{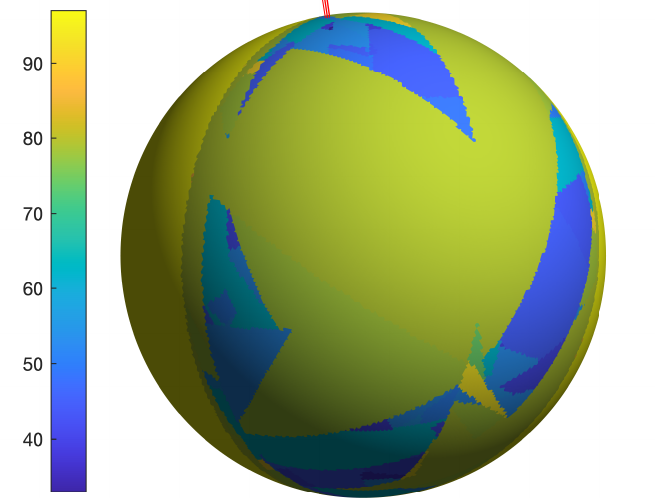}
  \includegraphics[trim = 60mm 103mm 65mm 100mm, clip, width=.49\columnwidth,keepaspectratio]{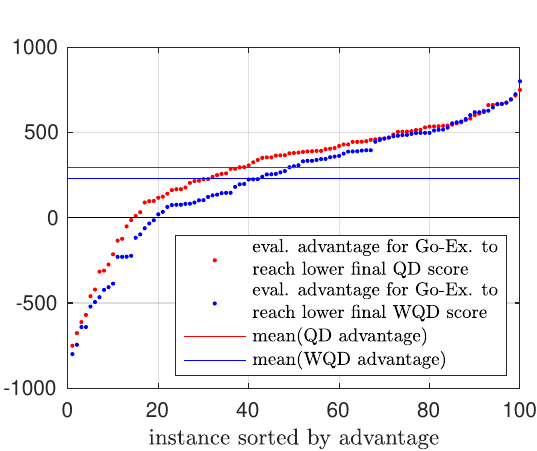}
\caption{(Left) An objective on $S^2$ obtained along the same lines as described in \S \ref{sec:CFG} for $S^9$, and evaluated on vertices of a geodesic polyhedron with 163842 vertices. The grid optimum is attained on just eight vertices near the top, indicated with a {\color{red}red} marker. (Right) Evaluation advantage of Algorithm \ref{alg:GoExploreDissimilarity} relative to the global generator $G$ to reach the lower (W)QD-score achieved by either over the course of $M=1000$ evaluations. Note that even though about 15-20 percent of instances are disadvantageous, overall there is still a 21-27 percent overall advantage for Algorithm \ref{alg:GoExploreDissimilarity}. This advantage is significant in light of the objective's behavior.}
  \label{fig:CFG_advantage}
\end{figure}

In Figure \ref{fig:CFG_coverage} we show two of the 100 examples of ``code coverage'' using Algorithm \ref{alg:GoExploreDissimilarity} \emph{versus} $G$ alone. Although on occasion $G$ produces better coverage by itself, this is comparatively uncommon, as the right panel of Figure \ref{fig:CFG_advantage} illustrates. It is worth noting that the objective here presents some fundamental difficulties for optimization: by construction, it is piecewise constant on intricate regions (see left panel of Figure \ref{fig:CFG_advantage}). From this perspective, the mere existence of a clear advantage is significant. 

\section{\label{sec:zeroScale}``Extremal'' diversity at scale zero}

In the limit $t \uparrow \infty$, a weighting tends to the vector $1$; this and consideration of examples such as in Figure \ref{fig:3pointSpace20210402} suggest that the opposite limit $t \downarrow 0$ encodes ``extremal'' information about diversity. Figure \ref{fig:maxQE} shows examples of weightings that maximize diversity at $t = 0$ and compares one of these to the corresponding weighting at $t = t_+$. It is clear that the former dramatically singles out ``boundary'' points of a sort. 
In fact for Euclidean examples, modulo a change of metric $d \mapsto \sqrt{d}$ and re-embedding into Euclidean space, \footnote{
See Proposition 4.12 of \cite{devriendt2022graph}.
}
such weightings actually single out points lying on a minimal enclosing sphere (see below).

An obvious question is whether or not this extremal notion of diversity can provide an advantage when incorporated into Algorithm \ref{alg:GoExploreDissimilarity}. However, any possibility of a positive answer to this question requires a reasonably efficient algorithm for actually computing the diversity at $t = 0$. We pursue this first, focusing on the case $q = 1$ as it corresponds most closely to Shannon entropy and turns out to admit an elegant algorithm (recall that the distribution maximizing diversity does not depend on $q$ in the end).

For $p$ in the simplex $\Delta_{n-1} := \{ p \in [0,1]^n : 1^T p = 1 \}$ of nonnegative probability distributions, the diversity of order $1$ is
\begin{equation}
\label{eq:diversity1}
D_1^{Z}(p) := \prod_{j:p_j > 0} (Zp)_j^{-p_j}
\end{equation}
and the corresponding similarity-sensitive generalization of Shannon entropy is
\begin{equation}
\label{eq:ssEntropy1}
\log D_1^{Z}(p) = - \sum_{j:p_j > 0} p_j \log (Zp)_j.
\end{equation}
We would like to compute the maximum value of \eqref{eq:ssEntropy1} for the common case $Z = \exp[-td]$ in the limit $t \downarrow 0$ for a generic invertible dissimilarity matrix $d$. While this goal turns out to be overly ambitious in practice, we can still manage fairly well.

The first-order approximation $Z = \exp[-td] \approx 11^T - td$ yields
\begin{equation}
\label{eq:ssEntropy1Approx}
\log D_1^{Z}(p) \approx t p^T d p.
\end{equation}
The term $p^T d p$ in the right hand side of \eqref{eq:ssEntropy1Approx} is the \emph{quadratic entropy} \cite{rao1982diversity} and the problem of maximizing it over $\Delta_{n-1}$ has been considered in, e.g., \cite{hjorth1998finite,izsak2002quadratic,pavoine2005measuring,izumino2006maximization}. In the Euclidean setting, \cite{pavoine2005measuring} points out that this maximum quadratic entropy is realized by the squared radius of a minimal sphere containing points with distance matrix $\sqrt{d}$ (which is also a Euclidean distance matrix); the support of $p$ corresponds to the subset of points on this sphere.

It can be shown (see, e.g., Example 5.16 of \cite{devriendt2022graph}) that maximizing $p^T d' p$ over $\Delta_{n-1}$ is $\mathbf{NP}$-hard for arbitrary $d'$. This is not surprising in light of the fact that quadratic programming is generally $\mathbf{NP}$-hard \cite{sahni1974computationally} and remains so even when the underlying matrix has only a single eigenvalue with a given sign \cite{pardalos1991quadratic}. While this sign condition is typical for Euclidean distance matrices \cite{schoenberg1937certain,bogomolny2007distance}, it nevertheless turns out that for $d$ a Euclidean distance matrix, $p^T d p$ is convex (see Theorem 4.3 of \cite{rao1984convexity} and Proposition 5.20 of \cite{devriendt2022graph}), so it can be efficiently maximized over any sufficiently simple polytope via quadratic programming. More generally, $p^T d p$ is convex if $d$ is \emph{strict negative type}, i.e., $x^T d x < 0$ for $1^T x = 0$: this entails that $Z$ is positive semidefinite for \emph{all} $t>0$. 
\footnote{
See also \cite{leinster2012measuring,meckes2013positive,leinster2016maximizing,leinster2021entropy}. Per Lemma 1.7 of \cite{parthasarathy1972positive} as rephrased in Theorem 4.2 of \cite{rao1984convexity}, we can encode the truth or falsity of the assertion that a symmetric nonnegative matrix $d$ of size $n$ is negative type (i.e., $x^T d x \le 0$ for $1^T x = 0$) in one line of MATLAB, viz. \texttt{eigs(d(2:n,2:n)-d(2:n,1)-d(1,2:n)+d(1,1),1,'smallestabs')<0}. 
}
In \S \ref{sec:maximizingQE} we exhibit a more practical algorithm than quadratic programming for maximizing the quadratic entropy of strict negative type metrics.

On the other hand, it appears likely that maximizing $p^T d p$ over $\Delta_{n-1}$ is still generally $\mathbf{NP}$-hard when $Z$ is positive definite only for all sufficiently small $t$. Yet in this intermediate case we can still do better than despairing at general intractability or resorting to an approach described below as a heuristic of uncertain effectiveness by developing a nontrivial bound. By Theorem 3.2 of \cite{izumino2006maximization} we have that 
$$\arg \max_{p \in \mathbb{R}^n : 1^T p = 1} p^T d p = \frac{d^{-1}1}{1^T d^{-1} 1},$$
though in general this extremum (which also turns out to equal the limiting weighting $\lim_{t \downarrow 0} Z^{-1} 1$) will have negative components. Thus the best practical recourse when $Z$ is positive definite only for all sufficiently small $t$ is to bound $p^T d p$ using
\begin{equation}
\label{eq:bound1}
\max_{p \in \Delta_{n-1}} p^T d p \le \max_{p \in \mathbb{R}^n : 1^T p = 1} p^T d p = \frac{1}{1^T d^{-1} 1}.
\end{equation}
This unpacks as
\begin{align}
\label{eq:bound2}
\lim_{t \downarrow 0} \frac{\log D_1^Z(p)}{\max_{p \in \Delta_{n-1}} \log D_1^Z(p)} & \ge \lim_{t \downarrow 0} \frac{\log D_1^Z(p)}{\max_{p \in \mathbb{R}^n : 1^T p = 1} \log D_1^Z(p)} \nonumber \\
& = (p^T d p) \cdot (1^T d^{-1} 1).
\end{align}

\subsection{\label{sec:maximizingQE}Maximizing quadratic entropy of strict negative type metrics}

Translated into our context, Proposition 5.20 of \cite{devriendt2022graph} states that if $d$ is strict negative type (again, recall this includes Euclidean distance matrices), then 
\begin{equation}
\label{eq:argMaxQE}
p_*(d) := \arg \max_{p \in \Delta_{n-1}} p^T d p
\end{equation}
is uniquely characterized by the conditions
\begin{itemize}
	\item[i)] $p_*(d) \in \Delta_{n-1}$
	\item[ii)] $e_j^T d p \ge e_k^T d p$ for all $j \in \text{supp}(p)$ and $k \in [n]$.
\end{itemize}

A generally efficient algorithm with good error bounds and an anytime convergence guarantee, and whose first stage is an excellent approximation that always runs in polytime, is described in \S 3.1 and detailed in \S G of [Huntsman, S. ``Peeling metric spaces of strict negative type.'' TAG-DS (2025); also at arXiv:2509.19565; we do not cite this in references because this is not a published version of record]. It addresses the $t \downarrow 0$ limit of an algorithm described in the preprint \cite{huntsman2022diversity} but omitted from the published version of record. See also Theorem 5.23 of \cite{devriendt2022graph}.

\subsection{\label{sec:zeroScaleExamples}Practicalities and examples}

As a practical matter, we have found that the approximate algorithm referenced in the preceding subsection also performs better than a quadratic programming solver: it is much faster (in MATLAB on $\approx 1000$ points, a few hundredths of a second versus several seconds for a quadratic programming solver with tolerance $10^{-10}$) and more accurate, in particular by handling sparsity exactly. Figure \ref{fig:maxQE} shows representative results. It is also very simple to implement.

\begin{figure}[h]
  \centering
  \includegraphics[trim = 65mm 110mm 65mm 100mm, clip, width=.49\columnwidth,keepaspectratio]{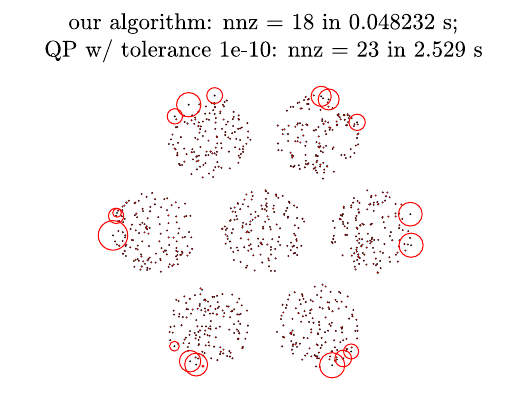}
  \includegraphics[trim = 65mm 110mm 65mm 100mm, clip, width=.49\columnwidth,keepaspectratio]{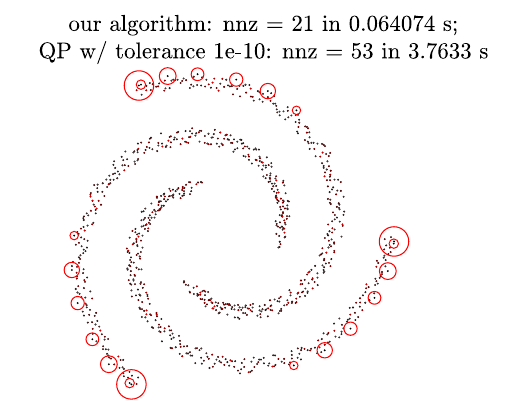}
  \includegraphics[trim = 65mm 109mm 65mm 109mm, clip, width=.49\columnwidth,keepaspectratio]{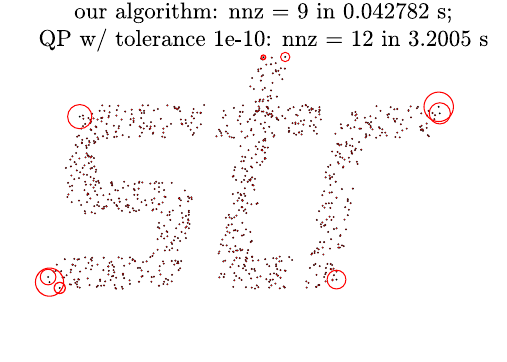}
  \includegraphics[trim = 65mm 109mm 65mm 109mm, clip, width=.49\columnwidth,keepaspectratio]{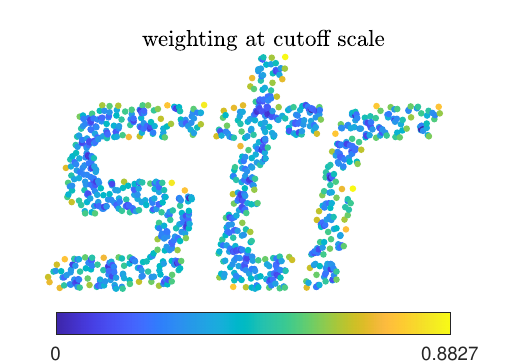}
\caption{(Upper left, upper right, and lower left) Outputs of the approximate algorithm on the Euclidean distance matrix of the $\approx 1000$ black points, indicated by {\color{red}red} circles with radius proportional to the corresponding entries of $p$. The numbers of nonzero (nnz) entries of the output are indicated along with the runtimes of the algorithm; the same numbers are reported for a quadratic programming run with tolerance $10^{-10}$. (Lower right) The weighting on the same points as in the lower left panel at the strong cutoff $t = t_+$. Recall that this scale is the least such that the weighting is nonnegative. Note that even at this scale, weighting components still tend to be at corners or at least boundaries.
  }
  \label{fig:maxQE}
\end{figure}

\subsection{\label{sec:zeroScaleQdAlgorithm}Incorporating diversity contributions at scale zero and/or a fuzzing-inspired exploration effort into Algorithm \ref{alg:GoExploreDissimilarity}}

There are two obvious places to incorporate scale zero computations into Algorithm \ref{alg:GoExploreDissimilarity}: on elites, and on probes/expeditions sent from elites. These respectively address the core ``go'' and ``explore'' mechanisms in our approach. We modified our original implementation to support both of these as options, along with the ``entropic power schedule'' of \cite{bohme2020boosting} as an option to govern the exploration effort. 

Perhaps surprisingly, applying these options to the Rastrigin function indicates that none of them has a positive effect, as Figure \ref{fig:maxQE} illustrates (we do not show the entropic power schedule variant as it was computed for fewer function evaluations and its poor performance was already evident). The poor performance of scale zero options appears to be because they respectively suppress exploration around ``interior'' elites and near elites generically for the cases in which the scale zero options are applied to elites and probes/expeditions. These factors in turn both impact the ability of corresponding variants of Algorithm \ref{alg:GoExploreDissimilarity} to find high-quality solutions.

\begin{figure}[h]
  \centering
  \includegraphics[trim = 64mm 103mm 65mm 102mm, clip, width=.49\columnwidth,keepaspectratio]{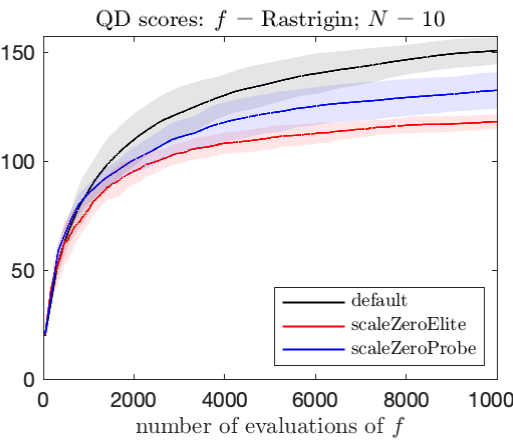}
  \includegraphics[trim = 64mm 103mm 65mm 102mm, clip, width=.49\columnwidth,keepaspectratio]{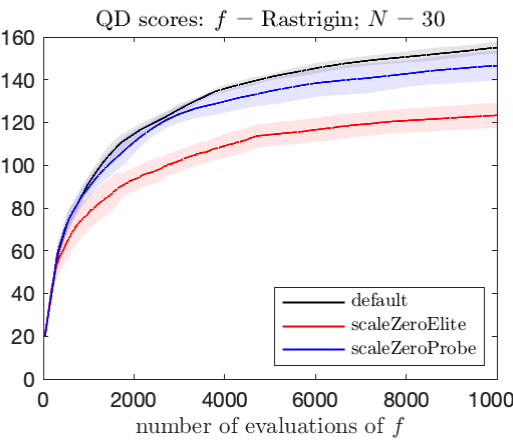}
\caption{(Left) QD-scores for variants of the approximate algorithm described above using scale zero constructions and applied to the Rastrigin function in dimension $N = 10$ (cf. Figure \ref{fig:RastriginScore_midDim}). The approximate algorithm, indicated as the default here, clearly outperforms its variants. (Right) As in the left panel, but for dimension $N = 30$.
  }
  \label{fig:maxQE}
\end{figure}

Our consideration of the entropic power schedule of \cite{bohme2020boosting} was inspired by our development of a cousin of Algorithm \ref{alg:GoExploreDissimilarity} for the fuzzing problem (see \S \ref{sec:CFG}), where we confirmed it offered leading performance in accordance with theoretical arguments behind it (details will be reported elsewhere). However, the underlying rationale of this construction is mismatched with typical considerations of QD algorithms \emph{per se} to a degree that was only obvious to us after doing an experiment. 

In the fuzzing problem, improving an input amounts to identifying an essentially new part of the behavior space. Doing the latter optimally is precisely the aim of the entropic power schedule. Meanwhile, the idea permeating much of the present paper of modeling an objective is fundamentally ill-posed in the context of fuzzing: once a program's behavior is reasonably well understood in a neighborhood of a given input, that input should be ignored in favor of others whose local behavior is not yet well understood. 

In short, while we believe a variant of Algorithm \ref{alg:GoExploreDissimilarity} can improve the state of the art for fuzzing, and ideas from both diversity optimization at scale zero and fuzzing and are relevant to quality-diversity, a naive transplantation of ideas is inadequate in either direction. Finally, combining these ideas will also require some combination of delicacy and/or approximation. If $d$ is strict negative type, we can resort to the approximate algorithm mentioned above as in applications of Go-Explore that seek to interpolate an objective on Euclidean space. However, for typical fuzzing applications, we must resort to the bound \eqref{eq:bound2} to govern a power schedule of our fuzzer at scale $t=0$.
\footnote{
Proposition 5.17 of \cite{devriendt2022graph} gives the generic bounds $p^T d p \in [\frac{1}{2},\frac{n-1}{n}] \cdot \max_{jk} d_{jk}$.
}

\section{\label{sec:conclusion}Conclusion}

Our formulation of Go-Explore illustrates that a single quality-diversity algorithm can be applied across very general settings. The algorithm centers on a mathematically principled definition of diversity that admits a tractable maximization scheme. The algorithm usefully separates concerns of problem details, surrogate construction, and its core quality-diversity mechanisms. 

Although we focused here on very structured input spaces in order to produce efficient surrogates for expensive objectives in the context of quality-diversity algorithms, the notions of diversity, magnitude, and weightings provide tools for building optimization algorithms more generally. In particular, the prospect of using neural surrogates and/or methods for accelerating the computation of weightings such as \cite{rouet2016distributed,chenhan2017n} suggest wider scope for applications. 

Finally, while the results of \S \ref{sec:zeroScale} do not yield improvements to Algorithm \ref{alg:GoExploreDissimilarity}, we believe they are of independent interest and can yield useful applications of their own.

\begin{acks}
Thanks to Zac Hoffman and Rachelle Horwitz-Martin for clarifying questions and suggestions, and to Karel Devriendt for an illuminating discussion about maximizing quadratic entropy. This research was developed with funding from the Defense Advanced Research Projects Agency (DARPA). The views, opinions and/or findings expressed are those of the author and should not be interpreted as representing the official views or policies of the Department of Defense or the U.S. Government. Distribution Statement “A” (Approved for Public Release, Distribution Unlimited).
\end{acks}

\bibliographystyle{ACM-Reference-Format}


\appendix

\section{\label{sec:OverviewOfAppendices}Overview of appendices}

\begin{itemize}
	\item \S \ref{sec:goAlternatives} sketches alternative ``go'' mechanisms.
	\item \S \ref{sec:couponCollection} details bounds governing ``how often to go.''
	\item \S \ref{sec:bandwidthNote} elaborates on bandwidth in $\mathbb{R}^N$.
	\item \S \ref{sec:exploreAlternatives} sketches alternative ``explore'' mechanisms. 
	\item \S \ref{sec:GoExploreBaseline} lists pseudocode for a baseline version of Go-Explore used in \S \ref{sec:benchmarkingReal}.
	\item \S \ref{sec:IntegerResults} shows additional results for \S \ref{sec:integer}.
	\item \S \ref{sec:regex} details another example over $\mathbb{Z}^N$ involving regular expression matching.
	\item \S \ref{sec:OtherBinaryExamples} elaborates on the binary example in the main text and details other examples.
	\item \S \ref{sec:BugFix} addresses the repair and consequent effects of a minor bug in \S \ref{sec:goExploreDissimiliarity.m}.
	\item \S \ref{sec:SourceCode} contains source code.
\end{itemize}

\section{\label{sec:goAlternatives}Alternative techniques for going}

Alternatives to a distribution on elites would be distributions on all evaluated states or on all cells. Because the number of evaluated states grows linearly, the cubic complexity of standard linear algebra on them can become untenable, so we do not elaborate on the former alternative here. 
\footnote{
Appendix E of \cite{huntsman2022parallel} discusses how to improve performance for \eqref{eq:weighting}: see also \S 4 of \cite{bunch2021weighting}.
}

\subsection{\label{sec:distributionsOnAllCells}Distributions on all cells}

A distribution on cells could aim to incorporate i) a weighting based on the underlying dissimilarity $d$ and ii) a global estimate of the objective $f$. Since \S \ref{sec:objectiveEstimate} largely addresses ii), we focus here on i).
\footnote{
Evaluating $f$ on every cell \emph{de novo} could also be done using a state generation technique described below that is applicable to many cases of practical interest. However, this is probably inappropriate for situations we are most concerned with.
}

An interesting idea is to use a distance on permutations to define a suitable probability distribution over all cells. Such distances can yield simple positive definite kernels that elegantly dovetail with the magnitude \cite{leinster2021entropy} framework. For example, if $d$ is the so-called \emph{Kendall tau/bubble sort} (resp., \emph{Cayley}) distance on $S_n$ that counts the number of transpositions of adjacent (resp., generic) elements needed to transform between two permutations, then the corresponding \emph{Mallows} (resp., \emph{Cayley}) kernel $\exp[-td]$ is positive definite for all $t \ge 0$ \cite{jiao2015kendall} (resp., for $t > \log(n-1)$) \cite{corfield2021fundamental}. The Kendall tau distance can be computed in quasilinear time \cite{chan2010counting} and the Cayley distance in linear time (by counting cycles in the quotient permutation).

However, there is not much point in pursuing this idea without a mechanism for generating states in cells on demand (necessary in general due to the curse of dimensionality). For $X = \mathbb{R}^N$ the state generation mechanism can be effected using linear programming, but for $X = \mathbb{Z}^N$ integer programming is required, and this may be prohibitively difficult.
\footnote{
We sketch the basic idea. A cell is determined by the $K$ nearest landmarks: w.l.o.g. (i.e., up to ignorable degeneracy), we have $d(x,x_{\mathcal{I}(1)}) < \dots < d(x,x_{\mathcal{I}(K)})$, where $\mathcal{I}$ is the set of $L$ landmark indices for the set $\{x_i\}_{i=1}^T$ of initial states. For Euclidean space, this is the same as $\|x_{\mathcal{I}(1)}\|^2 - 2 \langle x_{\mathcal{I}(1)},x \rangle < \dots < \|x_{\mathcal{I}(K)}\|^2 - 2 \langle x_{\mathcal{I}(K)},x \rangle$. A point $x$ on the boundary of a cell can thus be produced by linear (or, according to context, integer or binary) programming. Moving towards landmarks in a controlled manner then yields a point in the interior.
}
For other spaces there may not be a suitable state generation mechanism at all.

Moreover, if we do not incorporate $d$ into the generalized Kendall or Cayley distance, symmetry just leads to a uniform weighting on cells (some of which might also conceivably be degenerate), which will not be adequate for the purpose of promoting diversity beyond the generation of landmarks. While for the ordinary Kendall distance the corresponding similarity matrix (viz., the so-called Mallows kernel) is positive definite for any scale \cite{jiao2015kendall}, the analogous statement for a generalization along the lines of \cite{kumar2010generalized} would have to be established (or worked around by operating at scale $t_+$). See also \cite{fagin2003comparing,van2021using} for other relevant practicalities in this context.

\section{\label{sec:couponCollection}How often to go}

Given a distribution $p$ on elites, we want to sample from $p$ often enough so that a sufficient number of elites serve as bases for exploring the space, but not so many times as to be infeasible. Meanwhile, to mitigate bias in sampling, it makes sense to sample over the course of discrete epochs. The classical \emph{coupon collector's problem} \cite{flajolet1992birthday} provides a suitable framework in which a company issues a large pool of coupons with $n$ types that are distributed according to $p$.

Per Corollary 4.2 of \cite{flajolet1992birthday}, we have that the expected time for the event $C_m$ of collecting $m$ of $n$ coupon types via IID draws from the distribution $(p_1,\dots,p_n)$ satisfies 
\begin{equation}
\label{eq:partialCoupon}
\mathbb{E}(C_m) = \sum_{\ell=0}^{m-1} (-1)^{m-1-\ell} \binom{n-\ell-1}{n-m} \sum_{|L| = \ell} \frac{1}{1-P_L}
\end{equation}
with $P_L := \sum_{k \in L} p_k$. The specific case $m = n$ admits an integral representation that readily admits numerical computation, 
\footnote{
While an integral representation of $\mathbb{E}(C_m)$ exists for generic $m$, it is also combinatorial in form and the result \eqref{eq:partialCoupon} of evaluating it symbolically is easier to compute.
}
viz. 
\begin{equation}
\label{eq:totalCoupon}
\mathbb{E}(C_n) = \int_0^\infty \left ( 1 - \prod_{k=1}^n [1-\exp(-p_k t)] \right ) \ dt.
\end{equation} 
However, the sum \eqref{eq:partialCoupon} is generally hard or impossible to evaluate in practice due to its combinatorial complexity, and it is desirable to produce useful bounds.
\footnote{
A coarser approach to coupon collection than the granular approach of considering a distribution over elites would be to determine the number of samples from $p$ required to visit every $\sigma^{(1)}$ (recall that these are the Voronoi cells of landmarks). To do this, we only need to group and add the relevant entries of $p$, then apply \eqref{eq:totalCoupon}. However, we do not pursue this here.
}

\begin{figure}[h]
  \centering
  \includegraphics[trim = 72mm 120mm 79mm 115mm, clip, width=.32\columnwidth,keepaspectratio]{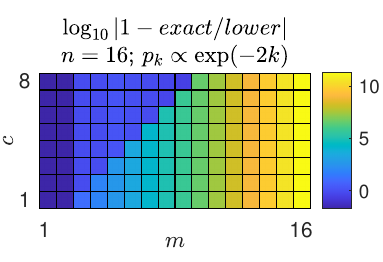}
  \includegraphics[trim = 72mm 120mm 79mm 115mm, clip, width=.32\columnwidth,keepaspectratio]{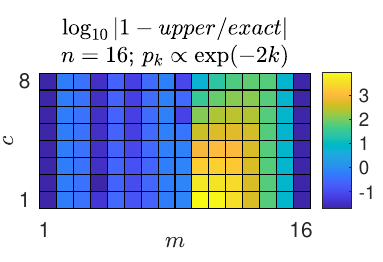}
  \includegraphics[trim = 72mm 120mm 79mm 115mm, clip, width=.32\columnwidth,keepaspectratio]{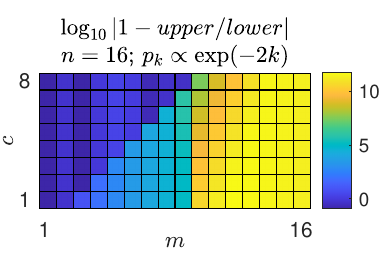} 
  \\
  \includegraphics[trim = 72mm 120mm 79mm 115mm, clip, width=.32\columnwidth,keepaspectratio]{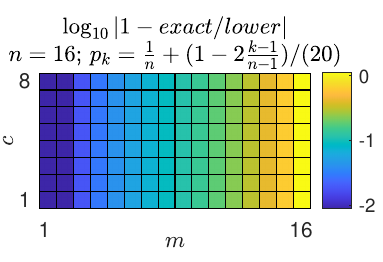}
  \includegraphics[trim = 72mm 120mm 79mm 115mm, clip, width=.32\columnwidth,keepaspectratio]{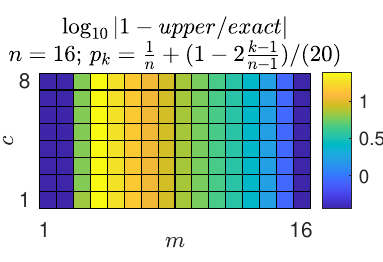}
  \includegraphics[trim = 72mm 120mm 79mm 115mm, clip, width=.32\columnwidth,keepaspectratio]{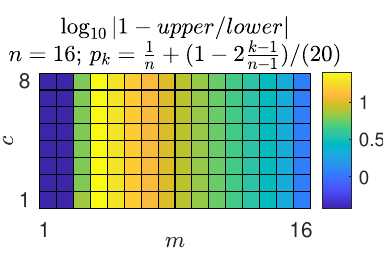} 
  \\
  \includegraphics[trim = 72mm 120mm 79mm 115mm, clip, width=.32\columnwidth,keepaspectratio]{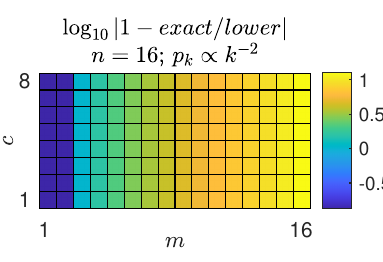}
  \includegraphics[trim = 72mm 120mm 79mm 115mm, clip, width=.32\columnwidth,keepaspectratio]{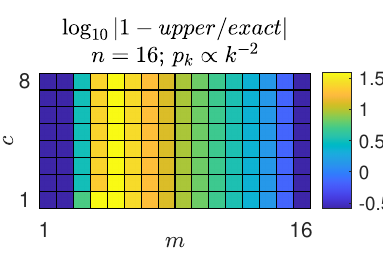}
  \includegraphics[trim = 72mm 120mm 79mm 115mm, clip, width=.32\columnwidth,keepaspectratio]{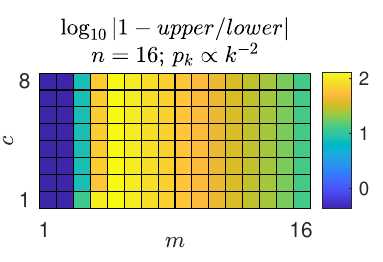}   \caption{Comparison of exact value and (merging of various) upper/lower bounds (omitting exact calculations which are feasible for $m \lesssim 20$ from the bounds themselves) for $\mathbb{E}(C_m)$ for various probability distributions with $n = 16$. These results suggest in particular that for the case where $p$ is approximately uniform, a readily computable lower bound is reasonably accurate.  
  }
  \label{fig:couponBounds}
\end{figure}

\begin{figure}[h]
  \centering
  \includegraphics[trim = 72mm 120mm 79mm 115mm, clip, width=.32\columnwidth,keepaspectratio]{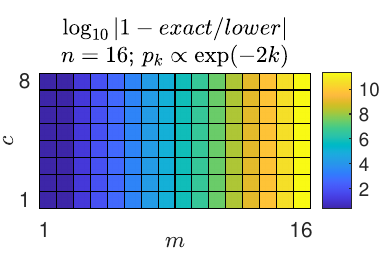}
  \includegraphics[trim = 72mm 120mm 79mm 115mm, clip, width=.32\columnwidth,keepaspectratio]{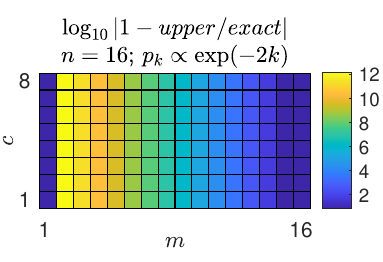}
  \includegraphics[trim = 72mm 120mm 79mm 115mm, clip, width=.32\columnwidth,keepaspectratio]{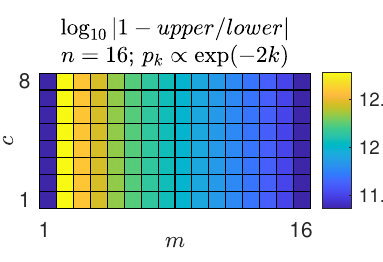} 
  \\
  \includegraphics[trim = 72mm 120mm 79mm 115mm, clip, width=.32\columnwidth,keepaspectratio]{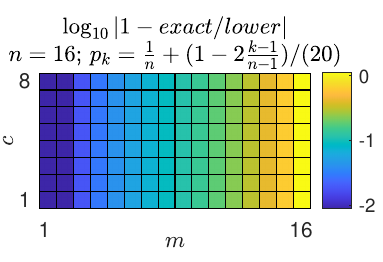}
  \includegraphics[trim = 72mm 120mm 79mm 115mm, clip, width=.32\columnwidth,keepaspectratio]{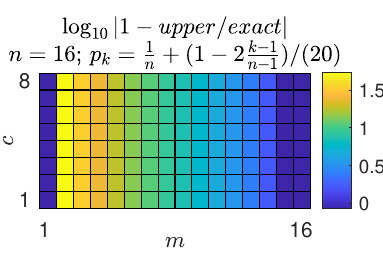}
  \includegraphics[trim = 72mm 120mm 79mm 115mm, clip, width=.32\columnwidth,keepaspectratio]{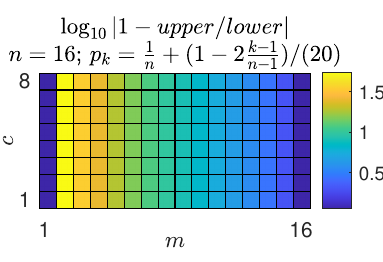} 
  \\
  \includegraphics[trim = 72mm 120mm 79mm 115mm, clip, width=.32\columnwidth,keepaspectratio]{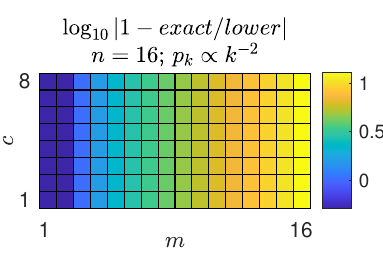}
  \includegraphics[trim = 72mm 120mm 79mm 115mm, clip, width=.32\columnwidth,keepaspectratio]{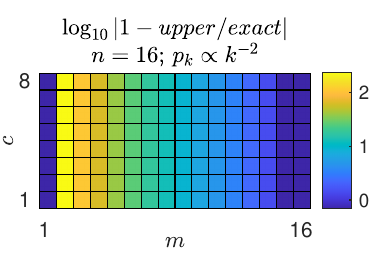}
  \includegraphics[trim = 72mm 120mm 79mm 115mm, clip, width=.32\columnwidth,keepaspectratio]{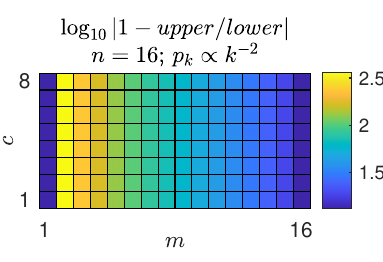}   \caption{As in Figure \ref{fig:couponBounds}, but without including bounds that involve a cutoff $c$. Again, these results suggest in particular that for the case where $p$ is approximately uniform, a readily computable lower bound is reasonably accurate.  
  }
  \label{fig:couponBoundsX}
\end{figure}

As Figures \ref{fig:couponBounds} and \ref{fig:couponBoundsX} illustrate, reasonably tight lower bounds turn out to be readily computable in practice, 

Towards this end, assume w.l.o.g. that $p_1 \ge \dots \ge p_n$, and let $c \le n$. (For clarity, it is helpful to imagine that $c < n$ and $p_c \gg p_{c+1}$, but we do not assume this.) To bound $\sum_{|L| = \ell} (1-P_L)^{-1}$, we first note that $\{L : |L| = \ell\}$ is the union of disjoint sets of the form $\{L : |L| = \ell \text{ and } L \cap [\lambda] = M\}$ for $M \in 2^{[\lambda]}$, where $\lambda := \min \{c,\ell\}$. Thus
\begin{equation}
\label{eq:bound1a}
\sum_{|L| = \ell} \frac{1}{1-P_L} = \sum_{M \in 2^{[\lambda]}} \sum_{\substack{|L| = \ell \\ L \cap [\lambda] = M}} \frac{1}{1-P_L}.
\end{equation}

Now $P_L = P_{L \cap [\lambda]} + P_{L \backslash [\lambda]}$. If we are given bounds of the form $\pi_* \le P_{L \backslash [\lambda]} \le \pi^*$, we get in turn that
$$\frac{1}{1-P_{L \cap [\lambda]}-\pi_*} \le \frac{1}{1-P_L} \le \frac{1}{1-P_{L \cap [\lambda]}-\pi^*}.$$
If furthermore $\pi_*$ and $\pi^*$ depend on $L$ only via $L \cap [\lambda]$, then 
\begin{align}
\label{eq:bound1b}
\frac{|\{L : |L| = \ell \text{ and } L \cap [\lambda] = M\}|}{1-P_M-\pi_*} & \le \sum_{\substack{|L| = \ell \\ L \cap [\lambda] = M}} \frac{1}{1-P_L} \nonumber \\
& \le \frac{|\{L : |L| = \ell \text{ and } L \cap [\lambda] = M\}|}{1-P_M-\pi^*}.
\end{align}
Meanwhile, writing $\mu := |M|$ and combinatorially interpreting the Vandermonde identity $\sum_\mu \binom{\lambda}{\mu} \binom{n-\lambda}{\ell-\mu} = \binom{n}{\ell}$ yields
\begin{equation}
\label{eq:bound1c}
|\{L : |L| = \ell \text{ and } L \cap [\lambda] = M\}| = \binom{n-\lambda}{\ell-\mu}
\end{equation}
and in turn bounds of the form
\begin{align}
\label{eq:bound1d}
\sum_{M \in 2^{[\lambda]}} \binom{n-\lambda}{\ell-\mu} \frac{1}{1-P_M-\pi_*} & \le \sum_{|L| = \ell} \frac{1}{1-P_L} \nonumber \\
& \le \sum_{M \in 2^{[\lambda]}} \binom{n-\lambda}{\ell-\mu} \frac{1}{1-P_M-\pi^*}.
\end{align}

Now the best possible choice for $\pi_*$ is  $P_{[\ell-\mu]+n-(\ell-\mu)}$; similarly, the best possible choice for $\pi^*$ is $P_{[\ell-\mu]+\min\{\lambda,n-(\ell-\mu)\}}$.
This immediately yields upper and lower bounds for \eqref{eq:partialCoupon}, though the alternating sign term leads to intricate expressions that are not worth writing down explicitly outside of code. 

The resulting bounds are hardly worth using in some situations, and quite good in others. We augment them with the easy lower bound obtained by using the uniform distribution in \eqref{eq:partialCoupon} \cite{anceaume2015new} and the easy upper bound obtained by taking $m = n$ and using \eqref{eq:totalCoupon}; we also use the exact results when feasible (e.g., $n$ small or $m = n$) as both upper and lower bounds. These basic augmentations have a significant effect in practice. 

Experiments on exactly solvable (in particular, small) cases show that though the bounds for $(1-P_L)^{-1}$ are good, the combinatorics involved basically always obliterates the overall bounds for distributions of the form $p_k \propto k^{-\gamma}$ with $\gamma$ a small positive integer. However, the situation improves dramatically for distributions that decay quickly enough.

We can similarly also derive bounds along the lines above based on the deviations $\delta_k := n p_k - 1$. The only significant difference in the derivation here \emph{versus} the one detailed above is that we are forced to consider absolute values of the deviations, so although these bounds are more relevant to our context, they are also looser in practice. We provide a sketch along preceding lines. Write $\Delta_L := \sum_{k \in L} \delta_k$, so that $P_L = (|L|+\Delta_L)/n$. Assuming w.l.o.g. that $|\delta_1| \ge \dots \ge |\delta_n|$, we have $\sum_{|L| = \ell} (1-P_L)^{-1} = \sum_{M \in 2^{[\lambda]}} \sum_{|L| = \ell; L \cap [\lambda] = M} (1-[\ell+\Delta_L]/n)^{-1}$. If $\pi_* \le \Delta_{L \backslash [\lambda]} \le \pi^*$ with $\pi_*$, $\pi^*$ depending only on $L \cap [\lambda]$, then $\sum_{M \in 2^{[\lambda]}} \binom{n-\lambda}{\ell-\mu} (1-[\ell+\Delta_M+*]/n)^{-1}$ is a lower (resp., upper) bound for $\sum_{|L| = \ell} (1-P_L)^{-1}$ for $*$ indicating $\pi_*$ (resp., $\pi^*$). Meanwhile, the best possible choice for $\pi^*$ is $\sum_{k = \min\{\lambda,n-(\ell-\mu)\}+1}^{\min\{\lambda,n-(\ell-\mu)\}+(\ell-\mu)} | \delta_k | \ge \Delta_{L \backslash [\lambda]}$ and $\pi_* = -\pi^*$.

In the near-uniform regime we also have the simple and tight lower bound $\mathbb{E}(C_m) \ge n(H_n - H_{n-m})$, where $H_n := \sum_{k=1}^n k^{-1}$ is the $n$th harmonic number \cite{flajolet1992birthday,anceaume2015new}. In fact this bound is quite good for the linear case in figures below and $n$ small, to the point that replacing harmonic numbers with logarithms can easily produce larger deviations from the bound than the error itself.

\section{\label{sec:bandwidthNote}Bandwidth in $\mathbb{R}^N$}

In the setting of $\mathbb{R}^N$, we can deploy some analysis beforehand. First, recall the standard formula $\int_{\mathbb{R}^N} \psi(|x|) \ dx = \omega_{N-1} \int_0^\infty \psi(r) r^{n-1} \ dr$ where $\omega_{N-1} = 2\pi^{N/2}/\Gamma(N/2)$ is the Hausdorff-Lebesgue measure (i.e., generalized notion of surface area) of $S^{N-1}$. Writing here $\nu(x) = (2\pi)^{-N/2} \exp(-|x|^2/2)$ for the standard Gaussian, we have that $\int_{B_R(0)} \nu(x) \ dx = 1 - \Gamma(N/2,R^2/2)/\Gamma(N/2)$, where the numerator in the rightmost expression is the (upper) incomplete gamma function. From this it follows in particular that $\int_{B_{\sqrt{N}}(0)} \nu(x) \ dx \approx 1/2$.  

This nominally gives us a way to relate the bandwidth (i.e., standard deviation) of a spherical Gaussian distribution to the geometry of a cell under the \emph{Ansatz} that we are sampling near the center of a roughly spherical cell. However, in high dimensions the vast majority of a cell's volume will be near its boundary, and a spherical (or for that matter, even ellipsoidal) approximation of the cell geometry will also generally be terrible. On the other hand, a slightly more detailed analysis along the lines above yields that all but exponentially little of the probability mass of the standard Gaussian lies in a thin spherical shell with radius centered at $\sqrt{N}$ \cite{blum2020foundations}. If now $\delta_*$ is the distance from the origin to the nearest part of its cell boundary and $\delta^*$ is the distance to the farthest part of its cell boundary, then we expect that $\int_{B_R(0)} \nu \ dx$ is nearly unity for $R < \delta_*$ and nearly zero for $R > \delta^*$, and typically with (something like) a plateau at $1/2$ for a not insignificant portion of the interval between these. In other words, we can still determine a characteristic radius of the cell by finding a bandwidth $\theta$ for Gaussian sampling that yields about half of the  sample points in the cell.

\section{\label{sec:exploreAlternatives}Alternative techniques for exploring}

For the sake of generality, first assume that we cannot estimate $f$ (say, because $X$ is not the sort of space that enables estimation). In this event, exploration is conceptually simple. 

If parallel evaluation is not possible, simply select the point among random candidates that maximizes the differential magnitude relative to prior points in (or that by prior exploration led to) the current cell. Because the number of prior points in any cell should be fairly small (because the objective is expensive), we can usually take them \emph{in toto}, but if this assumption is violated we can restrict consideration to at most a fixed number of points with the largest weighting components (at scale zero since we need not form a probability distribution, and possibly doing other tricks like reserving space for and uniformly sampling some additional prior points). In practice (because of approximate submodularity) we can usually settle for the approximation of a single weighting incorporating all candidate points. 

If parallel evaluation is possible, we can do something similar to the landmark generation process to pick a fixed-size subset of candidates that (approximately) maximally increase the magnitude of the prior points. Alternatively, we can compute the weighting of all prior and candidate points and select the top candidates that result. (This has the benefit of being faster.)

Suppose now that we can estimate $f$ (say by RBF interpolation): then we certainly should incorporate that estimate into the exploration process. 

Again, suppose further that parallel evaluation is not possible: then we should pick the candidate point that optimizes the (W)QD-score \eqref{eq:WQD} for $w$ the weighting on cells at the prior $t_+$. Alternatively, we can optimize whatever other gauge of performance we care about.

\section{\label{sec:GoExploreBaseline}Baseline version of Go-Explore for \S \ref{sec:benchmarkingReal}}

\begin{algorithm}
  \caption{\textsc{GoExploreBaseline}$(f,d,L,T,K,G,M,g)$}
  \label{alg:GoExploreBaseline}
\begin{algorithmic}[1]
  \STATE Generate landmarks as subset of initial states $X'$ using $G$ \hfill \emph{// Algorithm \ref{alg:GoExploreLandmarks}}
  \STATE Evaluate $f$ on $X'$
  \STATE Evaluate $\sigma^{(K)}$ on $X'$ and initialize history $h$ \hfill \emph{// Algorithm \ref{alg:GoExploreCell}}
  \WHILE{$|h|<M$}
   \STATE $E \leftarrow \bigcup_\tau \{ \arg \min_{x \in h: \sigma^{(K)}(x) = \tau} f(x) \}$ \hfill \emph{// Elites}
   \STATE Compute weighting $w$ at scale $t_+$ on $E$
   \STATE Form diversity-maximizing PDF $p$ \hfill \emph{// Pure exploration}
   \STATE Compute number $b$ of expeditions \hfill \emph{// $b \leftarrow \lceil |E| \log |E| \rceil$}
    \FOR{$b$ steps}
      \STATE Sample $x \sim p$ 
      \STATE Compute exploration effort $\mu_*$ \hfill \emph{// $ \mu_* \leftarrow 10$}
      \STATE \emph{// $g(\cdot | x, \theta_0) \leftarrow $ Gaussian with covariance $\theta_0^2 I$}
      \STATE Sample $X' \sim g^{\times \mu_*}(\cdot | x,\theta_0)$
    \ENDFOR
    \STATE Evaluate $f$ on $X'$
    \STATE Evaluate $\sigma^{(K)}$ on $X'$ and update $h$
  \ENDWHILE
  \ENSURE $h$ (obtain set $E$ of globally diverse and locally optimal elites as above)
\end{algorithmic}
\end{algorithm}

\begin{figure}[h]
  \centering
  \includegraphics[trim = 35mm 75mm 35mm 75mm, clip, width=.49\columnwidth,keepaspectratio]{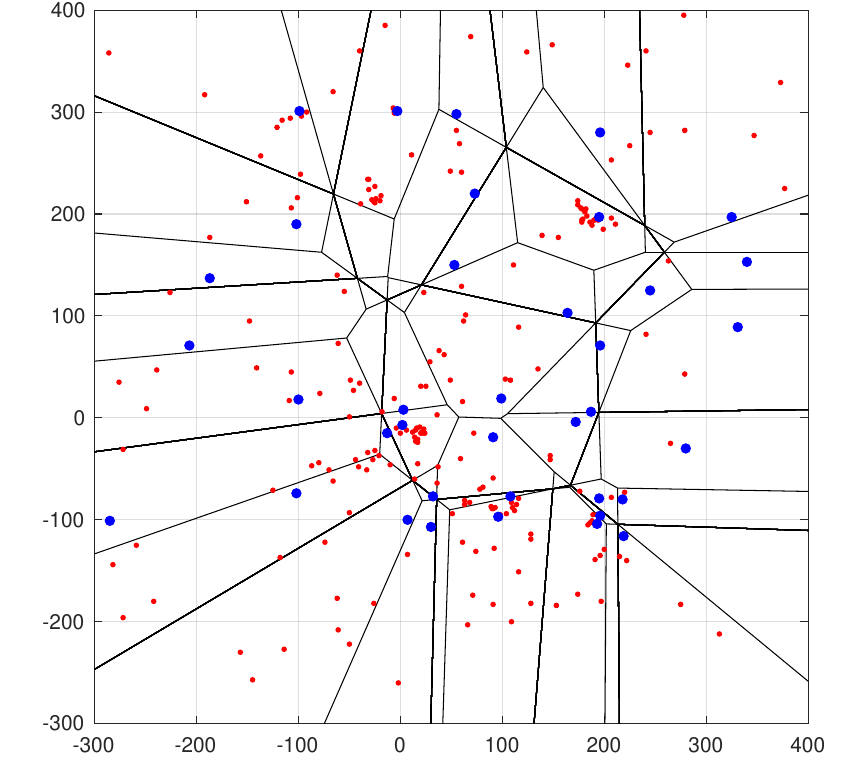}
  \includegraphics[trim = 35mm 75mm 35mm 75mm, clip, width=.49\columnwidth,keepaspectratio]{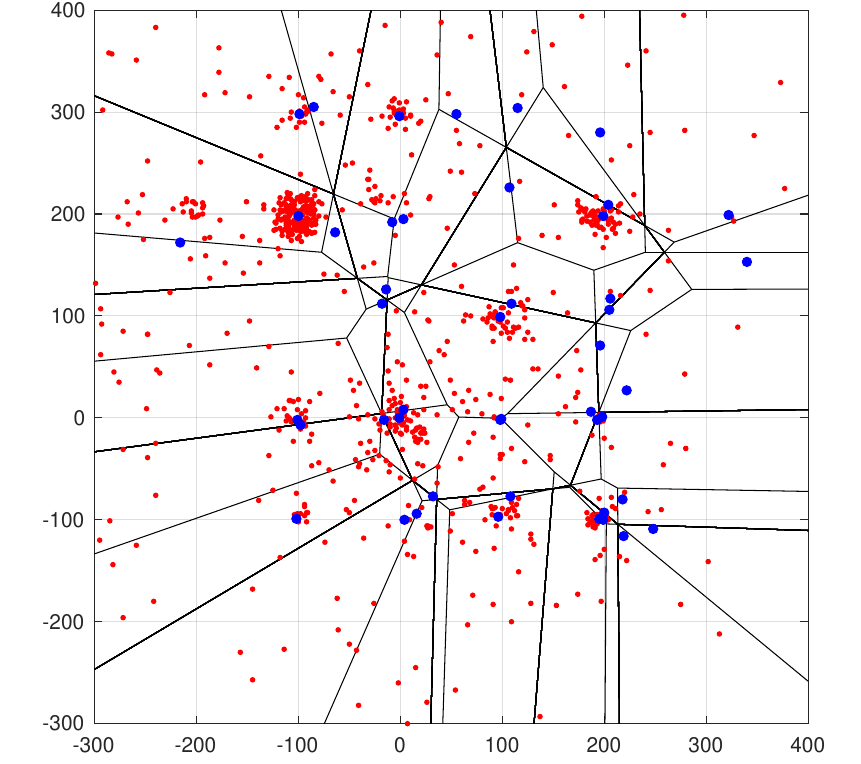}
\caption{(L) As in the right panel of Figure \ref{fig:Rastrigin20221212}, but for the scaled and discretized variant described in \S \ref{sec:integer} with $\lambda = 100$. The domain is $\mathbb{Z}^2$; $G$ is given by rounding the uniform distribution on $(\lambda \cdot [-2,3])^2$, and $g$ is given by rounding a spherical Gaussian away from zero. (R) As in the left panel, but for $M = 1000$.
  }
  \label{fig:RastriginInteger_300_20221212}
\end{figure}

\begin{figure}[h]
  \centering
  \includegraphics[trim = 35mm 75mm 35mm 75mm, clip, width=.49\columnwidth,keepaspectratio]{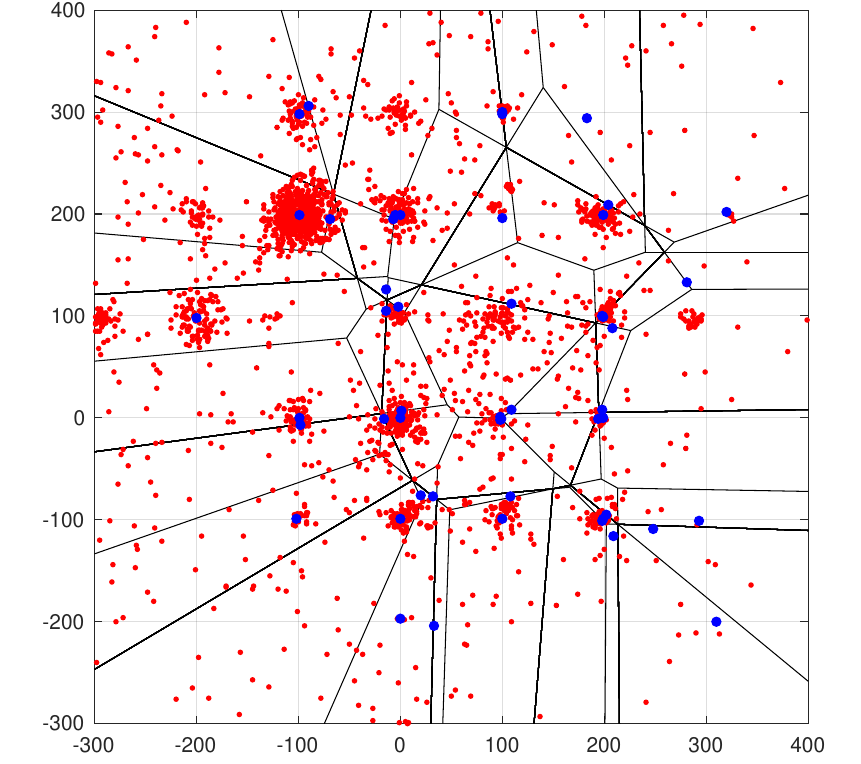}
  \includegraphics[trim = 35mm 75mm 35mm 75mm, clip, width=.49\columnwidth,keepaspectratio]{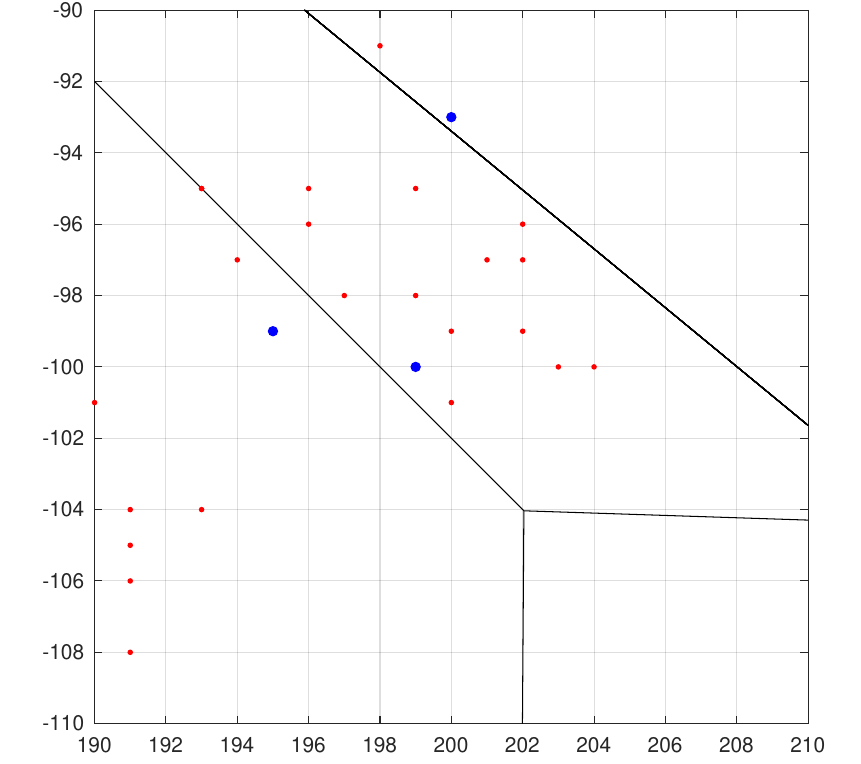}
\caption{(L) As in the right panel of Figure \ref{fig:RastriginInteger_300_20221212}, but for $M = 3000$. (R) Detail of the $M = 1000$ example from the left panel of Figure \ref{fig:RastriginInteger_300_20221212}.
  }
  \label{fig:RastriginInteger_3000_20221212}
\end{figure}

\section{\label{sec:IntegerResults}Results for \S \ref{sec:integer}: Figures \ref{fig:RastriginInteger_300_20221212}-\ref{fig:RastriginInteger_3000_20221212}}

\section{\label{sec:regex}Another example over $\mathbb{Z}^N$: regular expressions}

\begin{figure}[h]
  \centering
  \includegraphics[trim = 45mm 90mm 40mm 85mm, clip, width=.9\columnwidth,keepaspectratio]{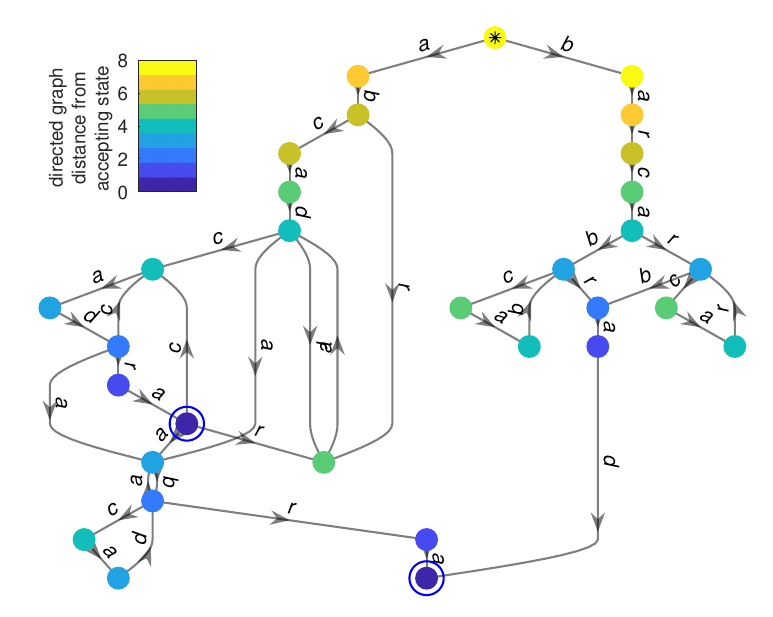}
\caption{Minimal DFA for the regular expression \eqref{eq:abracadabra}. The initial state is shown with an asterisk; the accepting states are shown with circles. For clarity, a garbage state and transitions to it are not shown. States are colored by their digraph distance from one of the two accepting states.
  }
  \label{fig:DFA_20220902}
\end{figure}

Figure \ref{fig:DFA_20220902} shows the minimal discrete finite automaton (DFA) for the regular expression
\begin{equation}
\label{eq:abracadabra}
\texttt{(ab(ra|cad)+(cad|ab)+ra)|(bar(car)+bad)|(bar(cab)+rad)}
\end{equation}
over the alphabet $\mathcal{A} := \{\texttt{a},\texttt{b},\texttt{c},\texttt{d},\texttt{r}\}$. Let $\Omega$ be the set of accepting states: both elements are circled in the figure.

Let $\iota : \{0,\dots,5\}^{16} \rightarrow \mathcal{A}^*$ be defined by mapping (only) any initial nonzero entries to the corresponding elements of $\mathcal{A}$, e.g., $\iota(1,2,5,1,0,\dots,0,3) = \texttt{abra}$ and $\iota(0,1,2,5,1,0,\dots,0,3) = \epsilon$, where $\epsilon$ indicates the empty string. 

\begin{figure}[h]
  \centering
  \includegraphics[trim = 60mm 100mm 60mm 100mm, clip, width=.75\columnwidth,keepaspectratio]{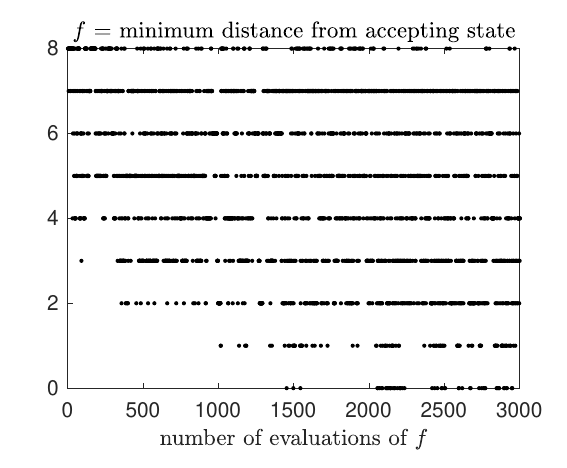}
\caption{Values of \eqref{eq:objectiveDFA} obtained from a run of Algorithm \ref{alg:GoExploreDissimilarity} with primary inputs as described in the text; $L = 15$, $T = \lceil L \log L \rceil = 41$, $K = 2$, and $\mu = 128$. Note that we reach an accepting state after fewer than $1500$ function evaluations, while there are $5^8 = 390625$ strings over $\mathcal{A}$ of length equal to the unique shortest valid string.
  }
  \label{fig:DFA_history_short_20221212}
\end{figure}

\begin{figure}[h]
  \centering
  \includegraphics[trim = 65mm 100mm 65mm 105mm, clip, width=.9\columnwidth,keepaspectratio]{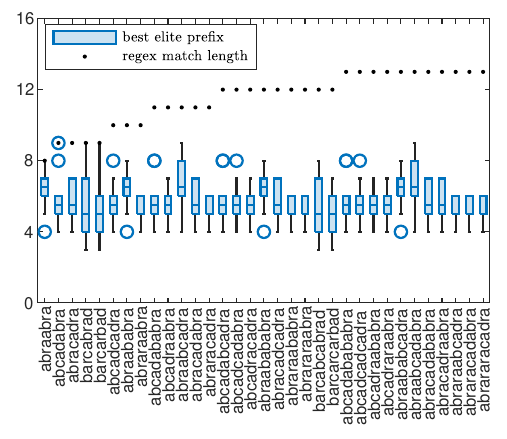}
\caption{Boxchart detailing the extent of the best elite prefix for each short regular expression match in an ensemble of 10 runs of Algorithm \ref{alg:GoExploreDissimilarity} with primary inputs as described in the text; $L = 15$, $T = \lceil L \log L \rceil = 41$, $K = 2$, $\mu = 128$, and $M = 1000$. The boxchart shows the central quartiles as boxes, outliers (determined via interquartile range) as circles, and the range without outliers as lines. Dots indicate the length of the string indicated on the horizontal axis. Note that strings with the prefix \texttt{bar} are an area of relative underperformance, presumably because this area has relatively fewer extrema---and higher barriers between them---in the first place. Note also that medians are typically $\ge 5$, while $5^5 = 3125 \gg M$.
  }
  \label{fig:regexProgress1000_20221212}
\end{figure}

\begin{figure}[h]
  \centering
  \includegraphics[trim = 65mm 100mm 65mm 105mm, clip, width=.9\columnwidth,keepaspectratio]{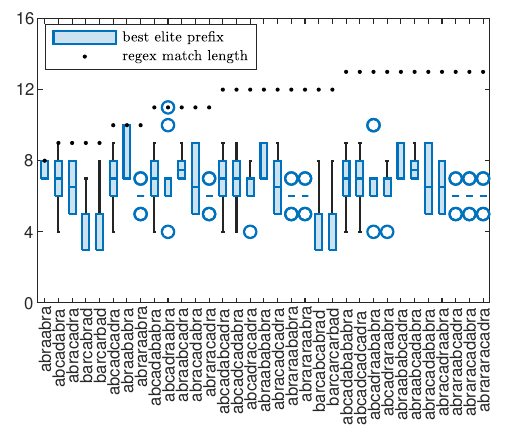}
\caption{As in Figure \ref{fig:regexProgress1000_20221212}, but for $M = 3000$.
  }
  \label{fig:regexProgress3000_20221212}
\end{figure}

\begin{figure}[h]
  \centering
  \includegraphics[trim = 65mm 100mm 65mm 105mm, clip, width=.9\columnwidth,keepaspectratio]{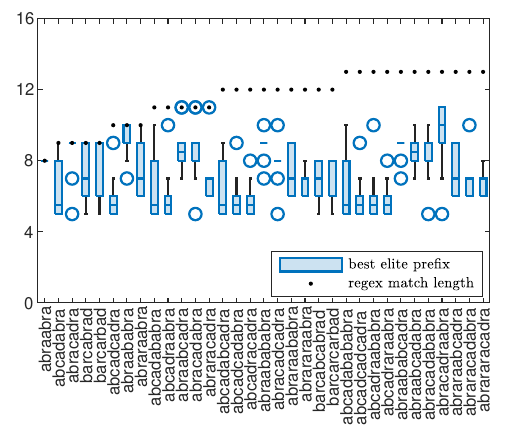}
\caption{As in Figure \ref{fig:regexProgress1000_20221212}, but for $M = 10000$. Note that one of the outliers has a prefix corresponding to the regular expression match \texttt{abracadabra}.
  }
  \label{fig:regexProgress_20221212}
\end{figure}

We define the objective
\begin{equation}
\label{eq:objectiveDFA}
f(x) := \min_{\text{prefixes } w \text{ of } \iota(x)} d_{DFA}(w,\Omega)
\end{equation}
where $d_{DFA}(w,\Omega)$ indicates the distance on the DFA digraph between the last state reached by $w$ and the accepting states $\Omega$. We take $d$ to be Hamming distance; the global generator $G$ to be sampling from $\mathcal{U}([5]^8 \times \{0\}^8)$; and the local generator $g(\cdot |x,\theta)$ to be uniformly changing, appending, or truncating nonzero initial entries $\theta$ times. The results of running Algorithm \ref{alg:GoExploreDissimilarity} are shown in Figures \ref{fig:DFA_history_short_20221212}-\ref{fig:regexProgress_20221212}.

The problem we are solving here is easier than but related to regular language induction from membership queries alone (which is not in $\mathbf{P}$: a polytime solution requires additional language queries \cite{angluin1987learning}). It is morally an instance of directed greybox fuzzing \cite{bohme2017directed} in which we try to obtain multiple valid strings and many long diverse prefixes of valid strings. In fact, this toy problem broadly illustrates directed greybox fuzzing in general: a control flow graph can be represented with a DFA, and in turn a regular expression \cite{toprak2014lightweight}. 

\section{\label{sec:OtherBinaryExamples}Other examples over $\mathbb{F}_2^N$}

\subsection{\label{sec:SKsupplement}Supplementary figures for \S \ref{sec:binary}}

\begin{figure}[h]
  \centering
  \includegraphics[trim = 35mm 100mm 35mm 90mm, clip, width=.9\columnwidth,keepaspectratio]{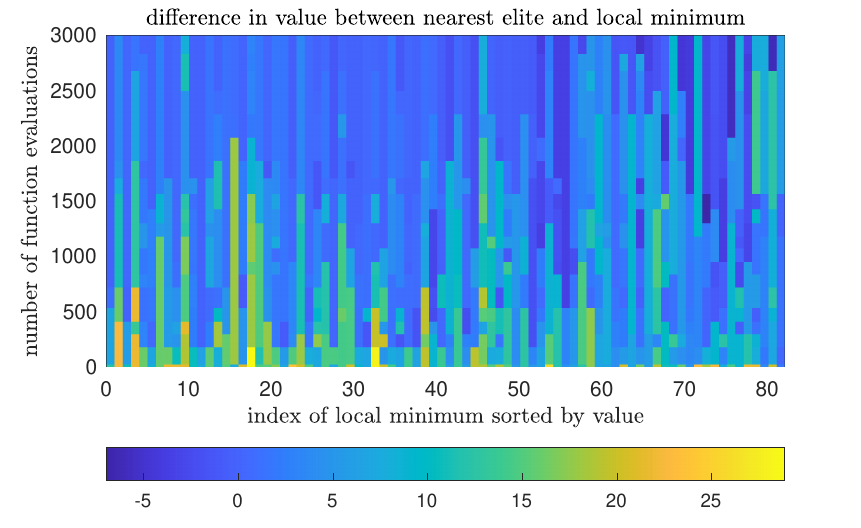}
\caption{Another view of performance with the same data used for Figure \ref{fig:barriersSK20221212}. Barriers are not shown in lieu of a more detailed time evolution.
  }
  \label{fig:valueSK20221212}
\end{figure}

Figure \ref{fig:valueSK20221212} shows the performance of Algorithm \ref{alg:GoExploreDissimilarity} on an instance of \eqref{eq:SK} with $N = 20$. 

\begin{figure}[h]
  \centering
  \includegraphics[trim = 35mm 100mm 35mm 90mm, clip, width=.9\columnwidth,keepaspectratio]{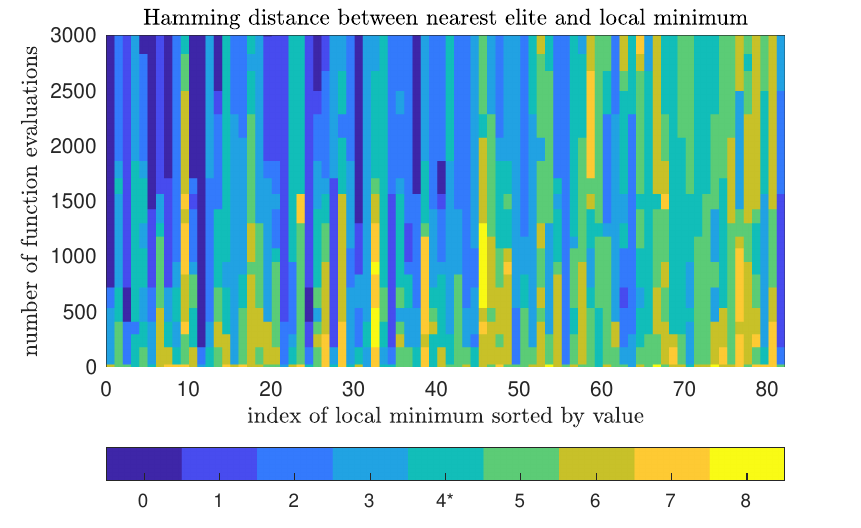}
\caption{As in Figure \ref{fig:valueSK20221212}, but for Hamming distance. The $*$ on the colorbar indicates the figure of merit $r'_E$ in \eqref{eq:figureOfMerit}.
  }
  \label{fig:distanceSK20221212}
\end{figure}

One aspect of Figure \ref{fig:distanceSK20221212}--the small asterisk $*$ in the colorbar indicating a figure of merit--requires some explanation. Let $A(N,r)$ be the maximum number of points in $\mathbb{F}_2^N$ with pairwise Hamming distance $\le r$. The classical binary Hamming bound \cite{moon2020error} is
\begin{equation}
\label{eq:HammingBound}
A(N,r) \le \frac{2^N}{\sum_{k=0}^r \binom{N}{k}},
\end{equation} 
and its proof just amounts to observing that the denominator on the RHS is the volume of a Hamming sphere of radius $r$. This immediately yields a figure of merit for the Hamming distance between elites and minima. For a set $E \subset \mathbb{F}_2^N$ of elites, let 
$$r_E := \max \{r \in \mathbb{N} : |E| \le A(N,r)\}.$$
In the extreme case that $E$ is a ``perfect'' error-correcting code, then $r_E$ is just the ``distance'' of the code. In general, the points of $E$ will not be so precisely equispaced as this, and $A(N,r)$ is not easily computable, but we can nevertheless get a simple and useful figure of merit as follows. By \eqref{eq:HammingBound}, 
\begin{equation}
\label{eq:figureOfMerit}
r_E \le r'_E:= \max \left \{r \in \mathbb{N} : |E| \le \frac{2^N}{\sum_{k=0}^r \binom{N}{k}} \right \}
\end{equation} 
is easily computable. If the distance between elites and local minima tends to be significantly less than $r'_E$, then because Hamming balls have volume exponential in $N$, the elites have earned their status. In Figure \ref{fig:distanceSK20221212}, this is evidently the case for most of the minima, with exceptions tending to be the shallower minima.

\subsection{\label{sec:LABS}Low-autocorrelation binary sequences}

An even harder problem than that of \S \ref{sec:binary} is that of finding low-autocorrelation binary sequences (LABS). As \cite{doerr2020benchmarking} points out, the LABS problem represents a ``grand combinatorial challenge with practical applications in radar engineering and measurement.'' Recall that the autocorrelation of $s \in \{\pm 1\}^N$ is $R_k(s) := \sum_{j=1}^{N-k} s_j s_{j+k}$. There are two common versions of the LABS problem, which can broadly be distinguished as being studied by physicists \cite{ferreira2000landscape,packebusch2016low} and by engineers \cite{brest2021low}. The former, called the Bernasconi model \cite{bernasconi1987low}, turns out to have higher barriers than the latter, suggesting a more structured landscape. It is defined via the energy function 
\begin{equation}
\label{eq:Bernasconi}
f(s) = \sum_{k=1}^{N-1} R_k^2(s),
\end{equation}
whereas the engineers' version is defined via $f(s) = \max_{k > 0} |R_k(s)|$. 

\begin{figure}[h]
  \centering
  \includegraphics[trim = 35mm 80mm 35mm 80mm, clip, width=.9\columnwidth,keepaspectratio]{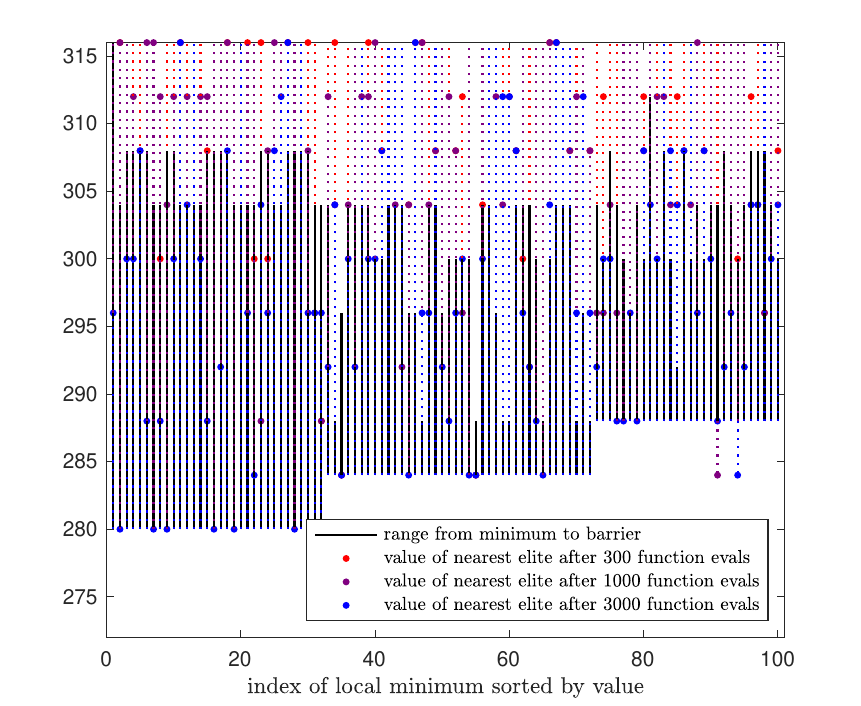}
\caption{As in Figure \ref{fig:barriersSK20221212}, but for \eqref{eq:Bernasconi} with $N = 16$. Only the 100 lowest minima are shown.
}
  \label{fig:barriersLABS20221212}
\end{figure}

\begin{figure}[h]
  \centering
  \includegraphics[trim = 35mm 100mm 35mm 90mm, clip, width=.9\columnwidth,keepaspectratio]{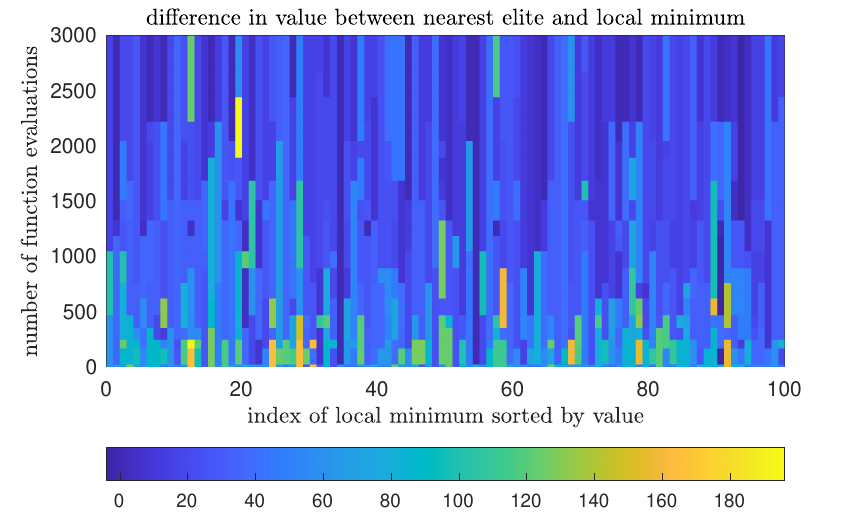}
\caption{As in Figure \ref{fig:valueSK20221212}, but for \eqref{eq:Bernasconi} with $N = 16$. Only the 100 lowest minima are shown.
  }
  \label{fig:valueLABS20221212}
\end{figure}

\begin{figure}[h]
  \centering
  \includegraphics[trim = 35mm 100mm 35mm 90mm, clip, width=.9\columnwidth,keepaspectratio]{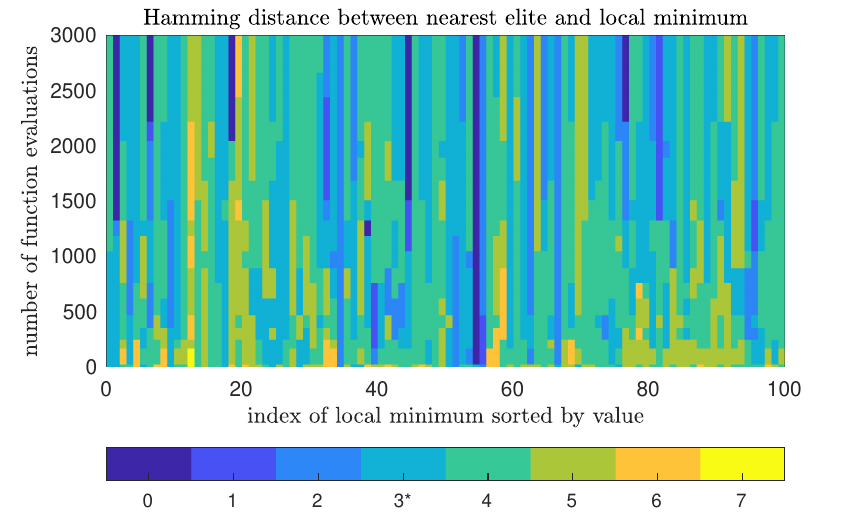}
\caption{As in Figure \ref{fig:distanceSK20221212}, but for \eqref{eq:Bernasconi} with $N = 16$. Only the 100 lowest minima are shown.
  }
  \label{fig:distanceLABS20221212}
\end{figure}

Figures \ref{fig:barriersLABS20221212}-\ref{fig:distanceLABS20221212} show results for \eqref{eq:Bernasconi} with $N = 16$. With only 3000 evaluations, we have found 6 of 32 optimal sequences in a space with $2^{16}$ elements, and 15 of the 100 most optimal sequences.


\subsection{\label{sec:checks}Checksums and cyclic redundancy checks}

Packet headers for version 4 of the Internet Protocol (IPv4) \cite{postel1981rfc791} include, among others, fields for the version (= 4), header length, and a 16-bit checksum \cite{braden1988rfc1071}. We ran our algorithm for $3000$ steps on $\mathbb{F}_2^{160}$ with 
\begin{align}
f & = 4 d_H(\text{version},4) + 4 d_H(\text{length},160) \nonumber \\
& + d_H(\text{nominal checksum},\text{IPv4 checksum}), \nonumber
\end{align}
$L = 6$, $T = \lceil L \log L \rceil$, $K = 2$, $G$ uniform, and $g$ given by bit flips with a Bernoulli parameter of $q$ for all bits except the ones in the objective $f$, which have a Bernoulli parameter of $10q$, and $\mu = 128$.

This resulted in 30 elites, four of which were ``perfect'' in that their version, header length, and checksum were valid. Note that (although this exercise was highly artificial) this means that we obtained four distinct IPv4 headers in which 24 bits were dynamically set to precise values using just 3000 evaluations of an objective. Moreover, most elites had near neighbors in Hamming space. Below, we show the packet headers and correct checksums in hexadecimal format, with correct checksums shown in {\color{blue}blue}.

\begin{center}
    \texttt{450555F106FE248015005326DE91FAC443C1CFD0}
    \texttt{450555F106FE248015005326DE91FAC443C1CFD4}
    \texttt{450755F3925640800700FB259E91F8414B80CE54}
    \texttt{450755F3925640800700FBA59E99DA4143C0CED4}
    \texttt{450755F3925640800700FBA59E99DB4143C0CED4}
    \texttt{450FDCB3AAC7EB8202009BF4DE3BCAC343C1CFD0}
    \texttt{450FDEF3AAC66B8203007D66DE3BCAC063C1CDD0}
    \texttt{458455F106FA248015005324DE99EAC443C0CFD0}
    \texttt{458555F106FE248015005326DE91EAC043C0CFD0}
    \texttt{459DFA69A85A324C6B8B6866CFA398CD748D2C65}
    \texttt{459DFA69A85A324C6B8B7066CF2398CD748D2C65}
    \texttt{459DFA69A85A324C6B8B7866CFA398CD748D2C65}
    \texttt{459DFAC88879365C4B8B79D4CDA318CD74896E61}
    \texttt{459DFAC88879365C6B8B{\color{blue}E9D4}CFA338CD74892E69}
    \texttt{459FFAC8985936586B8A4967CFB0194D748D6E68}
    \texttt{45C3034F5DBADCD3C3552E0274983B1687DC196F}
    \texttt{45C3834F1DBA9CD3C355680F70983B1687DC196F}
    \texttt{45C3834F5DB29CD3C3552607709A3A1687DC196F}
    \texttt{45C3834F5DBA9CD3C3552C09709A3A16875C196F}
    \texttt{45C3834F5DBA9CD3C355{\color{blue}2C0F}70983B1687DC196F}
    \texttt{45C3834F5DBB9CD3C3550C0F709A3B1687DE196E}
    \texttt{45C383CB5DBA9CD3C3552C81709A3A16875C19EF}
    \texttt{45C38B5B5DBA9F92C35D13B0608A3A5F865D393B}
    \texttt{45C38B5F5DBA9F92C35D{\color{blue}12F0}618A3A1F865D393B}
    \texttt{45C38B5F5DBA9F92C35D1330608A3A5F865D393B}
    \texttt{45C38B5F5DBA9F92C35D{\color{blue}13B0}608A3A5F865D393B}
    \texttt{45C38BDF5DBA9F92C35D13B0608A3A5F865D393B}
    \texttt{45DA7BFA1A2158A1C3AA3A7095541C1440E90080}
    \texttt{45DA7BFA1A2158A9C3A259F0955C1C1440ED0080}
    \texttt{45EA3895E95A129D3850AC069E146A8932968AEE}
\end{center}

Similar exercises using the 32-bit cyclic redundancy check (CRC) \cite{koopman200232} used in (e.g.) the IEEE 802.3 standard for Ethernet and many other protocols/programs suggest that more effort is required. Indeed, using the 16-bit CRC corresponding to the polynomial $x^{16} + x^{15} + x^2 + 1$ and $g$ corresponding to uniformly flipping bits of the CRC along with 32 other bits in a notional Ethernet frame, we get only two of 19
elites with the correct CRC after 15000 evaluations of an objective that is the Hamming distance between the CRC and its nominal bits.
This is in line with the general observation that reversing CRCs is nontrivial, though readily accomplished with dedicated algorithms \cite{stigge2006reversing}.

\section{\label{sec:BugFix}Effect of a bug in previous version of code in \S \ref{sec:goExploreDissimiliarity.m}}

After the first version of this paper was prepared, we identified and repaired a bug in the code in \S \ref{sec:goExploreDissimiliarity.m}, then reran the examples. The original code was 
{\fontsize{6}{7}
\begin{verbatim}
        nearInd = unique([find(inCell(:)'),nearest(1:numNearest)]);
\end{verbatim}
} 

For consistency with Algorithm \ref{alg:GoExploreDissimilarity}, we replaced this with
{\fontsize{6}{7}
\begin{verbatim}
        % BUG FIXED: inBase was inCell
        nearInd = unique([find(inBase(:)'),nearest(1:numNearest)]);
\end{verbatim}
} 

The bug meant that some of the points used to interpolate the objective were always chosen from the inhabited cell with lexicographically least $\sigma^{(K)}$ instead of the current ``go-to'' cell. 

Fixing this bug had little quantitative effect, but suggested that the intended interpolation encourages ``drilling down'' into initial minima slightly more relative to initial exploration. This effect was small because the nearest points were still chosen for interpolation, and only the relatively few points in the base cell that were not sufficiently near were missed.

\section{\label{sec:SourceCode}Source code}

NB. The LaTeX source of a preprint version of this document contains scripts that use the code here to reproduce results/figures throughout.

\subsection{\label{sec:goExploreDissimiliarity.m}goExploreDissimiliarity.m}

{\fontsize{6}{7}
\begin{verbatim}
function [history,landmarks] = goExploreDissimilarity(f,dis,L,T,K,...
    globalGenerator,objectiveEvalBudget,isPosDef,localGenerator,...
    maxExploreEffort)

% Go-Explore instantiation serving as a quality-diversity algorithm (see
% https://quality-diversity.github.io/, and cf. multimodal optimation,
% which this can also do) using mangitude-based constructs on a
% dissimilarity space, and fundamentally using not much else besides
% mechanisms for "globally" and "locally" generating points.
%
% NB. This particular instantiation also leverages linear radial basis
% function interpolation and thus is "only" suitable for boxes in Euclidean
% space, lattices therein, and bitvectors (these last two with only
% slightly more involved exploration function handles). However, in general
% any other approach to predicting the objective f (e.g., a neural net)
% could be cleanly substituted in the code. Still more generally, a pure
% exploration mechanism based on magnitude alone could be instantiated in
% the event that objective prediction is impossible in a given situation.
%
% Inputs (with examples):
%     f,    function handle for objective, e.g. Rastrigin function:
%               f = @(x) 10*numel(x)+x(:)'*x(:)-10*sum(cos(2*pi*x(:)));
%     dis,  function handle for dissimilarity (assumed symmetric), e.g.
%               dis = @(x0,x1) vecnorm(x0-x1);  % L2
%           NB. If dis is insufficiently regular, it may be profitable to
%           redefine it using the regularizeMagnitudeDimension function.
%           For this reason, we don't assume dis is the L2 distance
%     L,    number of landmarks (<= T)
%     T,    number of states to generate (including landmarks, so >= L)
%     K,    rank cutoff <= L
%     globalGenerator,  
%           function handle for global state generator, e.g.
%               dim = 2;
%               lb = -2*ones(dim,1);    % lower bounds
%               ub = 3*ones(dim,1);     % upper bounds
%               globalGenerator = ...
%                   @() diag(ub-lb)*rand(dim,1)+diag(lb)*ones(dim,1);
%     objectiveEvalBudget, 
%           number of evaluations of f to perform
%     isPosDef,  
%           flag to set (as per MATLAB logic, to any nonzero number) if dis
%           is known a priori to be positive definite. As
%           https://doi.org/10.1515/9783110550832-005 points out, this is
%           the case whenever dis is (e.g.)
%                 * the usual $\ell^p$ (or even $L^p$) norm for 1 <= p <= 2
%                 * the usual metric on hyperbolic space
%                 * an ultrametric
%                 * the usual metric on a weighted tree
%           Setting this flag allows us to avoid computing spectra that
%           would otherwise be required to maintain the connection between
%           weightings and diversity-saturating distributions
%     localGenerator,  
%           function handle for exploration subroutine, e.g.
%                 localGenerator = @(x,theta) x+theta*randn(size(x));
%           where here theta plays the role of standard deviation, and more
%           generally is expected to be governed by some scalar
%           "bandwidth." See note below on doing this for ints and bools,
%           and also a comment in code containing the string <BANDWIDTH>
%     maxExploreEffort,  
%           number bounding exploration effort, as measured in number of
%           evaluations of f per "expedition"
%
% Outputs: 
%     history,	
%           struct with fields
%                 * state       i.e., an input to f
%                 * cell        the cell containing the state
%                 * birth       the "epoch" in which the state was "born"
%                 * reign       the last epoch in which the state was elite
%                 * objective   the function objective value
%     landmarks,    
%           cell (row) array of landmark states
%
% Example: the Rastrigin function
%     rng('default');
%     f = @(x) 10*numel(x)+x(:)'*x(:)-10*sum(cos(2*pi*x(:)));   % Rastrigin
%     dis = @(x0,x1) vecnorm(x0-x1);
%     L = 15;
%     T = ceil(L*log(L));
%     K = 2;
%     dim = 2;
%     lb = -2*ones(dim,1);    % lower bounds
%     ub = 3*ones(dim,1);     % upper bounds
%     globalGenerator = @() diag(ub-lb)*rand(dim,1)+diag(lb)*ones(dim,1);
%     localGenerator = @(x,theta) x+theta*randn(size(x));
%     M = 3e2;
%     isPosDef = 1;
%     maxExploreEffort = 128;
%     warning off;    % to keep partialCouponCollection from complaining
%     [history,landmarks] = goExploreDissimilarity(f,dis,L,T,K,...
%         globalGenerator,M,isPosDef,localGenerator,maxExploreEffort);
%     warning on;
%     % Plot
%     isCurrentElite = ([history.reign]==max([history.reign]));
%     elite = {history(isCurrentElite).state};
%     h_fig = figure;
%     hold on;
%     bar = cell2mat(landmarks);
%     ind = nchoosek(1:size(bar,2),size(bar,2)-(K-1));
%     for ell = 1:size(ind,1)
%         h_fig = voronoi(bar(1,ind(ell,:)),bar(2,ind(ell,:)));
%         for j = 1:numel(h_fig)
%             set(h_fig(j),'Color',[0,0,0]);
%         end
%     end
%     foo = cell2mat(elite);
%     scatter(foo(1,:),foo(2,:),20,...
%         'MarkerEdgeColor',[0,0,1],'MarkerFaceColor',[0,0,1]); 
%     bar = cell2mat({history.state});
%     plot(bar(1,:),bar(2,:),'r.','MarkerSize',1);
%     grid on;
%     daspect([1,1,1]);
%     axis([lb(1)-1,ub(1)+1,lb(2)-1,ub(2)+1]);
%
% Example 2: a variant on the above with integer-valued states. Modify the
% above example along the following lines:
%     scale = 100;
%     f = @(x) 10*numel(x)+x(:)'*x(:)/scale^2-10*sum(cos(2*pi*x(:)/scale));
%     lb = -2*ones(dim,1)*scale;    % lower bounds
%     ub = 3*ones(dim,1)*scale;     % upper bounds
%     globalGenerator = @() round(diag(ub-lb)*rand(dim,1)+diag(lb)*ones(dim,1));
%     rafz = @(x) ceil(abs(x)).*sign(x);   % round away from zero
%     localGenerator = @(x,theta) x+rafz(theta*randn(size(x)));
% Zooming into the resulting figure will show the discrete nature of this
% example.
%
% NB. Elaborating on example 2, for a local generator on lattices or
% bitvectors, it is reasonable to just round/truncate the arguments of a
% Gaussian. To see why, we recall an example from section 4.1 of
% https://doi.org/10.1137/18M1164937 (preprint at
% https://arxiv.org/abs/1801.02373):
%     sample = [1,0,1,-2,1,2,3,-2,1,-1]; mu = mean(sample);      % 0.4
%     Sigma = var(sample);    % 2.7111 = (1.6465)^2 
%     % The corresponding discrete Gaussian has the following parameters. 
%     % Note that solving the gradient system that gives u and B requires 
%     % evaluating (indeed, differentiating) a Riemann theta function, 
%     % which is probably infeasible for situations of practical interest 
%     % from our perspective. 
%     u = .023; B = .0587; % Taking an ample integer interval
%     n = -100:100; 
%     % Form the discrete Gaussian 
%     p_d = exp(2*pi*(-.5*B*n.^2+n*u)); 
%     p_d = p_d/sum(p_d); 
%     % Form the continuous Gaussian 
%     p_c = exp(-.5*(n-mu).^2/Sigma); 
%     p_c = p_c/sum(p_c); 
%     % Round the argument of the continuous Gaussian 
%     x = linspace(min(n),max(n),10*numel(n)); 
%     p_x = exp(-.5*(round(x)-mu).^2/Sigma); 
%     p_x = p_x/trapz(x,p_x); 
%     % Rounding samples gives a pretty decent approximation to the 
%     % discrete Gaussian here. However, in high dimensions this will be 
%     % awful: for, as https://doi.org/10.1145/2746539.2746606 points out, 
%     % sampling from a discrete Gaussian is hard enough that an efficient 
%     % algorithm breaks lattice-based cryptosystems. On the other hand, 
%     % this means that we shouldn't waste our time trying to sample from a
%     % discrete Gaussian in % the first place, and rounding is a decent 
%     % hack for our purposes. 
%     figure; plot(n,p_d,'ko',n,p_c,'rx',x,p_x,'b+'); xlim([-10,10]);
%
% Example 3: a Sherrington-Kirkpatrick spin glass, i.e., an optimization
% problem on bit vectors whose global minimum is NP-hard to compute (though
% approximable in quadratic time by https://doi.org/10.1137/20M132016X)
%     rng('default');
%     N = 20; J = triu(randn(N),1); J = (J+J')/sqrt(N);
%     f = @(bitColVector) (2*bitColVector(:)-1)'*J*(2*bitColVector(:)-1); % SK
%     dis = @(x0,x1) vecnorm(x0-x1);  % square root of Hamming distance
%     L = 10;
%     T = ceil(L*log(L));
%     K = 2;
%     dim = N;
%     lb = 0*ones(dim,1);     % lower bounds: Boolean
%     ub = 1*ones(dim,1);     % upper bounds: Boolean
%     globalGenerator = ...   % note round!
%         @() round(diag(ub-lb)*rand(dim,1)+diag(lb)*ones(dim,1));
%     localGenerator = ...
%         @(x,theta) double(xor(x~=0,rand(size(x))<theta));	% bit flips
%     M = 3e2;
%     isPosDef = 1;
%     maxExploreEffort = 128;
%     warning off;    % to keep partialCouponCollection from complaining
%     [history,landmarks] = goExploreDissimilarity(f,dis,L,T,K,...
%         globalGenerator,M,isPosDef,localGenerator,maxExploreEffort);
%     warning on;
% It is instructive to perform this example in conjunction with an analysis
% of barriers between minima of the spin glass. This particular example has
% about 1.04M states and 82 minima.
%
% NB. Most of the expensive operations in this code (function evaluations
% and dissimilarity evaluations, though not cutoff scales) could be easily
% performed in parallel with minor modifications. See in particular the
% code comments "*** CAN USE parfor HERE ***" which are not exhaustive.
%
% Last modified 20220624 by Steve Huntsman
%
% Copyright (c) 2022, Systems & Technology Research. All rights reserved.

%% Generate landmarks
[initialStates,landmarkInd,~] = generateLandmarks(globalGenerator,dis,L,T);
landmarks = initialStates(landmarkInd);

%% Assign states to cells
% Note that we are sort of overloading MATLAB terminology for cellArray:
% this is a byproduct of sticking with Go-Explore's "cell" terminology
cellArray = stateCell(dis,landmarks,K,initialStates);   % a matrix
% Get unique cells to encode a set per se of inhabited cells--all
% implicitly, to save the expense of working with the cell representation.
% Use 'stable' option just so we can append efficiently later if any future
% code refactoring suggests it
[~,~,cellNumber] = unique(cellArray,'rows','stable');

%% Initialize history with states used to generate landmarks
% The history (to be instantiated shortly) will detail all the states for
% which the objective is ever evaluated, their corresponding cells, the
% "epoch" in which they were "born" (i.e., the outer loop iteration during
% which the objective was evaluated on them), the last epoch in which they
% were elite (0 if never elite), and their objective values
%
% Note that we can use this information to (cheaply) reconstruct, e.g.,
%     * the number K of closest landmarks used to define cells: this is
%         size([history.cell],1)
%     * the set of inhabited cells: up to ordering/transpose, this is
%         [history([history.reign]==max([history.reign])).cell]
epoch = 1;
% Objective *** CAN USE parfor HERE ***
objective = nan(size(initialStates)); 
for j = 1:numel(objective)
    objective(j) = f(initialStates{j});
end
% Reign
reign = zeros(size(initialStates));
for j = 1:max(cellNumber)
    inCell = find(cellNumber==j);  
    [~,indArgmin] = min(objective(inCell));
    reign(inCell(indArgmin)) = epoch;
end
% The instantiation of history as a struct has the practical benefit
% that--with the possible exception of states, which might not be (column)
% vectors in general, but often will be in practice--all of the fields of
% history can be realized as matrices a la [history.cell], etc. In the
% general case, even the states can be realized as cell arrays using
% braces
initialCells = ...
    mat2cell(cellArray',size(cellArray,2),ones(1,size(cellArray,1)));
initialBirth = num2cell(epoch*ones(size(initialStates)));
initialReign = num2cell(reign);
initialObjective = num2cell(objective);
initialHistory = [initialStates;initialCells;initialBirth;...
    initialReign;initialObjective];
history = cell2struct(initialHistory,...
    ["state","cell","birth","reign","objective"],1);

%% Main loop
objectiveEvalCount = numel(history);
% Use objectiveEvalCount in lieu of numel(history) per se in order to
% better respect the evaluation budget
while objectiveEvalCount < objectiveEvalBudget
        
    %% Identify elites (a/k/a the "archive" in Go-Explore paper-speak)
    % We keep the entire history to enhance exploration and in mind of the
    % expense of evaluating objectives for our applications
    isElite = [history.reign]==epoch;
    elite = {history(isElite).state};

    %% Begin a new epoch
    epoch = epoch+1;

    %% <GO>
    
    %% Compute dissimilarity matrix for elites
    dissimilarityMatrix = zeros(numel(elite));
    % Symmetry of dis is assumed (as otherwise the connection between
    % diversity saturation and weightings breaks down); hence symmetry is
    % built into dissimilarityMatrix
    for j1 = 1:numel(elite)
        for j2 = (j1+1):numel(elite)
            dissimilarityMatrix(j1,j2) = dis(elite{j1},elite{j2});
        end
    end
    dissimilarityMatrix = max(dissimilarityMatrix,dissimilarityMatrix');
    
    %% Compute diversity-maximizing distribution on elites    
    % Compute an appropriate cutoff scale to ensure a bona fide
    % diversity-maximizing distribution on elites
    if isPosDef
        % Save lots of time--avoid computing spectra over and over
        t = posCutoff(dissimilarityMatrix)*(1+sqrt(eps));
    else
        t = strongCutoff(dissimilarityMatrix)*(1+sqrt(eps));
    end
    Z = exp(-t*dissimilarityMatrix);
    % Solve for weighting while handling degeneracies 
    if max(abs(Z-ones(size(Z))),[],'all') < eps^.75 % sqrt(eps) too big
        w = ones(size(Z,1),1);
    else
        w = Z\ones(size(Z,1),1);
    end
    if any(w<0)
        warning(['min(w) = ',num2str(min(w)),' < 0: adjusting post hoc']); 
        w = w-min(w); 
    end
    maxDiversityPdf = w/sum(w);
    
    %% Construct "go distribution" goPdf balancing diversity and objective
    % The improvement of the objective values should drive progress instead
    % of (e.g.) a regularization coefficient a la temperature in simulated
    % annealing: i.e., this is only implicitly dynamic
    eliteObjectiveValue = [history(isElite).objective];
    % Match top and middle quantiles of logarithm of maximum diversity PDF
    % and elite objective values en route to producing the "go
    % distribution." Note that matching ranges or moments is complicated by
    % the fact that maxDivPdf has one entry that is basically zero, so its
    % logarithm is hard to normalize otherwise
    denom = max(log(maxDiversityPdf))-median(log(maxDiversityPdf));
    denom = denom+(denom==0);
    diversityTerm = (log(maxDiversityPdf)-median(log(maxDiversityPdf)))/...
        denom;
    denom = max(eliteObjectiveValue)-median(eliteObjectiveValue);
    denom = denom+(denom==0);
    objectiveTerm = (eliteObjectiveValue-median(eliteObjectiveValue))/...
        denom;
    % Encourage high diversity contributions and/or low objective values
    goPdf = exp(diversityTerm(:)-objectiveTerm(:));
    if numel(goPdf) == 1, goPdf = 1; end	% preclude possibility of a NaN 
    goPdf = goPdf/sum(goPdf);	% get a bona fide PDF
    goCdf = cumsum(goPdf);      % CDF for easy sampling
    
    %% Determine "go effort" expeditionsThisEpoch of sampling from goPdf    
    % Use lower bound for collecting half of the coupons from goPdf. The
    % rationale here is to systematically avoid dedicating lots of effort
    % to cells that aren't promising, but to have confidence that many
    % cells will still be explored
    [~,lowerBound,~] = partialCouponCollection(goPdf,...
        ceil(numel(goPdf)/2),numel(goPdf));
    expeditionsThisEpoch = ceil(lowerBound);
    disp(['objectiveEvalCount = ',num2str(objectiveEvalCount),...
        '; expeditionsThisEpoch = ',num2str(expeditionsThisEpoch),...
        '; objectiveEvalBudget = ',num2str(objectiveEvalBudget)]);
    
    %% </GO>
    
    %% Memorialize extrema of objective for normalization in loop below
    globalMax = max([history.objective]);
    globalMin = min([history.objective]);

    %% Inner loop over expeditions
    newState = [];
    for expedition = 1:expeditionsThisEpoch
        
        %% Sample from goCdf to get a "base elite" from which to explore
        % Recall that entries of goCdf (are presumed to) correspond to
        % inhabited cells
        baseIndex = find(rand<goCdf,1,'first');
        
        %% <EXPLORE>
        
        %% Determine exploration effort (from recent improvement pattern)
        % Get history of the base cell
        baseCell = stateCell(dis,landmarks,K,elite(baseIndex));
        inBase = find(ismember([history.cell]',baseCell,'rows'));        
        % Determine expeditions that visit the base cell (easy), rather
        % than expeditions that start from the base cell (hard).
        %
        % Note that our method of recording history (viz., state, cell,
        % birth, reign, and objective) makes it impossible to exactly
        % reconstruct the itinerary of expeditions (i.e., which cells
        % hosted bases during a given epoch). While we could guess at
        % this--or, more sensibly, augment our records to know this--it's
        % not clear that this would actually be useful, and it would
        % increase the complexity of the code. So we avoid this
        baseEpoch = unique([history(inBase).birth]);
        % In base cell from last expedition visiting it
        oneAgo = find([history(inBase).birth]==baseEpoch(end));
        % In base cell from penultimate expedition visiting it
        if numel(oneAgo) && numel(baseEpoch)>1
            twoAgo = find([history(inBase).birth]==baseEpoch(end-1));
        else
            twoAgo = oneAgo;	% could be empty in principle
        end
        % Determine normalized differential in objective over last two
        % expeditions visiting the base cell
        oneAgoObjective = [history(inBase(oneAgo)).objective];
        twoAgoObjective = [history(inBase(twoAgo)).objective];
        denom = globalMax-globalMin;
        denom = denom+(denom==0);
        bestOneAgoNormed = min((oneAgoObjective-globalMin)/denom);
        bestTwoAgoNormed = min((twoAgoObjective-globalMin)/denom);
        baseDelta = bestOneAgoNormed-bestTwoAgoNormed;  % in [-1,1]
        % Determine exploration effort on the basis of prior efforts
        if epoch > 2
            priorExploreEffort = ...
                nnz([history(inBase).birth]==baseEpoch(end));
        else
            % Geometric mean of 1 and maxExploreEffort seems appropriate
            priorExploreEffort = ceil(sqrt(maxExploreEffort));
        end
        foo = max(priorExploreEffort*2^-baseDelta,1);
        exploreEffort = ceil(min(foo,maxExploreEffort));

        %% Use nearby points and those in cell to interpolate f
        % Worry later about being efficient
        %
        % Rank states in the history by dissimilarity to the base elite
        numNearest = min(numel(history),ceil(maxExploreEffort/2));
        baseDissimilarity = nan(1,numel(history));
        for j = 1:numel(history)
            baseDissimilarity(j) = dis(elite{baseIndex},history(j).state);
        end
        [~,nearest] = sort(baseDissimilarity);
        % Combine the nearest states with those in the current cell and
        % do radial basis function interpolation with a linear kernel
        % BUG FIXED: inBase was inCell
        nearInd = unique([find(inBase(:)'),nearest(1:numNearest)]);
        x = [history(nearInd).state];
        y = [history(nearInd).objective];
        interpolant = linearRbfInterpolation(x,y);  
        
        %% Probe, iteratively reducing bandwidth as appropriate
        numProbes = 2*maxExploreEffort;
        probe = cell(1,numProbes);
        % Initial bandwidth is big--we will bring it down to size. Note
        % that this works just fine for Booleans and Hamming distance (or
        % its ilk) along with the localGenerator function of Example 3
        % above--more generally, one can generally retool the
        % localGenerator function to play nicely with this initialization.
        % <BANDWIDTH>
        bandwidth = max(dissimilarityMatrix(baseIndex,:));
        for j = 1:numel(probe)
            % In principle we could make the localGenerator function
            % interact with available information--for example, taking the
            % in-cell and/or nearby history with objective values below
            % some quantile, getting the mean and covariance of that data,
            % and sampling from a Gaussian with those same parameters. But
            % we aren't doing that now/yet, mainly because this
            % overspecializes.
            probe{j} = localGenerator(elite{baseIndex},bandwidth);
        end
        % Compare probe cells with base cell
        probeCellArray = stateCell(dis,landmarks,K,probe);   % a matrix
        probeInCell = ismember(probeCellArray,baseCell,'rows');
        % Reduce bandwidth until a significant number of probes remain in
        % the current cell (cf. Gaussian annulus theorem)
        while nnz(probeInCell) < numProbes/4
            bandwidth = bandwidth/2;
            for j = 1:numel(probe)
                probe{j} = localGenerator(elite{baseIndex},bandwidth);
            end
            probeCellArray = stateCell(dis,landmarks,K,probe);   % a matrix
            probeInCell = ismember(probeCellArray,baseCell,'rows');
        end
        % Out of an abundance of caution, avoid probe collisions (this has
        % never happened in practice if problems weren't absurdly small or
        % localGenerator functions weren't poorly constructed).
        uniqueProbe = unique(cell2mat(probe)','rows')';
        uniqueProbe = setdiff(uniqueProbe',...
            [history(nearInd).state]','rows')';
        probe = mat2cell(uniqueProbe,size(uniqueProbe,1),...
            ones(1,size(uniqueProbe,2)));
        
        %% Get biobjective data: estimated objective & weighting component
        % A local dissimilarity matrix considers all the probes at once.
        % One can easily imagine wanting to compute differential magnitudes
        % for each probe individually, but this has disadvantages. Setting
        % aside (mostly irrelevant) concerns about computational
        % complexity, the probes must eventually treated as a batch in
        % order to perform the biobjective analysis below.
        local = [{history(nearInd).state},probe];
        dissimilarityMatrix_local = zeros(numel(local));
        for j1 = 1:numel(local)
            for j2 = 1:numel(local)
                dissimilarityMatrix_local(j1,j2) = dis(local{j1},local{j2});
            end
        end
        % Compute an appropriate cutoff scale to ensure a bona fide
        % diversity-maximizing distribution on elites
        if isPosDef
            % Save time--avoid computing spectra over and over
            t_local = posCutoff(dissimilarityMatrix_local)*(1+sqrt(eps));
        else
            t_local = strongCutoff(dissimilarityMatrix_local)*(1+sqrt(eps));
        end
        % Local similarity matrix and weighting: latter is proportional to
        % diversity-maximizing distribution
        Z_local = exp(-t_local*dissimilarityMatrix_local);
        w_local = Z_local\ones(size(Z_local,1),1);
        % Estimate of objective
        f_estimate = [y,cellfun(interpolant,probe)];
        %
        biObjective = [f_estimate',-w_local]';

        %% Determine quantitative Pareto dominance
        % We will select the least-dominated points after normalizing both
        % objectives to zero mean and unit variance a la
        biObjectiveNormalized = ...
            diag(var(biObjective,0,2))\(biObjective-mean(biObjective,2));
        % we want to rank points by their Pareto domination a la
        dominatedBy = nan(size(biObjectiveNormalized,2),1);
        for j = 1:numel(dominatedBy)
            % This is slicker than repmat
            obj_j = biObjectiveNormalized(:,j)...
                *ones(1,size(biObjectiveNormalized,2));
            dominator = min(obj_j-biObjectiveNormalized);
            dominatedBy(j) = max(dominator);
        end

        %% Determine new states to evaluate after this loop
        cutoff = numel(local)-numel(probe);
        % 1:cutoff corresponds to stuff already in history        
        [~,dominanceInd] = sort(dominatedBy((cutoff+1):end));
        % Taking a minimum here to prevent an out-of-bounds error was
        % necessary on rare occasion (precisely once in dozens of uses
        % before instituted), for reasons that aren't immediately obvious:
        % no effort has yet been made to understand this
        toEvaluate = ...    
            probe(dominanceInd(1:min(exploreEffort,numel(dominanceInd))));
        % Update in a way that lets us terminate on budget
        newState = [newState,toEvaluate]; %#ok<AGROW>
        objectiveEvalCount = objectiveEvalCount+numel(toEvaluate);
        
        %% Respect the evaluation bound
        if objectiveEvalCount > objectiveEvalBudget
            objectiveEvalExcess = objectiveEvalCount-objectiveEvalBudget;
            newState = newState(1:(numel(newState)-objectiveEvalExcess));
            break; 
        end
        
        %% </EXPLORE>
    end
       
    %% Assign states to cells (cf. similar code @ initialization)
    state = [{history.state},newState];
    cellArray = stateCell(dis,landmarks,K,state);   % a matrix
    [~,~,cellNumber] = unique(cellArray,'rows','stable');

    %% Prepare rest of history updates (cf. similar code @ initialization)
    birth = [[history.birth],epoch*ones(size(newState))];
    % Objective *** CAN USE parfor HERE ***
    objective = nan(size(newState));
    for j = 1:numel(objective)
        objective(j) = f(newState{j});
    end
    objective = [[history.objective],objective]; %#ok<AGROW>
    % Reign
    reign = [[history.reign],zeros(size(newState))];
    for j = 1:max(cellNumber)
        inCell = find(cellNumber==j);
        [~,indArgmin] = min(objective(inCell));
        reign(inCell(indArgmin)) = epoch;        
    end
    
    %% Update history
    newState = state;
    newCell = mat2cell(cellArray',...
        size(cellArray,2),ones(1,size(cellArray,1)));   % a cell
    newBirth = num2cell(birth);
    newReign = num2cell(reign);
    newObjective = num2cell(objective);
    newHistory = [newState;newCell;newBirth;newReign;newObjective];
    history = cell2struct(newHistory,...
        ["state","cell","birth","reign","objective"],1);

end
\end{verbatim}
}

\subsection{stateCell.m}

{\fontsize{6}{7}
\begin{verbatim}
function cellId = stateCell(dis,landmarks,K,x)

% Produce cell identifiers for states
%
% Inputs:
%     dis,  function handle for dissimilarity (assumed symmetric)
%     landmarks,  
%           cell array of landmarks (versus a larger cell array of states 
%           and landmark indices)
%     K,    rank cutoff <= numel(lan) = L
%     x,    state(s) to map to cell
%
% Output: 
%     cellId,	array of size [numel(x),K] whose ith row is a sorted tuple
%               of the K closest landmarks (as indices, i.e., a subindex of
%               landmark indices)
%
% NB. To work with states and landmarkIndex as produced by
%     generateLandmarks, simply preclude executing this function by the
%     assignment
%         landmarks = states(landmarkIndex);
%
% Last modified 20220425 by Steve Huntsman
%
% Copyright (c) 2022, Systems & Technology Research. All rights reserved.

%% Check scalar input
% Other inputs are too tricky to check meaningfully here
if ~isscalar(K), error('K not scalar'); end
if ~isfinite(K), error('K not finite'); end
if ~isreal(K), error('K not real'); end
if K ~= round(K), error('K not integral'); end
if K < 1, error('K < 1'); end
L = numel(landmarks);
if K > L, error('K > L'); end

%%
dissimilarityFromLandmarks = nan(numel(x),L);
for i = 1:numel(x)
    for ell = 1:L
        dissimilarityFromLandmarks(i,ell) = dis(x{i},landmarks{ell});
    end
end
[~,ind] = sort(dissimilarityFromLandmarks,2);
cellId = ind(:,1:K);
\end{verbatim}
}

\subsection{generateLandmarks.m}

{\fontsize{6}{7}
\begin{verbatim}
function [initialStates,landmarkIndex,magnitude] = ...
    generateLandmarks(globalGenerator,dis,L,T)

% Generate diverse landmark states/points (we use the term state by
% reference to a dissimilarity framework for Go-Explore-type algorithms,
% which this was developed for)
%
% Inputs:
%     globalGenerator,  
%           function handle for global state generator
%     dis,  function handle for dissimilarity (assumed symmetric)
%     L,    number of landmarks (<= T)
%     T,    number of points to generate (including landmarks, so >= L)
%
% Output: 
%     initialStates,    cell array of all states produced by
%                       globalGenerator (for reuse) 
%     landmarkIndex,    indices of states that are landmarks 
%     magnitude,        magnitude of landmarks after given generations
%
% Example:
%     rng('default');
%     dim = 2;
%     lb = zeros(1,dim);
%     ub = ones(1,dim);
%     globalGenerator = @() diag(ub-lb)*rand(dim,1)+diag(lb)*ones(dim,1);
%     dis = @(x0,x1) vecnorm(x0-x1);
%     L = 15;
%     T = L^2;
%     [initialStates,landmarkInd,magnitude] = ...
%         generateLandmarks(globalGenerator,dis,L,T);
%     landmarks = initialStates(landmarkIndex);
%     foo = cell2mat(landmarks);
%     figure; plot(foo(1,:),foo(2,:),'ko')
%     figure; plot(magnitude)
%
% Last modified 20220425 by Steve Huntsman
%
% Copyright (c) 2022, Systems & Technology Research. All rights reserved.

%% Check scalar inputs
% Function handles are too tricky to check meaningfully
if ~isscalar(L), error('L not scalar'); end
if ~isfinite(L), error('L not finite'); end
if ~isreal(L), error('L not real'); end
if L ~= round(L), error('L not integral'); end
if L < 1, error('L < 1'); end
if ~isscalar(T), error('T not scalar'); end
if ~isfinite(T), error('T not finite'); end
if ~isreal(T), error('T not real'); end
if T ~= round(T), error('T not integral'); end
if T < L, error('T < L'); end

%% Initial states and landmark indices
initialStates = cell(1,T);
for ell = 1:L
    initialStates{ell} = globalGenerator();
end
landmarkIndex = 1:L;

%% Initial dissimilarity matrix, scale, weighting, etc.
d0 = zeros(L);
for j = 2:L
    for k = 1:(j-1)
        % There is some superfluous indexing here for illustration
        d0(j,k) = dis(initialStates{landmarkIndex(j)},...
            initialStates{landmarkIndex(k)});
    end
end
d0 = max(d0,d0');
% The t = 0 limit is not worth considering here, as it yields either unit
% magnitude or naughty behavior. For the sake of generality (and because
% it's a one-time cost) we invoke the strong cutoff scale.
t = strongCutoff(d0)*(1+sqrt(eps));
Z0 = exp(-t*d0);
w0 = Z0\ones(size(Z0,1),1);
magnitude = nan(1,T);
magnitude(L) = sum(w0);

%% Main loop
for i = (L+1):T
    %% Propose new state to replace the one with least weighting component
    [~,ind] = min(w0);
    newState = globalGenerator();
    
    %% Store new state
    initialStates{i} = newState;
    
    %% Gauge impact on magnitude at original (strong cutoff) scale
    newRow = zeros(1,L);
    for ell = 1:L
        if ell ~= ind
            newRow(ell) = dis(newState,initialStates{landmarkIndex(ell)});
        end
    end
    d1 = d0;
    d1(ind,:) = newRow;
    d1(:,ind) = newRow';
    Z1 = exp(-t*d1);
    w1 = Z1\ones(size(Z1,1),1);
    
    %% Update
    if sum(w1) > magnitude(i-1)
        d0 = d1;
        w0 = w1;
        landmarkIndex(ind) = i;
        magnitude(i) = sum(w1);
    else
        magnitude(i) = magnitude(i-1);
    end
end
\end{verbatim}
}

\subsection{linearRbfInterpolation.m}

{\fontsize{6}{7}
\begin{verbatim}
function interpolant = linearRbfInterpolation(x,y)

% Radial basis function (RBF) interpolation using linear function (which is
% also a polyharmonic spline with exponent 1). Returns the interpolant (as
% a function handle) that fits data a la
%     interpolant(x(:,j)) = y(j).
% This choice of RBF avoids the need for scaling, and in turn wrangling
% with conditioning or fancy arithmetic, and so can be implemented very
% simply. While it precludes exploiting sparsity, our intended applications
% don't presently leverage this anyway.
%
% Cf. polyharmonicRbfInterpolation.m, which generalizes this but whose
% output runs considerably slower even in the case k = 1 (NOPs take time).
%
% Example:
%     rng('default');
%     x = rand(1,10); 
%     y = rand(size(x,2),1);
%     lri = linearRbfInterpolation(x,y);
%     foo = linspace(min(x),max(x),1e4); 
%     bar = arrayfun(lri,foo); 
%     figure; 
%     plot(x,y,'ko',foo,bar,'r');
% 
% 2D example:
%     rng('default');
%     x = rand(2,10);
%     y = rand(size(x,2),1);
%     lri = linearRbfInterpolation(x,y);
%     [u1,u2] = meshgrid(linspace(0,1,1e2));
%     v = nan(size(u1));
%     for j1 = 1:size(v,1)
%         for j2 = 1:size(v,2)
%             v(j1,j2) = lri([u1(j1,j2);u2(j1,j2)]);
%         end
%     end
%     figure; 
%     surf(u1,u2,v); 
%     shading flat;
%     hold on;
%     plot3(x(1,:),x(2,:),y,'ko');
%
% Last modified 20220429 by Steve Huntsman
%
% Copyright (c) 2022, Systems & Technology Research. All rights reserved.

%% Check x and y
if ~ismatrix(x), error('x not matrix'); end
if ~ismatrix(y), error('y not matrix'); end
if size(x,2) ~= numel(y), error('x and y sizes incompatible'); end
if any(~isreal(x(:))), error('x not real'); end
if any(~isreal(y(:))), error('y not real'); end
if any(~isfinite(x(:))), error('x not finite'); end
if any(~isfinite(y(:))), error('y not finite'); end

%% Coefficients for RBF interpolation
interpMatrix = squareform(pdist(x'));
interpCoefficient = interpMatrix\y(:);

%% Interpolant (as function handle: q = query point)
interpolant = @(q) vecnorm(x-q*ones(1,size(x,2)),2,1)*interpCoefficient;\end{verbatim}
}

\subsection{posCutoff.m}

{\fontsize{6}{7}
\begin{verbatim}
function t = posCutoff(d)

% Minimal t such that exp(-u*d) admits a nonnegative weighting for any u >
% t. Here d is an extended real matrix with zero diagonal. 
%
% Last modified 20210224 by Steve Huntsman
%
% Copyright (c) 2021, Systems & Technology Research. All rights reserved.

%% Check d matrix, square, nonnegative extended real, zero diag
if ~ismatrix(d), error('d not matrix'); end
[m,n] = size(d);
if m ~= n, error('d not square'); end
isExtendedReal = (isinf(d)|isfinite(d))&isreal(d);
if any(any(~isExtendedReal|d<0)), error('d not nonnegative extended real'); end
if any(diag(d)~=0), error('d diagonal not zero'); end

%%
t = log(n-1)/min(min(d+diag(inf(1,n))));
lower = 0;
upper = t;
while 1-lower/upper>sqrt(eps)
    t = (lower+upper)/2;
    Z = exp(-t*d);
    w = Z\ones(size(Z,1),1);
    if all(w>0)
        upper = t;
    else
        lower = t;
    end
end
\end{verbatim}
}

\subsection{strongCutoff.m}

{\fontsize{6}{7}
\begin{verbatim}
function t = strongCutoff(d)

% Minimal t such that exp(-u*d) is positive semidefinite and admits a
% nonnegative weighting for any u > t Here d is a symmetric extended real
% matrix with zero diagonal.
%
% Last modified 20210319 by Steve Huntsman
%
% Copyright (c) 2021, Systems & Technology Research. All rights reserved.

%% Check d matrix, square, nonnegative extended real, zero diag
if ~ismatrix(d), error('d not matrix'); end
[m,n] = size(d);
if m ~= n, error('d not square'); end
isExtendedReal = (isinf(d)|isfinite(d))&isreal(d);
if any(any(~isExtendedReal|d<0)), error('d not nonnegative extended real'); end
if any(diag(d)~=0), error('d diagonal not zero'); end
if max(max(abs(d'./d-1))) > sqrt(eps), error('d not symmetric'); end

%%
t = log(n-1)/min(min(d+diag(inf(1,n))));
lower = 0;
upper = t;
while 1-lower/upper>sqrt(eps)
    t = (lower+upper)/2;
    Z = exp(-t*d);
    spec = eig(Z);
    if min(spec)>=0
        w = Z\ones(size(Z,1),1);
        if all(w>0)
            upper = t;
        else
            lower = t;
        end
    else
        lower = t;
    end
end
\end{verbatim}
}

\subsection{partialCouponCollection.m}

{\fontsize{6}{7}
\begin{verbatim}
function [exact,lowerBound,upperBound] = partialCouponCollection(p,m,c)

% Calculation of the expected time for the general coupon collection
% problem in which the ith kind of coupon is drawn with probability p(i),
% and m is the number of coupon types to collect. Using p = ones(1,m)/m
% thus recovers the classical result for the uniform case. The exact result
% is computed when reasonably cheap (for details, see Corollary 4.2 of
% https://doi.org/10.1016/0166-218X(92)90177-C; for the case m = numel(p),
% see Theorem 4.1), and bounds computed more generally based on the same
% approach (both for rapidly decreasing and approximately uniform p).
%
% Inputs:
%     p,      probability distribution
%     m,      number of coupons to collect
%     c,      cutoff parameter for computations (will compute 2^c terms
%             provided that c <= 16: otherwise, this will be avoided--this
%             can speed things up dramatically)
%
% Outputs:
%     exact expected time (= NaN if unknown) and lower/upper bounds
%
% For evaluation purposes it is useful to avoid taking exact results as
% bounds; see code cells starting with
%         %% OMIT THIS FOR EVALUATION PURPOSES
% and comment them out as warranted.
%
% By way of (internal) documentation, see the appended LaTeX snippet.
%
% NB. Note that gammaln is used instead of nchoosek except when actually
% producing (vs counting) the combinations. This approach is MUCH faster. 
%
% NB. Subsumes couponCollectionTotal.m and generalCouponCollection.m via
% special cases--these are thus deprecated.
%
% NB. To get just the lowerBound output argument anonymously, use something
% like this (for the second output argument):
% 
%     function nthArg = outargn(fun,n)%#ok<STOUT,INUSL> 
% 
%     % Return nth output argument from function fun. If [y_1,...,y_n,...,z]
%     % = fun(inputs) then y_n = outargn(@() fun(inputs),n). I think this is
%     % a tolerable use of eval.
% 
%     eval(['[',repmat('~,',1,n-1),'nthArg] = fun();']);
%
% Last modified 20220502 by Steve Huntsman
%
% Copyright (c) 2022, Systems & Technology Research. All rights reserved.

%% Check p matrix, finite, real, nonnegative, sums to unity
if ~ismatrix(p), error('p not matrix'); end
isReal = isfinite(p)&isreal(p);
if ~all(isReal(:)), error('p not real'); end
if any(p<0), error('p not nonnegative'); end
if abs(sum(p)-1) > sqrt(eps)
    warning('p does not sum to unity: normalizing');
end
p = p(:)'/sum(p);

%% Check m
if ~isscalar(m), error('m not scalar'); end
if ~isfinite(m), error('m not finite'); end
if ~isreal(m), error('m not real'); end
if m ~= round(m)
    error('m not integral');
end
if m < 1, error('m < 1'); end

%% Check c
if ~isscalar(c), error('c not scalar'); end
if ~isfinite(c), error('c not finite'); end
if ~isreal(c), error('c not real'); end
if c ~= round(c)
    error('c not integral');
end
if c < 1, error('c < 1'); end
if c > 16
    avoid_c = 1;
    warning('c too big to quickly compute corresponding bounds: avoiding');
else
    avoid_c = 0;
end

%% Check number of coupon types is achievable (note restriction to support)
if m > nnz(p), error('numCouponTypes > nnz(p)'); end

%% Annoying corner case
if m == 1
    warning('m = 1');
    exact = 1;
    lowerBound = 1;
    upperBound = 1;
    return;
end

%%
p = sort(p,'descend');
n = numel(p);
exact = NaN;

%% [EXACT] Produce exact result for both bounds if easy, else m = n bound
if n <= 16
    exact = 0;
    for ell = 0:(m-1)
        ind = nchoosek(1:n,ell);
        pInd = reshape(p(ind),size(ind));
        sign = (-1)^(m-1-ell);
        S = sum(1./(1-sum(pInd,2)));
        exact = exact+sign*nchoosek(n-ell-1,n-m)*S;
    end
    %% OMIT THIS FOR EVALUATION PURPOSES
    lowerBound = exact;
    upperBound = exact;
    return;
end

%% [TOTAL] Get "total" result for m = n, which is also an upper bound
% Theorem 4.1 of https://doi.org/10.1016/0166-218X(92)90177-C yields
% expectation for total coupon collection as an integral that admits
% straightforward numerical computation
% disp('Computing for case m = n');
f = @(t) 1-prod(1-exp(-p(:)*t),1);
x = 1; while f(x) > eps, x = 10*x; end  % OK upper limit for integral
t = linspace(0,x,1e4);
total = trapz(t,f(t));

%% Initialize bounds
lowerBound = 0;
upperBound = total;

%% Return total result if m = n
if m == n
    exact = total;
    %% OMIT THIS FOR EVALUATION PURPOSES
    lowerBound = exact;
    upperBound = exact;
    return;
end

%% Now we'll have to compute less trivial bounds (c permitting)...
% See notes 
if ~avoid_c
    %% Form power set of 1:c
    % disp('Generating power set of 1:c');
    powerSet = false(2^c,c);
    for j = 1:c
        powerSet(:,j) = bitget((0:(2^c-1))',j);
    end
    %% Bounds assuming rapid decay of p
    % Based on exact result of Corollary 4.2 in
    % https://doi.org/10.1016/0166-218X(92)90177-C
    % disp('Computing bounds for rapidly decaying p');
    lb0 = 0;
    ub0 = 0;
    for ell = 0:(m-1)
        % disp(['    ',num2str(ell+1),'/',num2str(m)]);
        lambda = min(c,ell);
        localPowerSet = powerSet(1:2^lambda,1:lambda);
        %%
        sum_ell_lb = 0;
        sum_ell_ub = 0;
        for j = 1:size(localPowerSet,1)
            M = localPowerSet(j,:);
            P_M = sum(p(M));
            mu = nnz(M);
            ind = 1:(ell-mu);
            if ell-mu >= 0 && ell-mu <= n-lambda
                % Indices for ell-mu smallest and largest components of p
                % that aren't already reserved for P_M
                ind_lb = ind+n-(ell-mu);
                ind_ub = ind+min(lambda,n-(ell-mu));
                % numer = nchoosek(n-lambda,ell-mu);
                % Much faster to use gammaln than nchoosek here; no warnings
                numer = exp(gammaln((n-lambda)+1)-gammaln((ell-mu)+1)...
                    -gammaln((n-lambda)-(ell-mu)+1));
                sum_ell_lb = sum_ell_lb+numer/(1-P_M-sum(p(ind_lb)));
                sum_ell_ub = sum_ell_ub+numer/(1-P_M-sum(p(ind_ub)));
            end
        end
        %%
        % coeff = nchoosek(n-ell-1,n-m)*(-1)^(m-1-ell);
        % Much faster to use gammaln than nchoosek here; no warnings
        foo = exp(gammaln((n-ell-1)+1)-gammaln((n-m)+1)...
            -gammaln((n-ell-1)-(n-m)+1));
        coeff = foo*(-1)^(m-1-ell);
        lb0 = lb0+min(coeff*sum_ell_lb,coeff*sum_ell_ub);
        ub0 = ub0+max(coeff*sum_ell_lb,coeff*sum_ell_ub);
    end
    lowerBound = max(lowerBound,lb0);
    upperBound = min(upperBound,ub0);
    %% Bounds assuming near-uniformity of p
    % Based on exact result of Corollary 4.2 in
    % https://doi.org/10.1016/0166-218X(92)90177-C. Similar mechanically to
    % preceding stuff.
    % disp('Computing bounds for nearly uniform p');
    lbu = 0;
    ubu = 0;
    delta = n*p-1;  % deviation from uniformity
    for ell = 0:(m-1)
        % disp(['    ',num2str(ell+1),'/',num2str(m)]);
        lambda = min(c,ell);
        localPowerSet = powerSet(1:2^lambda,1:lambda);
        %%
        sum_ell_lbu = 0;
        sum_ell_ubu = 0;
        for j = 1:size(localPowerSet,1)
            M = localPowerSet(j,:);
            Delta_M = sum(p(M));
            mu = nnz(M);
            % Indices for ell-mu largest magnitude deviations that aren't
            % already reserved for Delta_M
            ind = (1:(ell-mu))+min(lambda,n-(ell-mu));
            if ell-mu >= 0 && ell-mu <= n-lambda
                denomL = 1-(ell+Delta_M+sum(abs(delta(ind))))/n;
                denomU = 1-(ell+Delta_M-sum(abs(delta(ind))))/n;
                denomL = max(denomL,eps);   % to be safe
                denomU = max(denomU,eps);   % to be safe
                % numer = nchoosek(n-lambda,ell-mu);
                % Much faster to use gammaln than nchoosek here; no warnings
                numer = exp(gammaln((n-lambda)+1)-gammaln((ell-mu)+1)...
                    -gammaln((n-lambda)-(ell-mu)+1));
                sum_ell_lbu = sum_ell_lbu+numer/denomL;
                sum_ell_ubu = sum_ell_ubu+numer/denomU;
            end
        end
        %%
        % coeff = nchoosek(n-ell-1,n-m)*(-1)^(m-1-ell);
        % Much faster to use gammaln than nchoosek here; no warnings
        foo = exp(gammaln((n-ell-1)+1)-gammaln((n-m)+1)...
            -gammaln((n-ell-1)-(n-m)+1));
        coeff = foo*(-1)^(m-1-ell);
        lbu = lbu+min(coeff*sum_ell_lbu,coeff*sum_ell_ubu);
        ubu = ubu+max(coeff*sum_ell_lbu,coeff*sum_ell_ubu);
    end
    lowerBound = max(lowerBound,lbu);
    upperBound = min(upperBound,ubu);
end

%% Augment lower bound using uniform case
% This is an easy calculation from the Corollary 4.2 cited above: the fact
% that the uniform case provides a lower bound is both intuitively obvious
% and proved in https://doi.org/10.1239/jap/1437658606.
%
% (NB. It is elementary to show that the gradient of the expectation w/r/t
% coupon probabilities is zero at uniformity, and similarly that the
% Hessian is diagonal there [note that this tactic is not employed by the
% reference cited here in favor of a global argument].)
uniformLowerBound = n*(sum(1./(1:n))-sum(1./(1:(n-m))));
lowerBound = max(lowerBound,uniformLowerBound);

%% LaTeX documentation
% Per Corollary 4.2 of \cite{flajolet1992birthday}, we have that the
% expected time for the event $X_m$ of collecting $m$ of $n$ coupons via
% IID draws from the distribution $(p_1,\dots,p_n)$ satisfies
% \begin{equation} \label{eq:partialCoupon} \mathbb{E}(X_m) =
% \sum_{\ell=0}^{m-1} (-1)^{m-1-\ell} \binom{n-\ell-1}{n-m} \sum_{|L| =
% \ell} \frac{1}{1-P_L} \end{equation} with $P_L := \sum_{k \in L} p_k$.
% However, the sum \eqref{eq:partialCoupon} is generally difficult or
% impossible to evaluate in practice due to its combinatorial complexity,
% and it is desirable to produce useful bounds. \footnote{ The specific
% case $m = n$ admits an integral representation that readily admits
% numerical computation, viz. $\mathbb{E}(X_n) = \int_0^\infty \left ( 1 -
% \prod_{k=1}^n [1-\exp(-p_k t)] \right ) \ dt$. While an integral
% representation of $\mathbb{E}(X_m)$ also exists for generic $m$, it is
% also combinatorial in form and \eqref{eq:partialCoupon} (which is
% actually just the result of evaluating it symbolically) appears easier to
% compute. }
% 
% Towards this end, assume w.l.o.g. that $p_1 \ge \dots \ge p_n$, and let $c
% \le n$. (For clarity, it is helpful to imagine that $c < n$ and $p_c \gg
% p_{c+1}$, but we do not assume this.) To bound $\sum_{|L| = \ell}
% (1-P_L)^{-1}$, we first note that $\{L : |L| = \ell\}$ is the union of
% disjoint sets of the form $\{L : |L| = \ell \text{ and } L \cap [\lambda]
% = M\}$ for $M \in 2^{[\lambda]}$, where $\lambda := \min \{c,\ell\}$.
% Thus \begin{equation} \label{eq:bound1a} \sum_{|L| = \ell}
% \frac{1}{1-P_L} = \sum_{M \in 2^{[\lambda]}} \sum_{\substack{|L| = \ell
% \\ L \cap [\lambda] = M}} \frac{1}{1-P_L}. \end{equation}
% 
% Now $P_L = P_{L \cap [\lambda]} + P_{L \backslash [\lambda]}$. If we are
% given bounds of the form $\pi_* \le P_{L \backslash [\lambda]} \le
% \pi^*$, we get in turn that $$\frac{1}{1-P_{L \cap [\lambda]}-\pi_*} \le
% \frac{1}{1-P_L} \le \frac{1}{1-P_{L \cap [\lambda]}-\pi^*}.$$ If
% furthermore $\pi_*$ and $\pi^*$ depend on $L$ only via $L \cap
% [\lambda]$, then \begin{align} \label{eq:bound1b} \frac{|\{L : |L| = \ell
% \text{ and } L \cap [\lambda] = M\}|}{1-P_M-\pi_*} & \le
% \sum_{\substack{|L| = \ell \\ L \cap [\lambda] = M}} \frac{1}{1-P_L}
% \nonumber \\ & \le \frac{|\{L : |L| = \ell \text{ and } L \cap [\lambda]
% = M\}|}{1-P_M-\pi^*}. \end{align} Meanwhile, writing $\mu := |M|$ and
% combinatorially interpreting the Vandermonde identity $\sum_\mu
% \binom{\lambda}{\mu} \binom{n-\lambda}{\ell-\mu} = \binom{n}{\ell}$
% yields \begin{equation} \label{eq:bound1c} |\{L : |L| = \ell \text{ and }
% L \cap [\lambda] = M\}| = \binom{n-\lambda}{\ell-\mu} \end{equation} and
% in turn bounds of the form \begin{equation} \label{eq:bound1d} \sum_{M
% \in 2^{[\lambda]}} \binom{n-\lambda}{\ell-\mu} \frac{1}{1-P_M-\pi_*} \le
% \sum_{|L| = \ell} \frac{1}{1-P_L} \le \sum_{M \in 2^{[\lambda]}}
% \binom{n-\lambda}{\ell-\mu} \frac{1}{1-P_M-\pi^*}. \end{equation}
% 
% Now the best possible choice for $\pi_*$ is
% $P_{[\ell-\mu]+n-(\ell-\mu)}$; similarly, the best possible choice for
% $\pi^*$ is $P_{[\ell-\mu]+\min\{\lambda,n-(\ell-\mu)\}}$. This
% immediately yields upper and lower bounds for \eqref{eq:partialCoupon},
% though the alternating sign term leads to intricate expressions that are
% hardly worth writing down explicitly.
% 
% The resulting bounds are hardly worth using in some situations, and quite
% good in others. We augment them with the easy lower bound obtained by
% using the uniform distribution in \eqref{eq:partialCoupon}
% \cite{anceaume2015new} and the easy upper bound obtained by taking $m =
% n$; we also use the exact results when feasible (e.g., $n$ small or $m =
% n$) as both upper and lower bounds. These basic augmentations have a
% significant effect in practice.
% 
% Experiments on exactly solvable (in particular, small) cases show that
% though the bounds for $(1-P_L)^{-1}$ are good, the combinatorics involved
% basically always obliterates the overall bounds for distributions of the
% form $p_k \propto k^{-\gamma}$ with $\gamma$ a small positive integer.
% However, the situation improves dramatically for distributions that decay
% quickly enough.
% 
% We can similarly also derive bounds along the lines above based on the
% deviations $\delta_k := n p_k - 1$. The only significant difference in
% the derivation here \emph{versus} the one detailed above is that we are forced
% to consider absolute values of the deviations. In this regime we also
% have the simple and tight lower bound $\mathbb{E}(X_m) \ge n(H_n -
% H_{n-m})$, where $H_n := \sum_{k=1}^n k^{-1}$ is the $n$th harmonic
% number \cite{flajolet1992birthday,anceaume2015new}. In fact this bound is
% quite good for $n$ small, to the point that replacing harmonic numbers
% with logarithms can easily produce larger deviations from the bound than
% the error itself.
\end{verbatim}
}

\subsection{goExplorePerfEval.m}

{\fontsize{6}{7}
\begin{verbatim}
function [qd,wqd,numEvals,mag,misc] = ...
    goExplorePerfEval(history,dis,varargin)

% Optional third, fourth argument of minObjective and maxObjective to
% facilitate benchmarking
%
%     figure; plot(numEvals,wqd,'b.-',numEvals,qd,'r.-')
%
% Last modified 20220630 by Steve Huntsman
%
% Copyright (c) 2022, Systems & Technology Research. All rights reserved.

numEpochs = max([history.birth]);
% Only nominal error checking for minObjective, maxObjective arguments
narginchk(2,4)  % optional argument for minObjective, maxObjective
if nargin == 4
    minObjective = varargin{1};
    maxObjective = varargin{2};
elseif nargin == 3
    minObjective = varargin{1};
    maxObjective = max([history.objective]);
else
    minObjective = min([history.objective]);
    maxObjective = max([history.objective]);
end
qd = nan(1,numEpochs);
wqd = nan(1,numEpochs);
numEvals = nan(1,numEpochs);
mag = nan(1,numEpochs);

%%
misc = [];

for j = numEpochs:-1:1  % go backwards to get proper scale
    eliteInd = find(([history.reign]>=j)&([history.birth]<=j));
    elite = {history(eliteInd).state};
    objective = [history(eliteInd).objective];

    %% Dissimilarity matrix for elites
    dissimilarityMatrix = zeros(numel(elite));
    % Symmetry of dis is assumed (as otherwise the connection between
    % diversity saturation and weightings breaks down); hence symmetry is
    % built into dissimilarityMatrix
    for j1 = 1:numel(elite)
        for j2 = (j1+1):numel(elite)
            dissimilarityMatrix(j1,j2) = dis(elite{j1},elite{j2});
        end
    end
    dissimilarityMatrix = max(dissimilarityMatrix,dissimilarityMatrix');
    
    %% Diversity-maximizing distribution on elites AT COMMON SCALE
    if j == numEpochs
        t = posCutoff(dissimilarityMatrix)*(1+sqrt(eps));
    end
    Z = exp(-t*dissimilarityMatrix);
    w = Z\ones(size(Z,1),1);
    if any(w<0)
        warning(['min(w) = ',num2str(min(w)),' < 0: adjusting post hoc']); 
        w = w-min(w); 
    end

    %% Cell counts etc
    cellCount = nan(size(w));
    cellMax = nan(size(w));
    cellMin = nan(size(w));
    for k = 1:numel(elite)
        baseCell = history(eliteInd(k)).cell';
        inBase = find(ismember([history.cell]',baseCell,'rows'));        
        cellCount(k) = numel(inBase);
        cellMax(k) = max([history(inBase).objective]);
        cellMin(k) = min([history(inBase).objective]);
    end    
    
    %% Performance metrics per epoch
    qd(j) = sum(maxObjective-objective)/(maxObjective-minObjective);
    wqd(j) = (maxObjective-objective)*(w*numel(w)/sum(w))/(maxObjective-minObjective);
    numEvals(j) = nnz([history.birth]<=j);
    mag(j) = sum(w);                                        % magnitude
    % Miscellany
    misc(j).objective = objective;
    misc(j).w = w;
    misc(j).cellCount = cellCount;
    misc(j).t = t;
end
\end{verbatim}
}

%
%
%
%
%
%
%

\end{document}